\documentclass[journal]{IEEEtran}
\usepackage{cite}
\usepackage{amsmath,amssymb,amsfonts}
\usepackage{algorithmic,algorithm}
\usepackage{graphicx}
\usepackage{textcomp}
\usepackage{xcolor}
\usepackage{amsmath}
\usepackage{mathtools}
\usepackage{lineno}
\usepackage{booktabs}       
\usepackage{nicefrac}       
\usepackage{url}
\usepackage[hidelinks]{hyperref}
\usepackage{microtype} 
\usepackage{multirow}
\usepackage{array}
\usepackage{color} 

\begin{document}
%
\title{MEL: Efficient Multi-Task Evolutionary Learning for High-Dimensional Feature Selection}
%
%
%

\author{Xubin Wang, Haojiong Shangguan, Fengyi Huang, Shangrui Wu and Weijia Jia$^*$,~\IEEEmembership{Fellow,~IEEE}

\thanks{Corresponding authors (*): Weijia Jia.}
\thanks{Xubin Wang, Shangrui Wu and Weijia Jia are with the Hong Kong Baptist University and the Institute of Artificial Intelligence and Future Networks, Beijing Normal University (Zhuhai campus). Haojiong Shangguan and Fengyi Huang are with Hong Kong Baptist University. E-mail:  wangxb19@mails.jlu.edu.cn (Xubin Wang); shangguanhaojiong@uic.edu.cn (Haojiong Shangguan); huangfengyi@uic.edu.cn (Fengyi Huang); wushangrui@uic.edu.cn (Shangrui Wu);
jiawj@bnu.edu.cn (Weijia Jia).}

\thanks{Manuscript received March 10, 2023; revised June 26, 2023.}}



\maketitle

\begin{abstract}
Feature selection is a crucial step in data mining to enhance model performance by reducing data dimensionality. However, the increasing dimensionality of collected data exacerbates the challenge known as the ``curse of dimensionality", where computation grows exponentially with the number of dimensions. To tackle this issue, evolutionary computational (EC) approaches have gained popularity due to their simplicity and applicability. Unfortunately, the diverse designs of EC methods result in varying abilities to handle different data, often underutilizing and not sharing information effectively. In this paper, we propose a novel approach called PSO-based Multi-task Evolutionary Learning (MEL) that leverages multi-task learning to address these challenges. By incorporating information sharing between different feature selection tasks, MEL achieves enhanced learning ability and efficiency. We evaluate the effectiveness of MEL through extensive experiments on 22 high-dimensional datasets. Comparing against 24 EC approaches, our method exhibits strong competitiveness. Additionally, we have open-sourced our code on GitHub \footnote{https://github.com/wangxb96/MEL}.
\end{abstract}


%
\IEEEpeerreviewmaketitle

\section{Introduction}
%
%
%
%

\IEEEPARstart{W}{ith} the rapid advancement of technology, the prevalence of large-scale and high-dimensional data is becoming increasingly common in various real-world applications. A notable example is DNA microarrays, which yield a vast amount of information regarding gene sequences in a single test \cite{glaab2009arraymining}. While high-dimensional features offer the potential to define data samples more precisely, they also present several challenges, including model overfitting \cite{bermingham2015application}, model complexity \cite{james2013introduction}, and lengthy model training times \cite{li2017feature}. Moreover, the computational complexity during model construction increases exponentially with dimensionality, a phenomenon often referred to as the ``curse of dimensionality" in machine learning and data mining \cite{kramer1991nonlinear}. Consequently, there is an escalating demand for the development of efficient feature selection techniques to effectively handle high-dimensional data.
 
Feature selection is an effective method to deal with such problems \cite{guyon2003introduction}. There are generally three types of methods used for feature selection: filter methods \cite{yu2003feature}, wrapper methods \cite{phuong2005choosing}, and embedded methods \cite{saghapour2017novel}. Filter methods evaluate features based on intrinsic properties of the data without involving any learning algorithm \cite{guyon2003introduction}. Common filter criteria include information measures such as variance, correlation with labels, and mutual information \cite{guyon2003introduction} \cite{yu2003feature}. For example, features with near-zero variance do not differentiate samples and can be eliminated. Welch's t-test and eigenvector centrality are also widely used for filter-based feature selection \cite{zhang2015detection} \cite{roffo2016features}. Filter methods are fast but ignore interactions between features and the predictive model. Wrapper and embedded methods take the predictive model into account during feature selection. Essentially, the feature selection method can be viewed as a combinatorial optimization problem, where $2^n$ - 1 feature combinations exist for data containing $n$-dimensional features. Wrapper methods treat feature selection as a search problem and evaluate feature subsets using resampling methods and predictive accuracy on a held-out validation set \cite{kohavi1997wrappers}. Brute force and sequential search algorithms like sequential forward selection (SFS) and sequential backward selection (SBS) are commonly used \cite{figueroa2013learning} \cite{alenezi2021majority}. However, these deterministic searches can get stuck in local optima. Stochastic search methods like genetic algorithm (GA) \cite{oh2004hybrid}, ant colony optimization (ACO) \cite{ma2021two} and particle swarm optimization (PSO) \cite{wang2022feature}. Nevertheless, evolutionary algorithms incur high computational overhead. Embedded methods perform feature selection during model training by assigning feature importance weights, e.g., with decision tree based methods and regularization terms in logistic regression \cite{zhang2021fs} \cite{bach2008bolasso}. 

The goal of feature selection is to identify a compact feature subset that maximizes classification accuracy while reducing dimensionality. However, feature selection is an NP-hard problem, especially in high-dimensional datasets, as the computational cost of finding the optimal solution can be prohibitively expensive \cite{kohavi1997wrappers}. Recently, evolutionary computation (EC) approaches have received significant attention for feature selection due to their ability to effectively explore large search spaces and obtain approximate solutions \cite{xue2019self} \cite{wang2022self} \cite{zhou2022lagam}. Nevertheless, high-dimensional problems pose substantial challenges. The large search space exacerbates the ``curse of dimensionality", rendering exhaustive search intractable. Additionally, the increased risk of getting trapped in local optima hinders EC methods' ability to efficiently solve real-world feature selection problems on high-dimensional data, due to their inherently high computational complexity. Therefore, more work is still needed to develop EC techniques that can better scale to large-scale, high-dimensional datasets while mitigating computational costs and avoiding premature convergence. 

Particle swarm optimization (PSO) is an effective metaheuristic for optimization problems. However, in high dimensions, its performance degrades as particles become sparse in the vast search space \cite{xue2012particle}. Multi-task learning (MTL) can leverage common patterns across related tasks to boost learning \cite{zhang2021survey}. When combined, MTL and PSO offer notable advantages for addressing high-dimensional feature selection problems. First, MTL maintains a population of interacting sub-swarms, retaining diversity essential for exploring complex landscapes. Knowledge transfer across sub-swarms circumvents premature convergence. Second, MTL guides particles toward regions exhibiting consistency across tasks. This favors selection of generalizable features and mitigates overfitting single tasks. Third, cooperative co-evolution accelerates convergence by sharing informative samples between sub-swarms. This enhances PSO's abilities to construct high-quality solutions efficiently. In this paper, we aim to explore the potential advantages of integrating MTL with PSO to improve the learning ability and efficiency of the algorithm.

To summarize, we utilize three E's (Easy, Effective, Efficient) to encapsulate our contributions in this work:

\begin{itemize} 
    \item \textbf{Easy:} We propose a simple yet effective PSO-based Multi-task Evolutionary Learning (MEL) approach for high-dimensional feature selection. MEL leverages multi-task learning to jointly optimize related tasks and transfer knowledge between them without complex operations.
    \item \textbf{Effective:} Extensive experiments on 22 benchmark datasets demonstrate MEL significantly outperforms 24 state-of-the-art EC algorithms. Results show improvements in both classification performance and compact feature subset selection.
    \item \textbf{Efficient:} By dividing the search into cooperative subtasks, MEL learns feature importance across tasks to guide optimization while sharing information efficiently. Experiments reveal MEL not only achieves competitive accuracy but also faster execution times than standard PSO on high-dimensional problems.
\end{itemize}

\section{Related Work}
\subsection{Particle Swarm Optimization}
The particle swarm optimization (PSO) algorithm was developed by Eberhart and Kennedy and mimics the behavior of a flock of birds foraging for food \cite{kennedy1995particle}. Assume a $\mathcal{D}$-dimensional search space and a particle swarm containing $\mathcal{N}$ particles searching for the global optimal solution under its constraints, each particle contains the information of three $\mathcal{D}$-dimensional vectors, namely: velocity vector $ \Vec{\mathcal{V}_{i}} = \{\Vec{\mathcal{V}}_{i}^{1}, \Vec{\mathcal{V}}_{i}^{2}, ..., \Vec{\mathcal{V}}_{i}^{\mathcal{D}} \}$, position vector $ \Vec{\mathcal{X}}_{i} = \{\Vec{\mathcal{X}}_{i}^{1}, \Vec{\mathcal{X}}_{i}^{2}, ..., \Vec{\mathcal{X}}_{i}^{\mathcal{D}}\}$ and its own optimal position vector $ \Vec{\mathcal{P}}_{i} = \{\Vec{\mathcal{P}}_{i}^{1}, \Vec{\mathcal{P}}_{i}^{2}, ..., \Vec{\mathcal{P}}_{i}^{\mathcal{D}} \}$. The velocity vector represents the distance that the particle has passed in each dimension, and the optimal position vector of itself is the best position that each particle has reached individually. The population should consider the global optimal position, i.e., the optimal position of the particle in the population that makes the best fitness value, denoted as $ \Vec{\mathcal{P}}_{g} = \{\Vec{\mathcal{P}}_{g}^{1}, \Vec{\mathcal{P}}_{g}^{2}, ..., \Vec{\mathcal{P}}_{g}^{\mathcal{D}} \}$. The state of a particle is characterized by two factors, its position and velocity, whose update rules are expressed by the following two equations:
\begin{equation}
\Vec{\mathcal{V}}_{i}^{k} = {\omega}\Vec{\mathcal{V}}_{i}^{k-1} + {c_{1}}{r_{1}}(\Vec{\mathcal{P}}_{i}^{k-1}-\Vec{\mathcal{X}}_{i}^{k-1}) + {c_{2}}{r_{2}}(\Vec{\mathcal{P}}_{g}^{k-1}-\Vec{\mathcal{X}}_{i}^{k-1}) 
\end{equation}
\begin{equation}
\Vec{\mathcal{X}}_{i}^{k} = \Vec{\mathcal{X}}_{i}^{k-1}+\Vec{\mathcal{V}}_{i}^{k-1}
\end{equation}
where $k$ denotes the number of iterations; $i$ is the $ith$ particle in the population; $\omega$ is the inertia weight, which indicates the degree of the particle influenced by itself, and it can adjust the flight speed of the particle so that the particle tends to converge; $c_{1}$ and $c_{2}$  are learning factors, which indicate the degree of the particle influenced by individual experience and population experience, respectively.

PSO has been widely used for feature selection due to its simplicity and ease of implementation. A few early studies have applied the basic PSO to select features in medium-scale datasets. For example, Prasad \textit{et al.} \cite{prasad2015gene} used a PSO-based method to refine the gene space to a fine grained one from microarray data. Cheng \textit{et al.} \cite{cheng2014illumination} introduced PSO-SQI for illumination normalization using PSO to select features in quotient images. However, directly applying PSO becomes inefficient for high-dimensional datasets due to the large search space. To address this issue, several variations of PSO have been proposed for feature selection \cite{hu2020multiobjective} \cite{mistry2016micro}. While these methods achieved improved performance over the basic PSO, they did not fully address the computational challenges of high-dimensionality. Directly evaluating all features in each iteration remains costly as the dimensionality increases. Additionally, existing PSO variants do not take into account the dependencies between features.

\subsection{Multi-task Learning}
\textbf{Definition 1 (\textit{Multi-task Learning})}. \textit{A machine learning method based on shared representation, which learns $n$ related tasks $\{\mathcal{T}_i\}^n_{i=1}$ together and uses the association information between tasks $\{\mathcal{T}_i\}^n_{i=1}$ to improve generalization} \cite{zhang2021survey}.

Multi-task learning is very related to transfer learning. Unlike transfer learning, which focuses more on improving the target task, multi-task learning emphasizes on improving the performance of each task equally through shared information \cite{zhang2021survey} \cite{caruana1998multitask}. A key idea in multi-task learning is that the tasks are related and information can be shared across tasks. A very interesting real-life example is that if a person learns to play ping pong, he may soon be able to learn to play tennis. After he learns to play tennis, these skills will also help him improve his performance in playing badminton. The common skills acquired in the process of learning these sports can help improve performance in each of them.

There are a few key approaches in multi-task learning. One approach is parameter sharing, where parts of the machine learning model, such as the lower layers of a neural network, are shared among all tasks during training. Tasks can also be clustered based on relatedness, with more related tasks sharing more parameters. Multi-task learning is considered to be a promising field of machine learning and has achieved a wide range of applications. For example, it has been applied to multimodal fields \cite{hu2021unit}, deep reinforcement learning \cite{yu2020gradient}, hotspot detection \cite{chen2020semisupervised}, App survival prediction \cite{zhang2020app}, and natural language processing \cite{alqahtani2020multitask}. Chen \textit{et al.} \cite{chen2018multi} jointly predicted the status of MGMT and IDH1 genes using a multi-task feature selection approach. Moreover, Liu \textit{et al.} proposed a graph-guided regularization multi-task learning method to perform feature selection \cite{liu2018multitask}.

Recently, we have noticed an increasing number of studies applying multi-task learning in evolutionary computation (EC) field to improve performance \cite{zhang2022task} \cite{liu2022evolutionary} \cite{he2022multitask} \cite{zheng2022evolutionary}. Moreover, feature selection methods based on evolutionary multitasking have also been proposed. For instance, Zhang \textit{et al.} \cite{zhang2021evolutionary} designed a multitask evolutionary machine learning framework (MEML) and conducted a case study on feature selection. However, they mainly focus on large-instance data and do not study high-dimensional data enough. In their experiments, the medium-sized data selected is smaller than or equal to 500 dimensions, with the exception of a 5,000-dimensional data. In large-sized data, all data is smaller than 200 dimensions. Chen \textit{et al.} \cite{chen2020evolutionary} proposed an evolutionary multitask feature selection method for high-dimensional classification by evaluating the importance of features and establishing two related tasks about the concept of goals. They designed a crossover operator for information sharing, and proposed two mechanisms called variable-range strategy and subset updating to reduce the searching space and increase the diversity of population. In their subsequent research \cite{chen2021evolutionary}, they start from the perspective of transforming high-dimensional feature selection problems into multiple low-dimensional feature selection problems, and design a new task generation strategy and a knowledge transfer mechanism. We note that in the above two studies, ReliefF was used to calculate the importance of features and a knee point selection scheme was designed to select the appropriate subset of features. However, as they also mention in the article, selecting an appropriate subset requires domain knowledge. If a feature subset is obtained by preselection through a inappropriate threshold, the valid information in the original dataset is likely to be lost. At the same time, they designed many mechanisms to ensure the effectiveness of the proposed algorithm, which also increased the complexity of the algorithm.

\section{Methodology}
\subsection{Overview of the Proposed MEL}

\begin{figure}[!htb]
\centering
\includegraphics[scale = 0.222]{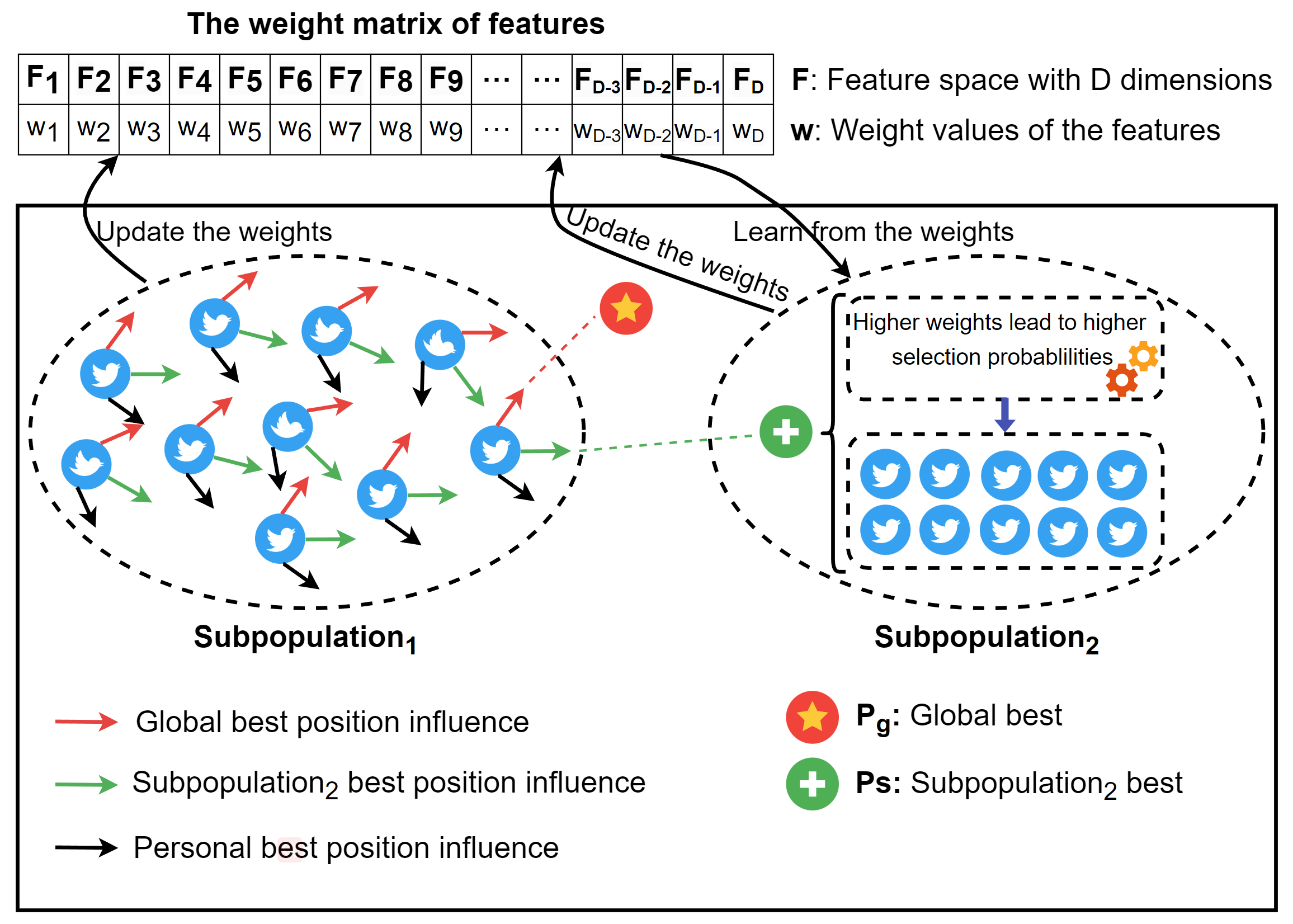}
\caption{A schematic diagram of the proposed MEL method. The parent population is divided into two subpopulations: $\Vec{Sub_1}$learns the feature importance during evolution, and its search is affected by $\Vec{Sub_2}$ best. $\Vec{Sub_2}$ also learns the importance of features during evolution, and searches for the optimal feature subset based on the results learned from $\Vec{Sub_1}$ and $\Vec{Sub_2}$. In particular, features with higher weights have a higher probability of being selected.}
\end{figure}

In this study, we develop an evolutionary learning approach that incorporates multitasking learning for high-dimensional feature selection. The main idea is to divide the parent population into two subpopulations, dubbed $subpopulation_1$ ($\Vec{Sub_1}$) and $subpopulation_2$ ($\Vec{Sub_2}$), each dedicated to searching for an optimal feature subset independently in its own search mode. To illustrate this process clearly, we provide a visual representation in Figure 1. Figure 1 demonstrates that the particles in $\Vec{Sub_1}$ are influenced by $\Vec{Sub_2}$ during the search process. This means that the knowledge and experience gained by $\Vec{Sub_2}$ will guide the search of $\Vec{Sub_1}$. Specifically, $\Vec{Sub_1}$ conducts its search and updates the learned feature importance. Simultaneously, $\Vec{Sub_2}$ also learns the feature importance during its search process. Furthermore, the search of $\Vec{Sub_2}$ is guided by the combined feature importance learned by both subpopulations. In this guidance mechanism, features with higher weights have a greater probability of being selected. To provide a comprehensive understanding of the proposed approach, we outline the detailed steps in Algorithm 1.

\begin{algorithm}[htb]
\caption{Multi-Task Evolutionary Learning Method}
\label{alg:algorithm}
\textbf{Input}: 
NP: the size of population, $\mathcal{T}$: the number of iterations.\\
\textbf{Output}: num: The number of selected features; acc: The classification accuracy.
\begin{algorithmic}[1] 
\STATE \textbf{Initialize} a PSO population with NP individuals;
\STATE \textbf{Initialize} the weights of all features to 0;
\STATE \textbf{Divide} the population into two equal-sized subpopulations $\Vec{Sub_1}$ and $\Vec{Sub_2}$ that each perform a task;
\STATE \textbf{Evaluate} each particle, the individual best $\Vec{\mathcal{P}}_i$, subpopulation best $\Vec{\mathcal{P}}_s$ and global best $\Vec{\mathcal{P}}_g$ are recorded;
\WHILE{t $\leq$ $T$}
     \STATE \textbf{$\Vec{Sub_1}$}: \textbf{Get} the knowledge from $\Vec{Sub_2}$ and search based on it; \textbf{Update} the weights of all features, $\Vec{\mathcal{P}}_i$, $\Vec{\mathcal{P}}_s$ and $\Vec{\mathcal{P}}_g$;
     \STATE \textbf{$\Vec{Sub_2}$}: The search is performed based on the learned feature weights, and features with higher weights have a higher probability of being selected; \textbf{Update} the weights of all features, $\Vec{\mathcal{P}}_i$, $\Vec{\mathcal{P}}_s$ and $\Vec{\mathcal{P}}_g$;
\ENDWHILE
\STATE \textbf{return} num, acc $\xleftarrow{}$ $\Vec{\mathcal{P}}_g$.
\end{algorithmic}
\end{algorithm}

\subsection{Evolutionary Initialization}
In the particle swarm optimization population, each particle $\Vec{\mathcal{P}^i}$ can be represented as: $\Vec{\mathcal{P}^i} = \{ \mathcal{F}_1, \mathcal{F}_2, ..., \mathcal{F}_{\mathcal{D}}\}$, where $\mathcal{F}$ is the feature space, $\mathcal{D}$ denotes dimensions of the input features. We then initialize the particles randomly, with the value of each particle expressed as a real number. To select a subset of features for evaluation in evolution, we use a threshold value $\theta$ for binarization operation to determine the selected features. In particular, if $\mathcal{F}_n$ is greater than $\theta$, it means that the corresponding feature is selected for evaluation; otherwise, the corresponding feature is not selected. At the same time, we set a weight matrix to record the weight of each feature. Initially, the weights of all features are set to 0. 
After initializing the population, we record the individual best ($\Vec{\mathcal{P}}_i$), global best ($\Vec{\mathcal{P}}_g$), and subpopulation best ($\Vec{\mathcal{P}}_s$) solutions. Subsequently, each particle's selected feature subset is evaluated, and based on the evaluation results, the feature weights are updated.

\subsection{Knowledge Learning}
In this section, we aim to address the following fundamental question, which serves as the core motivation behind our algorithm:

\textbf{Question 1. \textit{How can we identify valuable features?}}

High-dimensional data typically contains a mixture of relevant, irrelevant and redundant features \cite{guyon2003introduction}. This motivates us to select as many relevant features as possible while avoiding irrelevant and redundant ones. Ideally, all the selected features would be relevant. However, in wrapper methods, feature subsets are typically selected and evaluated by a learner, making it challenging to directly measure the importance of each feature. Additionally, the evolutionary process is iterative. Hence, we propose utilizing historical information to learn the importance of features and employ this knowledge in subsequent evolutionary iterations to achieve improved learning outcomes. Consequently, this study incorporates the concept of feature importance, enabling us to identify valuable features through their respective importance scores.

\textbf{Definition 2 (\textit{Feature Importance}).} \textit{The importance of a feature is measured by the change in accuracy resulting from its appearance or disappearance during the evolutionary process.}

To provide a clearer explanation of feature importance, we will further elaborate on the above definition. Specifically, after one iteration of evolution, if the classification accuracy of an individual $\Vec{\mathcal{P}^i}$ improves, the weight of the newly selected features $\mathcal{F}_n\uparrow$ in that iteration will be increased by the increment in accuracy. Conversely, the weight of the discarded features $\mathcal{F}_n\downarrow$ in that iteration will be decreased by the same amount of increment. On the other hand, if the classification accuracy of an individual $\Vec{\mathcal{P}^i}$ decreases, the weight of the newly selected features $\mathcal{F}_n\uparrow$ in that iteration will be decreased by the decrement in accuracy, while the weight of the discarded features $\mathcal{F}_n\downarrow$ in that iteration will be increased by the same amount of decrement. We denote $acc_i$ as the accuracy obtained in the current generation and $acc_{i-1}$ as the accuracy obtained in the previous generation. For each evaluation, two cases are considered: 

\textbf{Case 1: \textit{The $acc_i$ obtained by $\Vec{\mathcal{P}^i}$ is greater than $acc_{i-1}$.}}

\begin{equation}
     \mathcal{W}(\mathcal{F}_n) =\begin{dcases}
\mathcal{W}(\mathcal{F}_n) + (acc_i - acc_{i-1}),&\mathcal{F}_n \uparrow\\
\mathcal{W}(\mathcal{F}_n) - (acc_i - acc_{i-1}),&\mathcal{F}_n \downarrow 
\end{dcases}
\end{equation}

\textbf{Case 2: \textit{The $acc_i$ obtained by $\Vec{\mathcal{P}^i}$ is less than $acc_{i-1}$.}}
\begin{equation}
     \mathcal{W}(\mathcal{F}_n) =\begin{dcases}
\mathcal{W}(\mathcal{F}_n) - (acc_{i-1} - acc_i),&\mathcal{F}_n \uparrow\\
\mathcal{W}(\mathcal{F}_n) + (acc_{i-1} - acc_i),&\mathcal{F}_n \downarrow 
\end{dcases}
\end{equation}

We employ $\mathcal{W}(\mathcal{F}_n)$ to denote the weight of feature $\mathcal{F}_n$. The notation $\mathcal{F}_n \uparrow$ represents a feature that was not present in the previous round of evolution but has been newly selected in the current round. Conversely, $\mathcal{F}_n \downarrow$ represents a feature that was present in the previous round but has been discarded in the current round. For features that have remained unchanged, no specific action is taken. Our evaluation process for feature importance is based on each function evaluation during the evolutionary process. Unlike the approaches in \cite{chen2020evolutionary} and \cite{chen2021evolutionary}, which employ Relief to evaluate features at the initial stage, our method ensures that the information from the original data is retained. In each iteration, every feature has a probability of being selected, but features with higher weights are more likely to be chosen, indicating their significance. 

\textbf{Question 2. \textit{How can we maintain the diversity of the population?}}

In traditional EC algorithms, the evolution process often becomes trapped in local optima, which limits the algorithm's global search ability. To address this issue, the use of a multi-population mechanism has been proposed as an effective method to improve population diversity and avoid premature convergence. Several studies have demonstrated its successful application in the field of feature selection \cite{wang2022self} \cite{kilicc2021novel}. Collaboration among multi-populations presents a promising application scenario for multitask learning, which can enhance the evolution's performance. Therefore, in this study, we integrate the concept of multitask learning into the multi-population mechanism, allowing each subpopulation to independently search for the optimal feature subset using their own search strategy.

To develop an efficient yet simple method, only two subpopulations are used in our approach. Maintaining two distinctive subpopulations can help balance exploration and exploitation during the search. Additionally, since we have a population size of 20, setting too many subpopulations would result in too few particles in each subpopulation, rendering it inappropriate. Since all particles are learning knowledge (feature importance), our idea is to set up two subpopulations, where one subpopulation is influenced by the other subpopulation on the basis of the original PSO search mechanism, while the other subpopulation searches based on feature importance to increase search efficiency. Here, we give the definition of two tasks.

\textbf{Definition 3 (\textit{Task 1}).} \textit{$Subpopulation_1$ ($\Vec{Sub_1}$) conducts the search for the optimal feature subset based on the influence of individual best, $subpopulation_2$ ($\Vec{Sub_2}$) best, and global best.}

The search strategy in Task 1 can be based on the original PSO mechanism or any other suitable search algorithm. For example, each particle in $\Vec{Sub_1}$ can update its position by considering its individual best solution, the best solution found by any particle in $\Vec{Sub_2}$, and the global best solution found by any particle in the population. This strategy allows $\Vec{Sub_1}$ to explore the search space by leveraging the collective knowledge of both subpopulations.

\textbf{Definition 4 (\textit{Task 2}).} \textit{$Subpopulation_2$ ($\Vec{Sub_2}$) searches the optimal feature subset based on the feature importance information learned by the two subpopulations ($\Vec{Sub_1}$ and $\Vec{Sub_2}$).}

The search strategy in Task 2 focuses on exploiting the feature importance knowledge acquired during the learning phase. This can involve methods such as ranking the features based on their importance scores and selecting the most informative ones. $\Vec{Sub_2}$ can update its position and velocity accordingly to guide the search towards the most relevant features.

\subsection{Knowledge Transfer}
In our algorithm, each subpopulation learns independently and enhances its search ability through knowledge transfer. Specifically, we will use two cases to illustrate the knowledge transfer to $\Vec{Sub_1}$ and $\Vec{Sub_2}$ respectively.

\textbf{Case 1: \textit{Knowledge transfer to $\Vec{Sub_1}$}.}

In traditional particle swarm optimization (as shown in formulas (1) and (2)), particles modify their position and velocity by taking into account their past movements (inertia), their personal experience (cognitive state), and information about the best particle in the swarm (socialization). In our approach, which involves two subpopulations, we allow the first subpopulation $\Vec{Sub_1}$ to learn from the second subpopulation $\Vec{Sub_2}$. Here, we let $\Vec{Sub_1}$ be influenced by the subpopulation best of $\Vec{Sub_2}$. The update process is as follows:

\begin{equation}
\begin{split}
    \Vec{\mathcal{V}}_{i}^{k} &= {\omega}\Vec{\mathcal{V}}_{i}^{k-1} + {c_{1}}{r_{1}}(\Vec{\mathcal{P}}_{i}^{k-1}-\Vec{\mathcal{X}}_{i}^{k-1}) + {c_{2}}{r_{2}}(\Vec{\mathcal{P}}_{g}^{k-1}-\Vec{\mathcal{X}}_{i}^{k-1}) 
    \\&+ {c_{3}}{r_{3}}(\Vec{\mathcal{P}}_{s}^{k-1}-\Vec{\mathcal{X}}_{i}^{k-1}) 
\end{split}
\end{equation}
\begin{equation}
\Vec{\mathcal{X}}_{i}^{k} = \Vec{\mathcal{X}}_{i}^{k-1}+\Vec{\mathcal{V}}_{i}^{k-1}
\end{equation}
where $c_3$ is the learning factor learns knowledge from another subpopulation, $r_3$ is a random value range from [0,1], and $\Vec{\mathcal{P}}_{s}^{k-1}$ is the best individual in the second subpopulation $\Vec{Sub_2}$.

\textbf{Case 2: \textit{Knowledge transfer to $\Vec{Sub_2}$}.} 

When we utilize information from the feature weight matrix, we need to consider that after the operations in formulas (3) and (4), the weight values associated with each feature in the weight matrix may fall into one of three categories: less than 0, equal to 0, or greater than 0. In our algorithm, $\Vec{Sub_2}$ aims to select features for evaluation by considering their importance during the evolution process. It excludes features that have importance values less than or equal to 0, which helps in reducing evaluation costs and improving the efficiency of the feature search. This approach allows $\Vec{Sub_2}$ to concentrate its search resources on the most important features. The selection process based on feature importance in $\Vec{Sub_2}$ can be summarized as follows:
 
\begin{equation}
   \delta = \sum_{n=1}^\mathcal{D} \mathcal{W}(\mathcal{F}_n), \forall \mathcal{W}(\mathcal{F}_n) > 0
\end{equation}

We first calculate the sum of the weight values of all features whose weight values $\mathcal{W}(\mathcal{F}_n)$ are greater than 0, denoted as $\delta$. When selecting features in $\Vec{Sub_2}$, the probabilities of all features with weights less than or equal to 0 are set to 0, meaning that we do not select these features in this iteration. Here, features with weights greater than 0 are considered important features. These features are selected with the following probabilities $\rho$:

\begin{equation}
    \rho = \frac{\mathcal{W}(\mathcal{F}_n)}{\delta}, \forall \mathcal{W}(\mathcal{F}_n) \ge 0
\end{equation}

\subsection{Fitness Function}
In this study, our focus is on discovering the smallest feature subset while achieving the largest classification result. Therefore, two aspects need to be considered in the design of objective function: classification accuracy and feature subset size. Here, we aim to select a feature subset that is as small as possible to avoid the computational cost and high model complexity caused by high-dimensional features.
Simultaneously, we need to maintain an adequate number of features to ensure high classification accuracy. Based on the above considerations, we set the objective function as follows:

\begin{equation}
    f = \alpha \cdot error\_rate + \beta \frac{num\_of\_SF}{|\mathcal{F}|}
\end{equation}
where $error\_rate$ is the classification error rate, $num\_of\_SF$ denotes the size of selected features, $|\mathcal{F}|$ represents the number of features in the feature search space, $\alpha$ is the control weight of classification results and $\beta$  is the feature size control weight. The settings of parameters $\alpha$ (0.9) and $\beta$ (0.1) refer to reference \cite{chen2021evolutionary}.

\section{Implementation}
\subsection{Datasets}
We use 12 public high-dimesnional genetic datasets for our experiments \cite{yu2021ilrc}. Among these data, most of them have more than 7,000 dimensions, with the largest reaching 54,675 dimensions. They are available on Github \footnote{https://github.com/xwdshiwo}, and the details of these datasets are presented in TABLE I \footnote{The two class labels ``Pos" and ``Neg" in TABLE I represent the positive and negative class labels, respectively. }. 

\begin{table}[!htb]
\centering
\begin{tabular}{| l | l | l | l | l | }
\hline
\textbf{Dataset} & \textbf{Samples}& \textbf{Features} & \textbf{Classes (Pos / Neg)}\\
\hline \hline
Adenoma & 36 & 7457 & 18 / 18 \\ \hline
ALL$\_$AML & 72 & 7,129 & 47 / 25 \\ \hline
ALL3 & 125 & 12,625 & 24 / 101 \\ \hline
ALL4 & 93 & 12,625 & 26 / 67 \\ \hline
CNS & 60 & 7,129 & 39 / 21 \\ \hline
Colon &	62	&	2,000	&  40 / 22 \\
\hline
DLBCL & 77 & 7,129 & 58 / 19 \\ \hline
Gastric & 65 & 22,645 & 29 / 36 \\ \hline
Leukemia &	72	&	7,129	&	 47 / 25 \\
\hline
Lymphoma &	45	&	4,026	&	22 / 23\\
\hline
Prostate &	102	&	12,625	&	52 / 50	\\ 
\hline
Stroke &	40	&	54,675	&	20 / 20	\\ 
\hline
\end{tabular}
\caption{12 high-dimensional genetic datasets.}
\end{table}

To assess the impact of our proposed algorithm on datasets with a larger number of samples, we conducted experimental evaluations using high-dimensional and larger-sample datasets from various domains. These datasets include Text Data, Face Image Data, and Hand Written Image Data. The Pancancer dataset was sourced from reference \cite{li2021identification}, while the HAPTDataSet, and MultipleFeaturesDigit datasets were obtained from reference \cite{wang2022feature}. The remaining datasets can be downloaded from the ``Feature Selection" website \footnote{https://jundongl.github.io/scikit-feature/datasets.html}. For detailed information about these datasets, please refer to Table II.

\begin{table}[!htb]
\centering
\begin{tabular}{| l | l | l | l | l | }
\hline
\textbf{Dataset} & \textbf{Samples}& \textbf{Features} & \textbf{Classes}\\
\hline \hline
BASEHOCK	&	1,993	&	4,862	&	2	\\	\hline
COIL20	&	1,440	&	1,024	&	20	\\	\hline
HAPTDataSet	&	1,200	&	561	&	12	\\	\hline
Isolet	&	1,560	&	617	&	26	\\	\hline
madelon	&	2,600	&	500	&	2	\\	\hline
MultipleFeaturesDigit	&	1,000	&	649	&	10	\\	\hline
Pancancer	&	4,759	&	20,486	&	2	\\	\hline
PCMAC	&	1,943	&	3,289	&	2	\\	\hline
RELATHE	&	1,427	&	4,322	&	2	\\	\hline
USPS	&	9,298	&	256	&	10	\\	\hline
\end{tabular}
\caption{10 high-dimensional datasets with thousands of samples.}
\end{table}

\subsection{Baselines}
\subsubsection{Representative Classic Methods}
To illustrate the performance of our method, 18 evolutionary computation algorithms were employed. According to the classification methods in literature \cite{kumar2020comparative}, we selected four \textbf{swarm-based methods} (inspired by mutual behavior of swarm creatures), four \textbf{nature-inspired methods} (inspired by natural system), two \textbf{evolutionary algorithms} (inspired by natural selection concepts), four \textbf{bio-stimulated methods} (inspired by the foraging and hunting behavior in the wild) and four \textbf{physics-based methods} (inspired by physical rules). 
These meta-heuristic optimization methods basically represent the most classical, representative, and widely used methods in the field. 

\begin{itemize}
    \item \textbf{Swarm-based methods}: Artificial Bee Colony (ABC) \cite{karaboga2007powerful}, Ant Colony Optimization (ACO) \cite{aghdam2009text}, Particle Swarm Optimization (PSO) \cite{shi1998modified} and Monarch Butterfly Optimization (MBO) \cite{wang2019monarch}.
    \item \textbf{Nature-inspired methods}: Bat Algorithm (BAT) \cite{yang2010new}, Cuckoo Search Algorithm (CS) \cite{yang2009cuckoo}, Firefly Algorithm (FA) \cite{yang2010firefly} and Flower Pollination Algorithm (FPA) \cite{yang2014flower}.
    \item \textbf{Evolutionary algorithms}: Differential Evolution (DE) \cite{storn1997differential} and Genetic Algorithm (GA) \cite{huang2006ga}.
    \item \textbf{Bio-stimulated methods}: Fruit Fly Optimization Algorithm (FOA) \cite{pan2012new}, Grey Wolf Optimizer (GWO) \cite{mirjalili2014grey} and Harris Hawks Optimization (HHO) \cite{heidari2019harris}.
    \item \textbf{Physics-based methods}: Simulated Annealing (SA) \cite{kirkpatrick1983optimization}, Harmony Search (HS) \cite{geem2001new}, Gravitational Search Algorithm (GSA) \cite{rashedi2009gsa} and Multi Verse Optimizer (MVO) \cite{mirjalili2016multi}.
\end{itemize}
\subsubsection{Recently Published Evolutionary Methods}
In recent years, the field of feature selection has witnessed remarkable advancements and novel approaches. With the goal of exploring the latest developments and expanding the scope of our study, we have incorporated six evolutionary methods for feature selection published in recent four years (SaWDE \cite{wang2022self}, FWPSO \cite{wang2022feature}, VGS-MOEA \cite{cheng2022variable}, MTPSO \cite{chen2021evolutionary}, PSO-EMT \cite{chen2020evolutionary} and DENCA \cite{hancer2023evolutionary}). These methods represent the cutting-edge research in the field and offer promising solutions to the challenges at hand. By incorporating these state-of-the-art evolutionary algorithms, we aim to enhance the effectiveness and robustness of our proposed framework.

\subsection{Experimental Setup}
In Section V parts A-E, our experiments are run on a CentOS server equipped with an Intel(R) Xeon(R) Silver 4215R CPU @ 3.20GHz, 16GB of RAM using MATLAB R2019b. Furthermore, in Section V parts F-G, our experiments are run on MATLAB R2023b with a 13th generation i7-13700KF CPU and 64GB of memory. In particular, this study quantified running time in seconds. We obtained most of our baseline approaches from this online toolbox\footnote{https://github.com/JingweiToo/Wrapper-Feature-Selection-Toolbox}. \textit{To ensure the fairness of the experiment, we respect the original setting of the algorithm in the toolkit.} 
In addition, for all algorithms, we set the number of populations (NP) to 20 and the maximum number of iterations (T) to 100. At the same time, we set a threshold value ($\theta$) of 0.6 to convert the value of the real number field into a discrete value, so as to decide whether to choose the feature of the corresponding field. In this study, KNN classifier is employed with K equals to 3. Our classification accuracy is calculated using a five-fold cross-validation average. We performed each experiment ten times and averaged the data to make sure the experimental results were stable.

\section{Results and Analysis}
Our experiments included a comprehensive comparison of our method with various influential and diverse approaches in the field, such as swarm-based, nature-inspired, evolutionary algorithms, bio-inspired, and physics-based methods. This comparison aimed to demonstrate the breadth of our method's capabilities and its ability to outperform and complement existing techniques in feature selection. In addition, we specifically compared our method with the latest advancements in EC methods for feature selection published in the past four years. This comparison was crucial to assess the novelty and competitiveness of our approach against the most recent developments in the field. Furthermore, we conducted experiments on larger datasets to showcase the effectiveness of our method in handling datasets with a larger sample size. This analysis aimed to highlight the scalability and generalizability of our method, ensuring its applicability to real-world scenarios with extensive and diverse data. For each experiment, we compared and discussed the classification accuracy, feature subset size, and training time. In terms of these metrics, higher classification accuracy is desirable, while smaller feature subset size and shorter training time are preferable. Each experiment was repeated 10 times, and the average values (denoted as ``\textbf{Mean}") and standard deviations (denoted as ``\textbf{Std}") were calculated. The best results are highlighted in bold. In addition, it's worth noting that for some datasets, the results are indicated by `-' because the methods were unable to obtain results due to limitations in memory, running time or the algorithms themselves.

Overall, in Section V parts A-F, our MEL method achieved the best average performance across 12 datasets compared to the 18 EC methods and the additional five recently published EC methods specifically designed for feature selection. Our method outperformed other methods in terms of classification accuracy and ranked second in terms of feature subset size, following the FWPSO method. In terms of running time, our method performed slightly slower than SaWDE and Simulated Annealing (SA). In Section V part G, we also conducted experiments on an additional set of 10 datasets with a larger number of samples. In this case, we compared our method with five representative algorithms. The results demonstrated the superior overall performance of our method in terms of various evaluation metrics.

\subsection{Comparison with Swarm-based Heuristic Methods}
TABLE III presents the classification accuracy results of MEL and four swarm-based heuristic methods. It is evident that MEL achieves the highest classification accuracy on 8 out of 12 datasets and demonstrates the best overall performance. Notably, due to excessive memory requirements, ACO was unable to process the Stroke dataset, and therefore its results are not available for comparison. ABC and ACO exhibit strengths in handling certain datasets, while MBO's average performance ranks second only to MEL among the 18 algorithms compared. TABLE IV displays the sizes of feature subsets obtained by these algorithms. It is noteworthy that, in comparison to the second-ranked ACO algorithm, MEL achieves an average subset size of only approximately 33\%. These findings indicate that the MEL algorithm significantly reduces model complexity while maintaining high accuracy in inference.

\begin{table}[!htb]
\centering
\resizebox{87mm}{!}{
\begin{tabular}{ l l l l l l }
\hline
\multirow{2}{*}{\textbf{Dataset}} & \textbf{ABC}	&	\textbf{ACO}	&	\textbf{PSO}	&	\textbf{MBO}	&	\textbf{MEL (Ours)} \\ 
 \cline{2-6}
   \textbf{ } & \textbf{Mean $\pm$ Std} & \textbf{Mean $\pm$ Std}  & \textbf{Mean $\pm$ Std} & \textbf{Mean $\pm$ Std}  & \textbf{Mean $\pm$ Std} \\
\hline
Adenoma	&	0.9750	±	0.0000	&	\textbf{1.0000}	±	0.0000	&	0.9750	±	0.0000	&	0.9750	±	0.0000	&	0.9975	±	0.0079	\\ \hline
ALL$\_$AML	&	0.9907	±	0.0064	&	\textbf{1.0000}	±	0.0000	&	0.9907	±	0.0064	&	0.9880	±	0.0042	&	0.9907	±	0.0064	\\ \hline
ALL3	&	0.8160	±	0.0053	&	0.8142	±	0.0120	&	0.8192	±	0.0067	&	0.8296	±	0.0076	&	\textbf{0.8528}	±	0.0137	\\ \hline
ALL4	&	0.8376	±	0.0074	&	0.8495	±	0.0045	&	0.8433	±	0.0120	&	0.8542	±	0.0192	&	\textbf{0.9035}	±	0.0219	\\ \hline
CNS	&	0.7850	±	0.0254	&	0.7833	±	0.0096	&	0.8017	±	0.0214	&	0.8350	±	0.0214	&	\textbf{0.8450}	±	0.0223	\\ \hline
Colon	&	0.8969	±	0.0083	&	0.9118	±	0.0125	&	0.9049	±	0.0141	&	0.9197	±	0.0178	&	\textbf{0.9282}	±	0.0136	\\ \hline
DLBCL	&	\textbf{0.9913}	±	0.0060	&	0.9885	±	0.0035	&	0.9875	±	0.0000	&	0.9888	±	0.0040	&	0.9873	±	0.0004	\\ \hline
Gastric	&	\textbf{1.0000}	±	0.0000	&	\textbf{1.0000}	±	0.0000	&	\textbf{1.0000}	±	0.0000	&	\textbf{1.0000}	±	0.0000	&	\textbf{1.0000}	±	0.0000	\\ \hline
Leukemia	&	0.9933	±	0.0070	&	\textbf{0.9950}	±	0.0069	&	0.9893	±	0.0056	&	0.9907	±	0.0064	&	0.9917	±	0.0071	\\ \hline
Lymphoma	&	0.9400	±	0.0211	&	0.9222	±	0.0119	&	0.9422	±	0.0115	&	0.9444	±	0.0117	&	\textbf{0.9756}	±	0.0164	\\ \hline
Prostate	&	0.8805	±	0.0078	&	0.8837	±	0.0136	&	0.8886	±	0.0090	&	0.8981	±	0.0095	&	\textbf{0.9004}	±	0.0259	\\ \hline
Stroke	&	\textbf{1.0000}	±	0.0000	&	--	&	\textbf{1.0000}	±	0.0000	&	\textbf{1.0000}	±	0.0000	&	\textbf{1.0000}	±	0.0000	\\ \hline
\textbf{Average}	&	0.9255	±	0.0079	&	0.9226	±	0.0068	&	0.9285	±	0.0072	&	0.9353	±	0.0085	&	\textbf{0.9477}	±	0.0113	\\ \hline
\end{tabular}}
\caption{Comparison with Swarm-based Heuristic Methods on Accuracy}
\end{table}

\begin{table}[!htb]
\centering
\resizebox{87mm}{!}{
\begin{tabular}{ l l l l l l }
\hline
\multirow{2}{*}{\textbf{Dataset}} & \textbf{ABC}	&	\textbf{ACO}	&	\textbf{PSO}	&	\textbf{MBO}	&	\textbf{MEL (Ours)} \\ 
 \cline{2-6}
   \textbf{ } & \textbf{Mean $\pm$ Std} & \textbf{Mean $\pm$ Std}  & \textbf{Mean $\pm$ Std} & \textbf{Mean $\pm$ Std}  & \textbf{Mean $\pm$ Std} \\
\hline
Adenoma	&	2986.3	±	17.7	&	353.0	±	161.9	&	2926.6	±	130.4	&	2989.3	±	49.3	&	\textbf{271.1}	±	844.3	\\ \hline
ALL$\_$AML	&	2840.0	±	51.5	&	\textbf{695.2}	±	264.8	&	2863.6	±	134.7	&	2785.4	±	101.8	&	1050.8	±	1281.9	\\ \hline
ALL3	&	5081.0	±	55.2	&	1426.2	±	354.0	&	5188.2	±	272.4	&	4780.6	±	134.9	&	\textbf{1.0}	±	0.0	\\ \hline
ALL4	&	5015.9	±	44.1	&	1158.5	±	552.0	&	5376.9	±	195.3	&	4750.2	±	121.3	&	\textbf{34.1}	±	24.9	\\ \hline
CNS	&	2874.7	±	42.1	&	2481.4	±	1505.8	&	3023.1	±	113.3	&	2666.4	±	69.5	&	\textbf{57.2}	±	74.0	\\ \hline
Colon	&	813.5	±	19.9	&	580.5	±	442.6	&	845.1	±	32.7	&	737.6	±	19.3	&	\textbf{100.1}	±	266.1	\\ \hline
DLBCL	&	2858.0	±	45.2	&	3632.3	±	1256.2	&	2911.0	±	128.2	&	2708.5	±	77.7	&	\textbf{2697.8}	±	146.6	\\ \hline
Gastric	&	9048.1	±	64.2	&	9113.6	±	5925.7	&	9048.2	±	57.8	&	9066.9	±	89.5	&	\textbf{1.0}	±	0.0	\\ \hline
Leukemia	&	2859.4	±	51.8	&	1909.4	±	1497.8	&	2838.8	±	179.6	&	2781.1	±	87.9	&	\textbf{1793.4}	±	1236.5	\\ \hline
Lymphoma	&	1612.2	±	42.3	&	1353.1	±	731.7	&	1668.1	±	50.7	&	1512.3	±	27.4	&	\textbf{35.6}	±	46.3	\\ \hline
Prostate	&	5044.2	±	58.2	&	\textbf{2935.8}	±	3447.2	&	5390.7	±	236.2	&	4769.4	±	127.8	&	3055.8	±	2579.6	\\ \hline
Stroke	&	21870.5	±	131.2	&		--		&	21873.4	±	85.8	&	21780.4	±	151.4	&	\textbf{1.0}	±	0.0	\\ \hline
\textbf{Average}	&	5242.0	±	52.0	&	2330.8	±	1467.2	&	5329.5	±	134.8	&	5110.7	±	88.2	&	\textbf{758.2}	±	541.7	\\ \hline
\end{tabular}}
\caption{Comparison with Swarm-based Heuristic Methods on Subset Size}
\end{table}

\begin{table}[!htb]
\centering
\resizebox{87mm}{!}{
\begin{tabular}{ l l l l l l }
\hline
\multirow{2}{*}{\textbf{Dataset}} & \textbf{ABC}	&	\textbf{ACO}	&	\textbf{PSO}	&	\textbf{MBO}	&	\textbf{MEL (Ours)} \\ 
 \cline{2-6}
   \textbf{ } & \textbf{Mean $\pm$ Std} & \textbf{Mean $\pm$ Std}  & \textbf{Mean $\pm$ Std} & \textbf{Mean $\pm$ Std}  & \textbf{Mean $\pm$ Std} \\
\hline
Adenoma	&	71.0	±	1.8	&	1965.7	±	38.0	&	85.4	±	1.8	&	82.0	±	3.4	&	\textbf{55.5}	±	1.9	\\ \hline
ALL$\_$AML	&	90.2	±	2.3	&	2029.5	±	1431.3	&	101.6	±	4.2	&	94.7	±	2.2	&	\textbf{66.2}	±	5.8	\\ \hline
ALL3	&	194.7	±	4.8	&	4812.2	±	1545.9	&	207.3	±	8.3	&	192.4	±	2.8	&	\textbf{62.9}	±	2.3	\\ \hline
ALL4	&	145.7	±	1.1	&	8200.0	±	11224.5	&	166.8	±	4.8	&	150.4	±	1.7	&	\textbf{66.3}	±	5.6	\\ \hline
CNS	&	82.7	±	0.7	&	1711.0	±	817.9	&	95.3	±	2.8	&	86.1	±	1.5	&	\textbf{53.1}	±	0.9	\\ \hline
Colon	&	58.5	±	0.4	&	266.5	±	86.7	&	60.1	±	2.3	&	60.6	±	0.4	&	\textbf{44.9}	±	1	\\ \hline
DLBCL	&	92.9	±	0.9	&	1757.7	±	900.8	&	104.4	±	2.6	&	96.6	±	1.2	&	\textbf{72.3}	±	0.8	\\ \hline
Gastric	&	150.8	±	1.7	&	20462.4	±	12263.5	&	199.1	±	7.0	&	178.8	±	0.6	&	\textbf{74.4}	±	1	\\ \hline
Leukemia	&	89.9	±	1.2	&	1297.4	±	424.0	&	101.1	±	3.1	&	94.3	±	1.4	&	\textbf{66.8}	±	6.1	\\ \hline
Lymphoma	&	61.4	±	0.4	&	495.6	±	147.3	&	67.2	±	1.3	&	65.3	±	0.7	&	\textbf{48.2}	±	1.3	\\ \hline
Prostate	&	159.6	±	1.3	&	3908.6	±	1441.5	&	183.1	±	9.9	&	162.0	±	3.1	&	\textbf{98}	±	19	\\ \hline
Stroke	&	177.4	±	1.1	&		--		&	294.9	±	6.3	&	247.7	±	0.8	&	\textbf{115.7}	±	1.3	\\ \hline
\textbf{Average}	&	114.6	±	1.5	&	4264.2	±	2756.5	&	138.9	±	4.5	&	125.9	±	1.7	&	\textbf{68.7}	±	3.9	\\ \hline
\end{tabular}}
\caption{Comparison with Swarm-based Heuristic Methods on Running Time}
\end{table}

Figure 2 and Figure 3 depict the convergence curves of these algorithms in terms of classification accuracy and feature subsets, respectively. From Figure 2, it is evident that MEL exhibits superior search capability across most datasets, particularly in cases such as ALL3, ALL4, CNS, Colon, and Lymphoma. Figure 3 illustrates that both ACO and MEL demonstrate strong search abilities for relevant features. However, unlike ACO, which exhibits some fluctuations in the curve during the search process, MEL maintains better stability. TABLE V displays the running times of the algorithms on all datasets, including their average performance. It is noteworthy that MEL exhibits the shortest running time among all datasets, indicating its efficient search capability. In comparison to the original PSO algorithm, MEL achieves a running time that is nearly half. Conversely, ACO exhibits significantly longer running times across all datasets compared to other EC algorithms, suggesting that ACO may not be suitable for processing high-dimensional data.

\begin{figure*}[htp!]
  \centering
  \begin{minipage}[b]{0.1612\textwidth}
    \includegraphics[width=\textwidth]{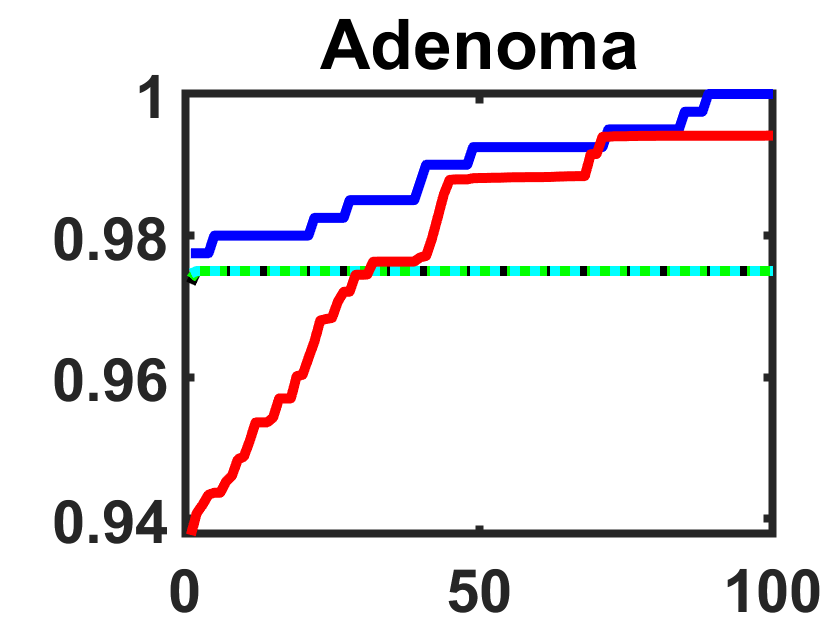}
  \end{minipage}
  \begin{minipage}[b]{0.1612\textwidth}
    \includegraphics[width=\textwidth]{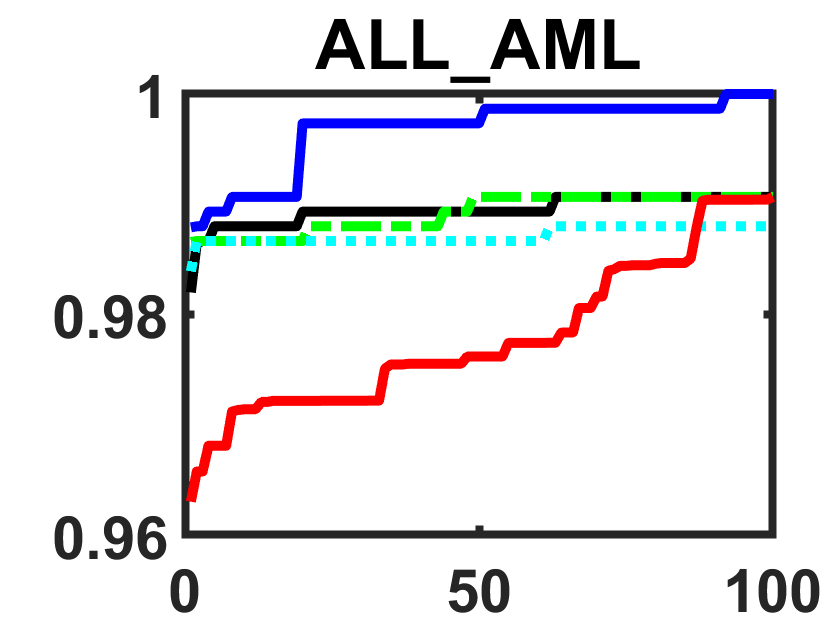}
  \end{minipage}
  \begin{minipage}[b]{0.1612\textwidth}
    \includegraphics[width=\textwidth]{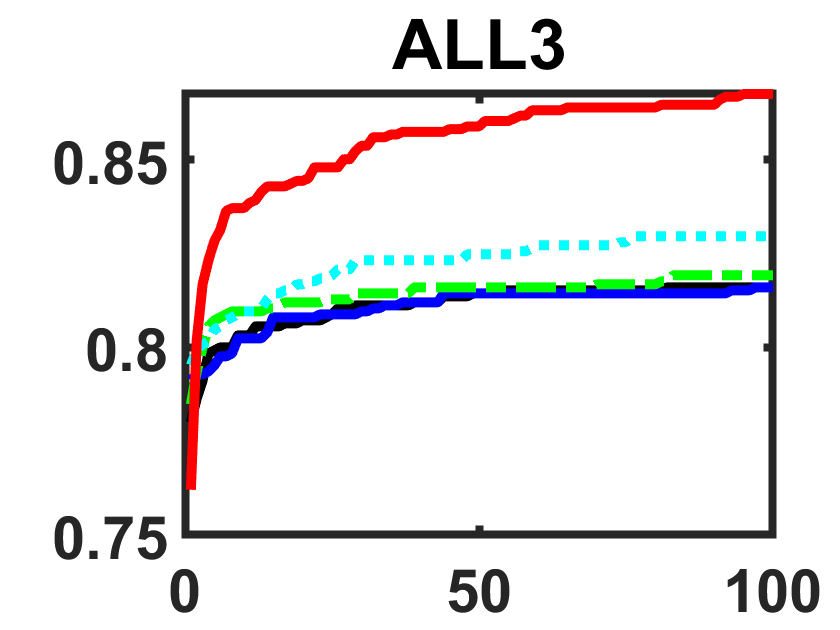}
  \end{minipage}
  \begin{minipage}[b]{0.1612\textwidth}
    \includegraphics[width=\textwidth]{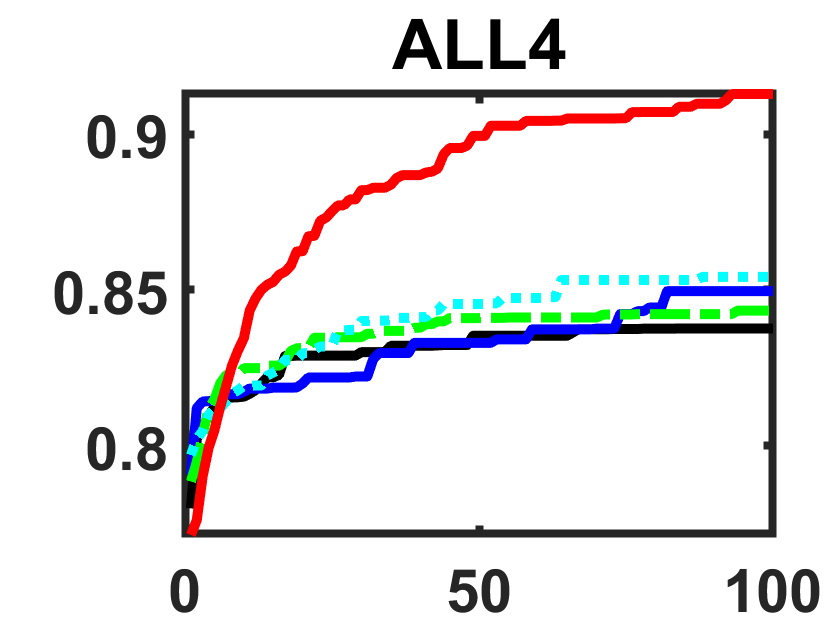}
  \end{minipage}
  \begin{minipage}[b]{0.1612\textwidth}
    \includegraphics[width=\textwidth]{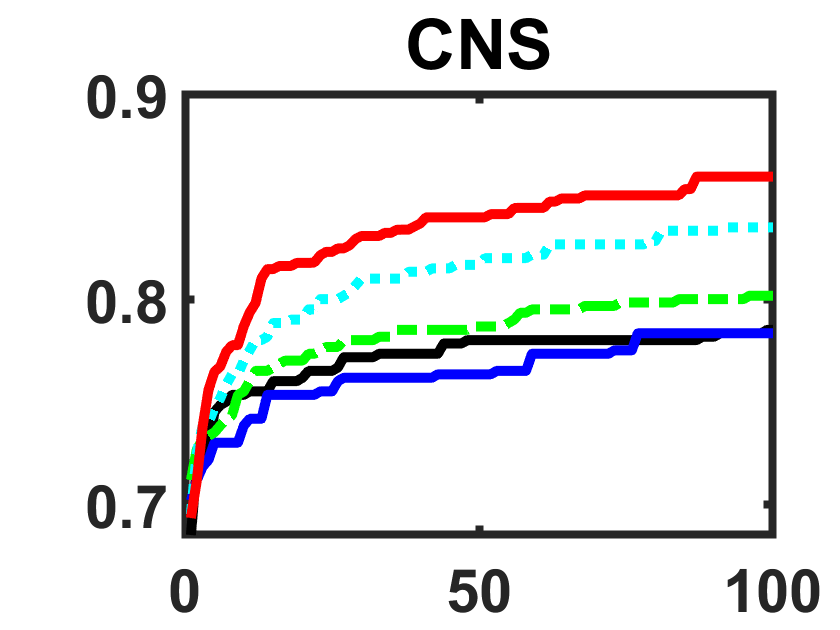}
  \end{minipage}
  \begin{minipage}[b]{0.1612\textwidth}
    \includegraphics[width=\textwidth]{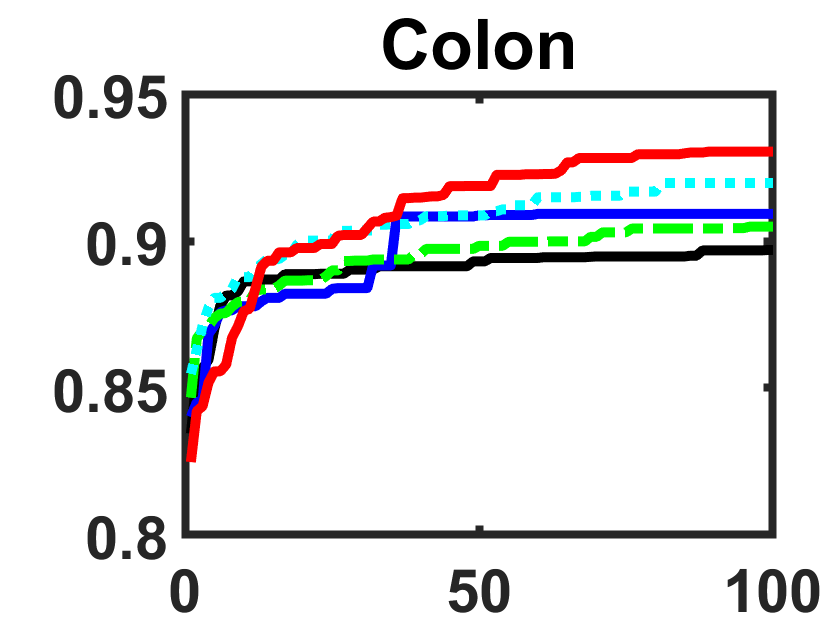}
  \end{minipage}
    \centering
  \begin{minipage}[b]{0.1612\textwidth}
    \includegraphics[width=\textwidth]{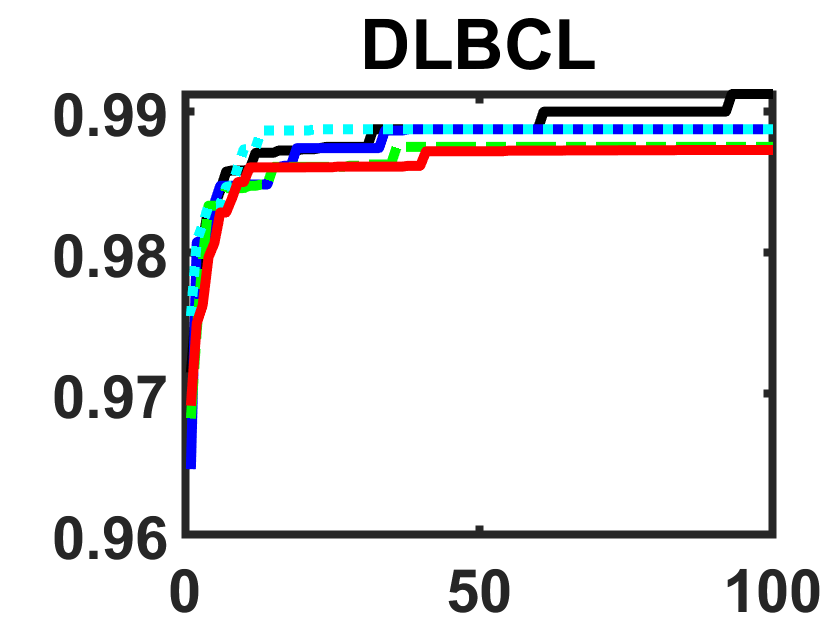}
  \end{minipage}
  \begin{minipage}[b]{0.1612\textwidth}
    \includegraphics[width=\textwidth]{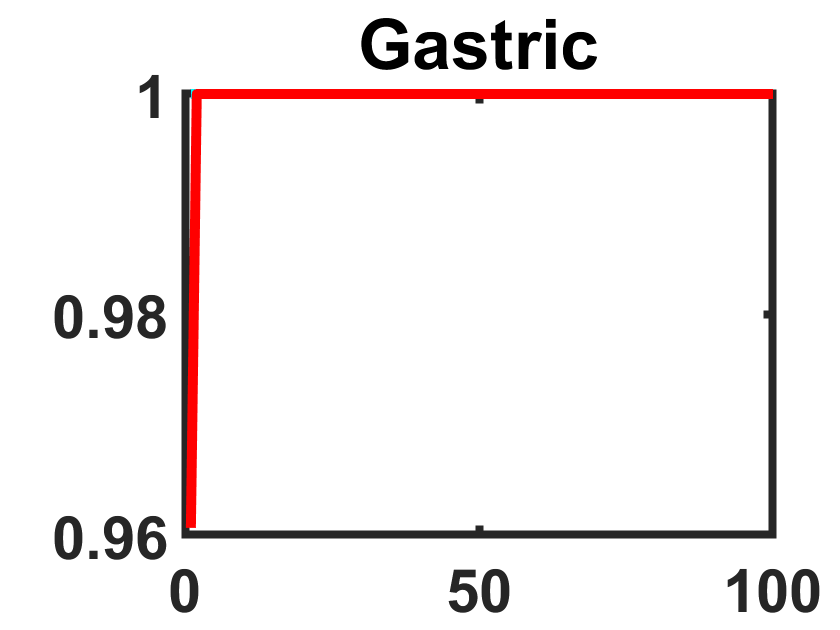}
  \end{minipage}
  \begin{minipage}[b]{0.1612\textwidth}
    \includegraphics[width=\textwidth]{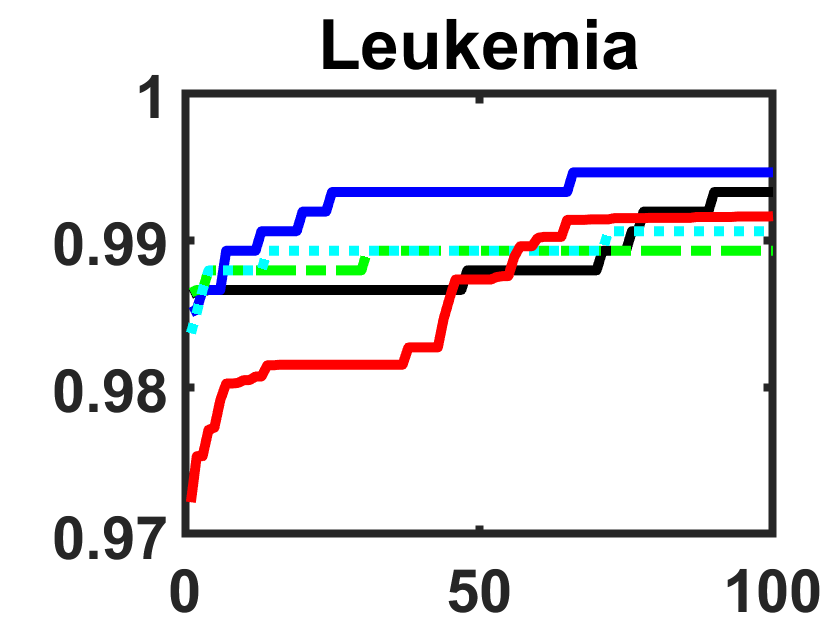}
  \end{minipage}
  \begin{minipage}[b]{0.1612\textwidth}
    \includegraphics[width=\textwidth]{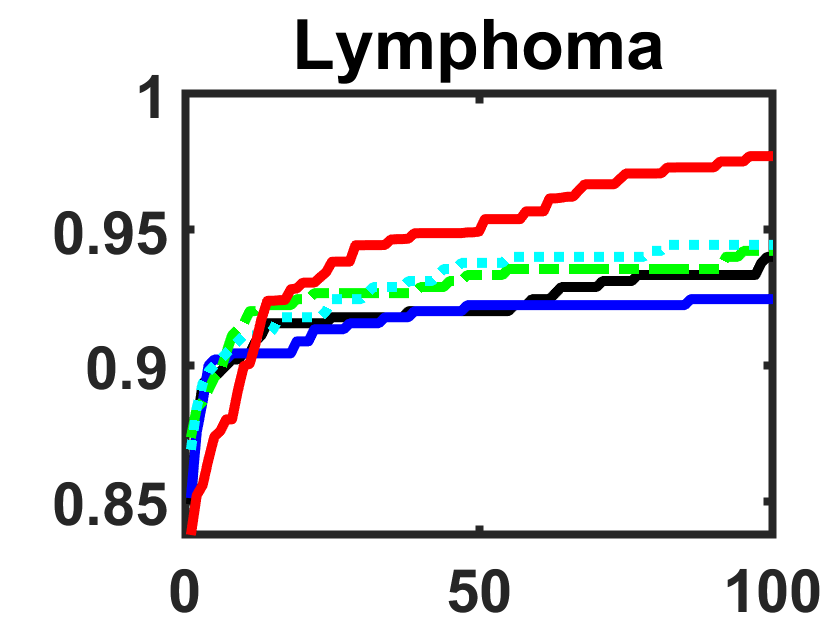}
  \end{minipage}
  \begin{minipage}[b]{0.1612\textwidth}
    \includegraphics[width=\textwidth]{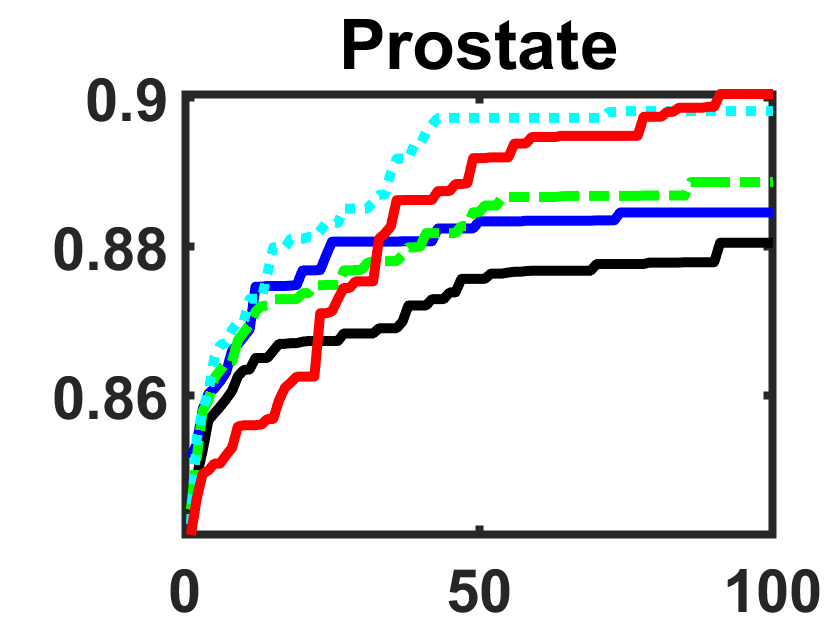}
  \end{minipage}
  \begin{minipage}[b]{0.1612\textwidth}
    \includegraphics[width=\textwidth]{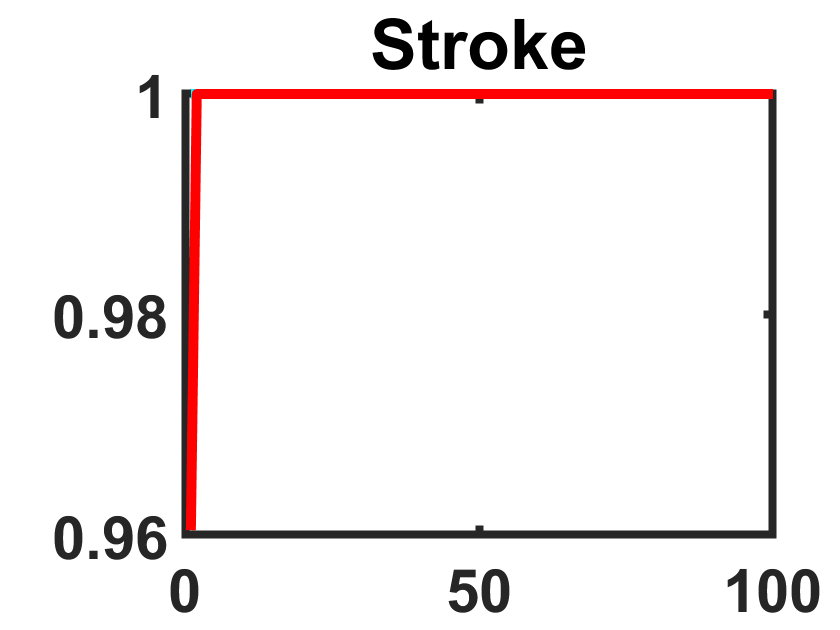}
  \end{minipage}
  \begin{minipage}[b]{0.18\textwidth}
    \includegraphics[width=\textwidth]{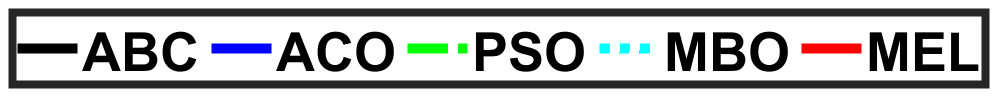}
  \end{minipage}
  \caption{Convergence Curves of Swarm-based EC Algorithms in Terms of Accuracy}
\end{figure*}

\begin{figure*}[htp!]
  \centering
  \begin{minipage}[b]{0.1612\textwidth}
    \includegraphics[width=\textwidth]{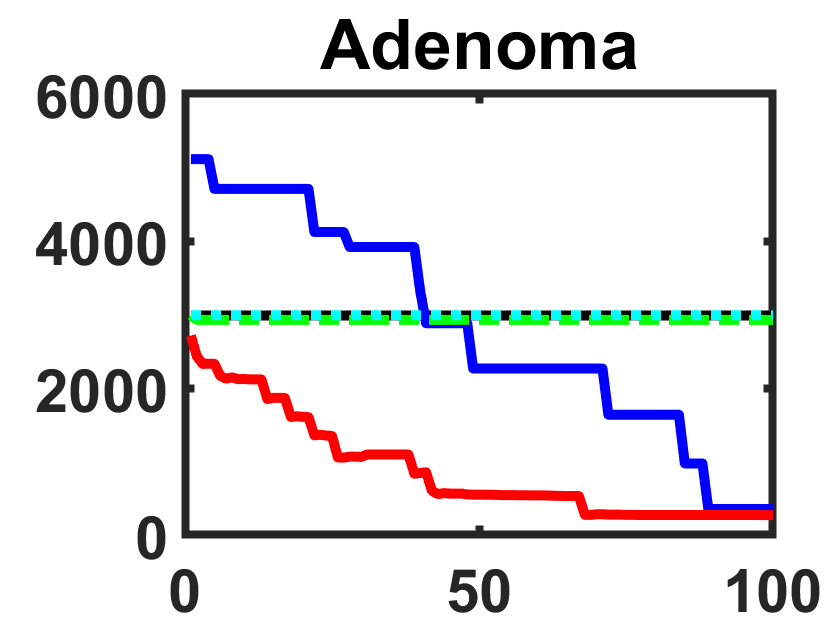}
  \end{minipage}
  \begin{minipage}[b]{0.1612\textwidth}
    \includegraphics[width=\textwidth]{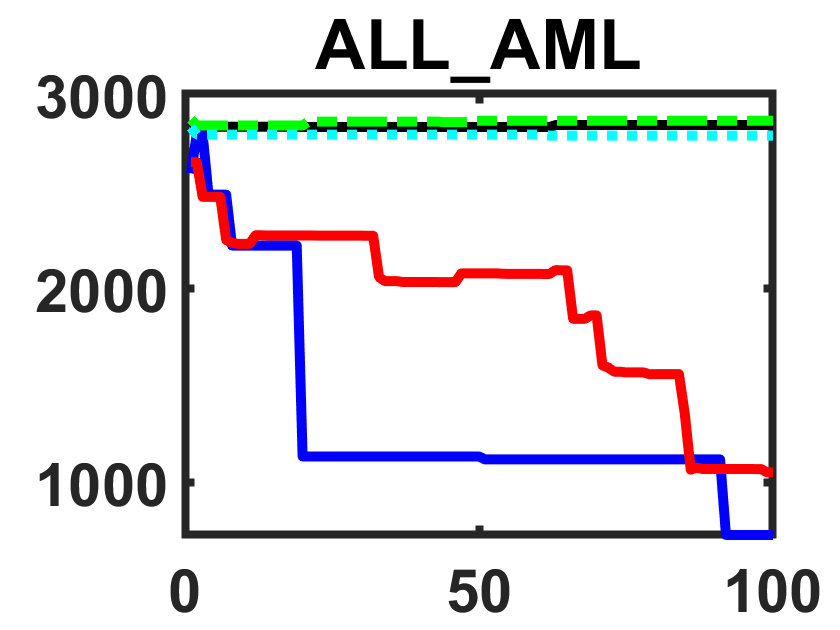}
  \end{minipage}
  \begin{minipage}[b]{0.1612\textwidth}
    \includegraphics[width=\textwidth]{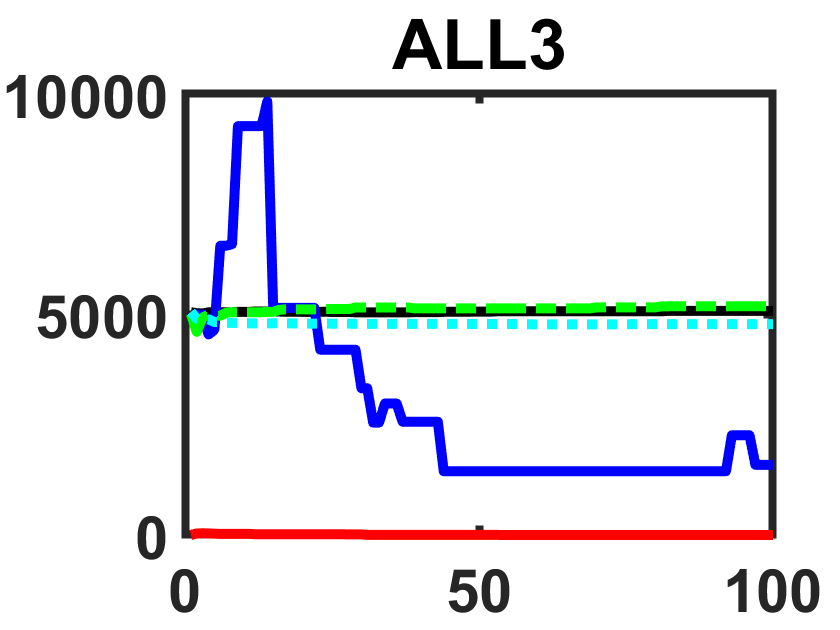}
  \end{minipage}
  \begin{minipage}[b]{0.1612\textwidth}
    \includegraphics[width=\textwidth]{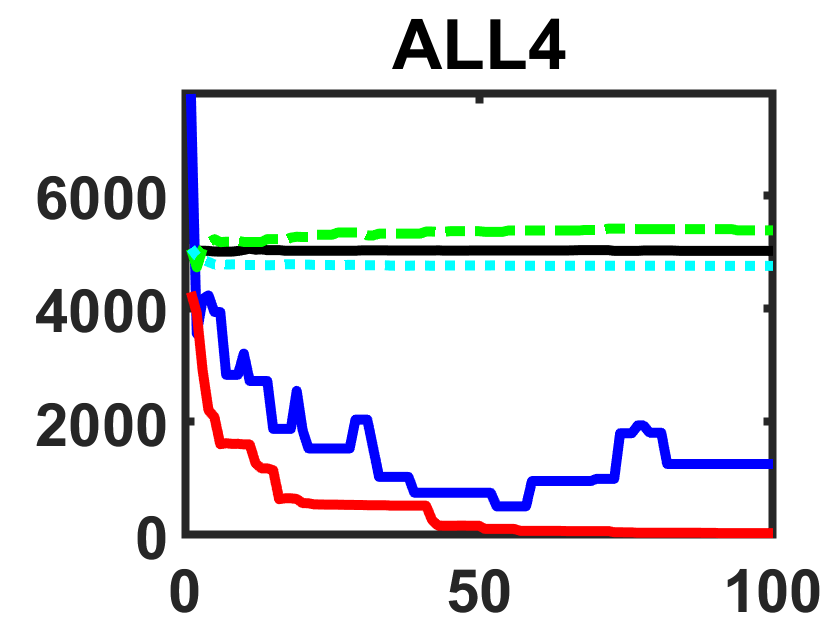}
  \end{minipage}
  \begin{minipage}[b]{0.1612\textwidth}
    \includegraphics[width=\textwidth]{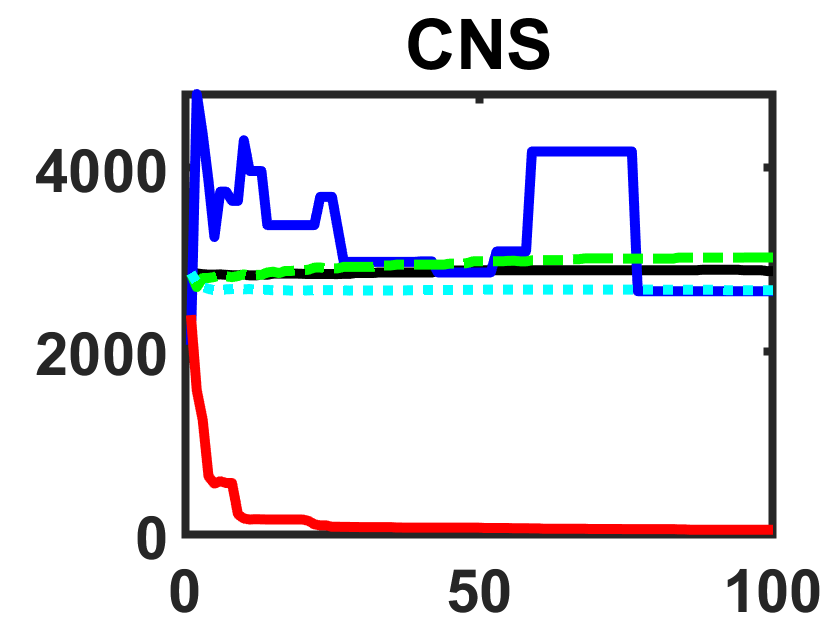}
  \end{minipage}
  \begin{minipage}[b]{0.1612\textwidth}
    \includegraphics[width=\textwidth]{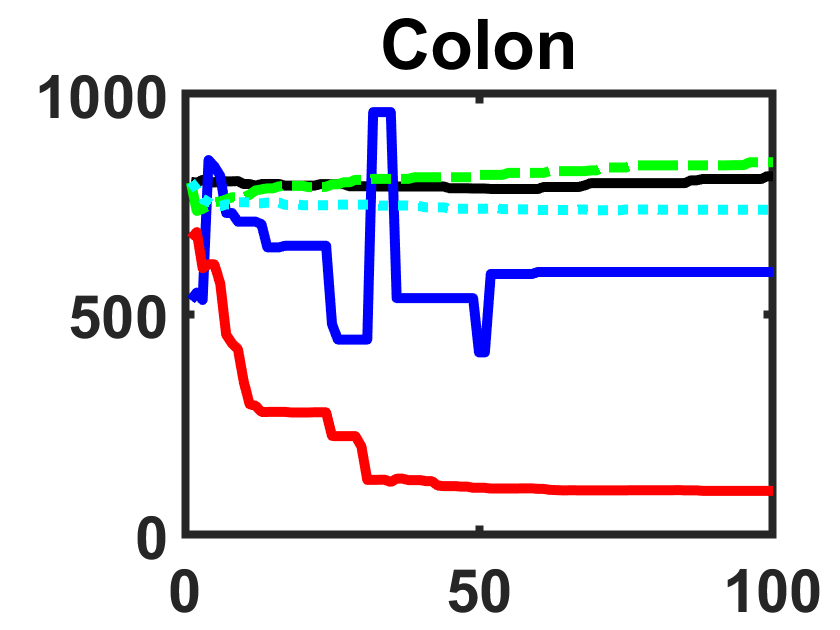}
  \end{minipage}
    \centering
  \begin{minipage}[b]{0.1612\textwidth}
    \includegraphics[width=\textwidth]{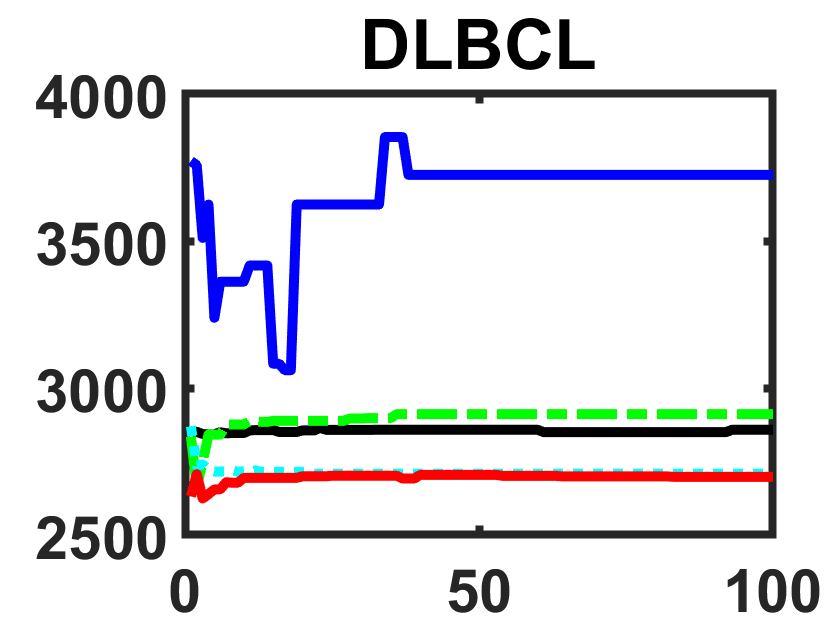}
  \end{minipage}
  \begin{minipage}[b]{0.1612\textwidth}
    \includegraphics[width=\textwidth]{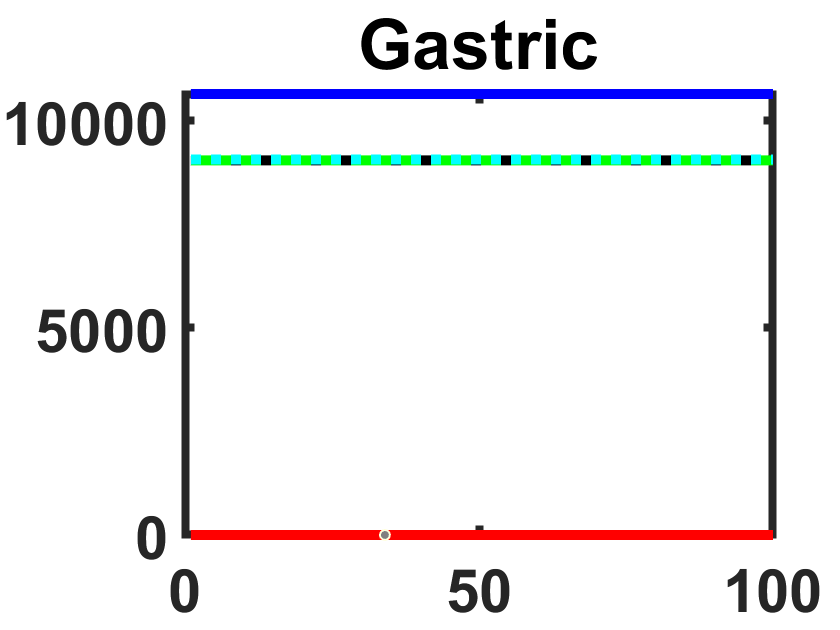}
  \end{minipage}
  \begin{minipage}[b]{0.1612\textwidth}
    \includegraphics[width=\textwidth]{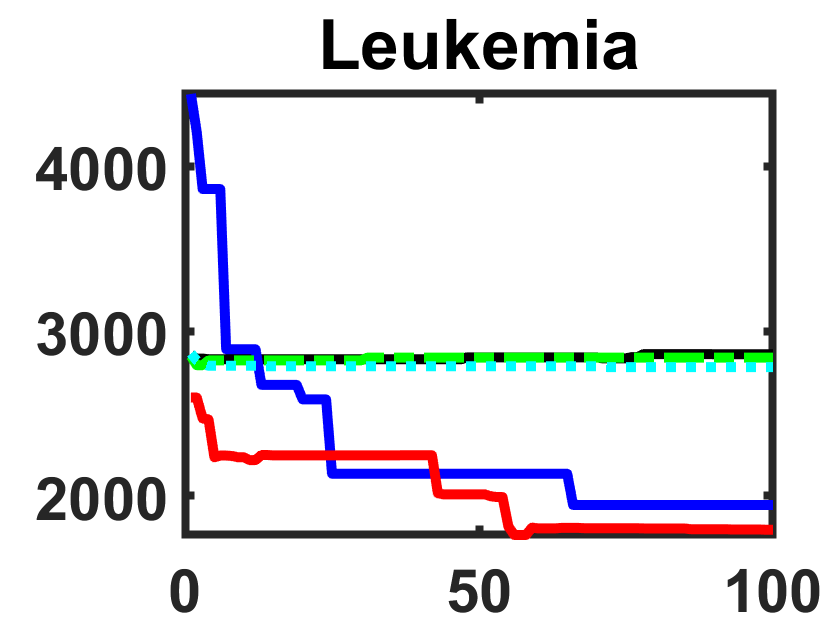}
  \end{minipage}
  \begin{minipage}[b]{0.1612\textwidth}
    \includegraphics[width=\textwidth]{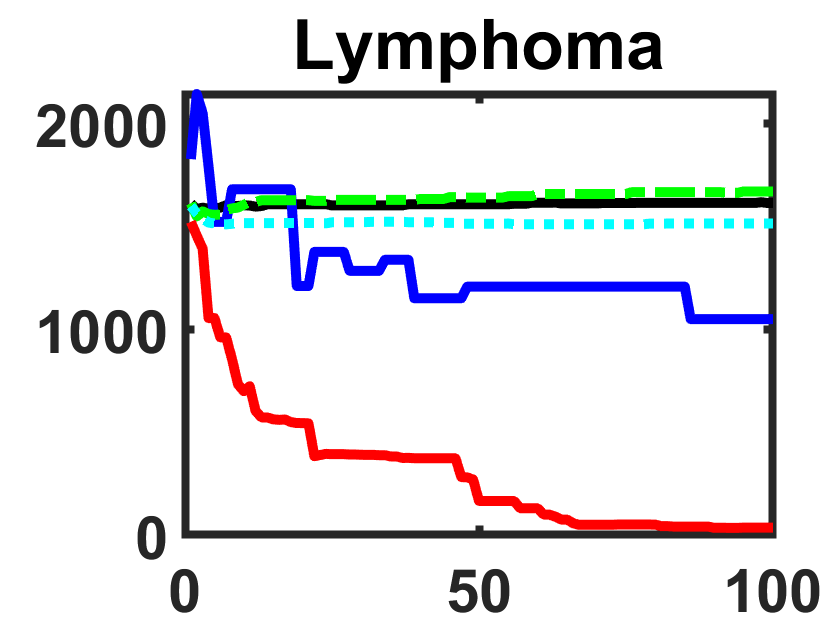}
  \end{minipage}
  \begin{minipage}[b]{0.1612\textwidth}
    \includegraphics[width=\textwidth]{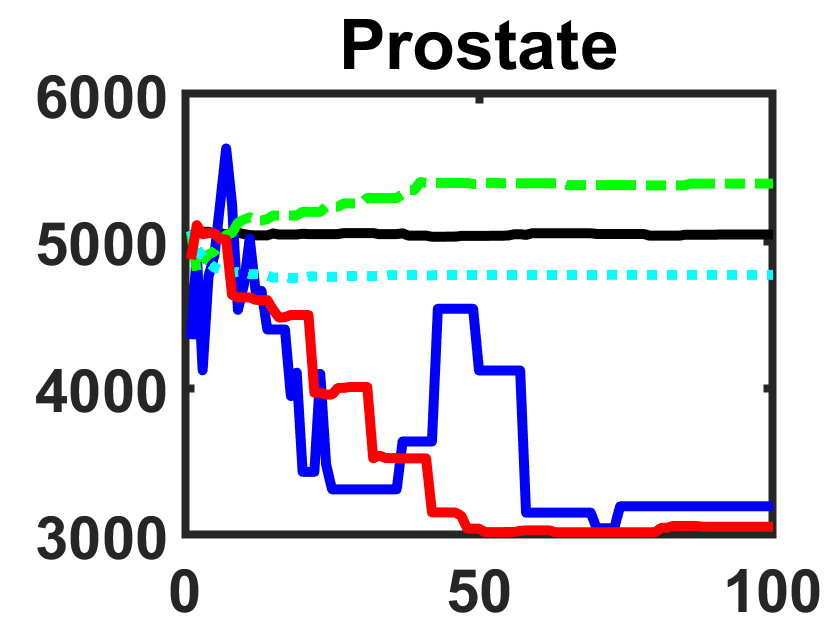}
  \end{minipage}
  \begin{minipage}[b]{0.1612\textwidth}
    \includegraphics[width=\textwidth]{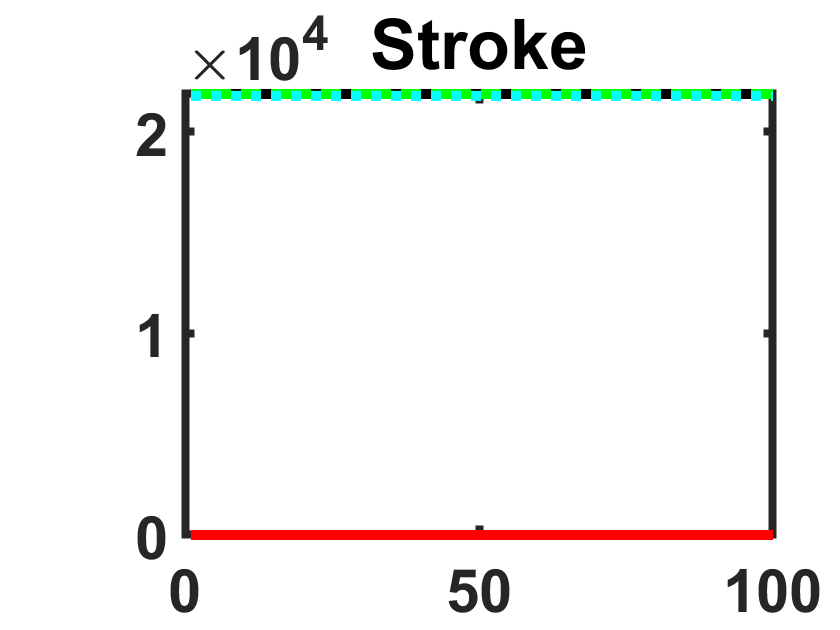}
  \end{minipage}
  \begin{minipage}[b]{0.18\textwidth}
    \includegraphics[width=\textwidth]{swarm/label.png}
  \end{minipage}
  \caption{Convergence Curves of Swarm-based EC Algorithms in Terms of the Size of Feature Subset}
\end{figure*}

\begin{figure*}[htp!]
  \centering
  \begin{minipage}[b]{0.1612\textwidth}
    \includegraphics[width=\textwidth]{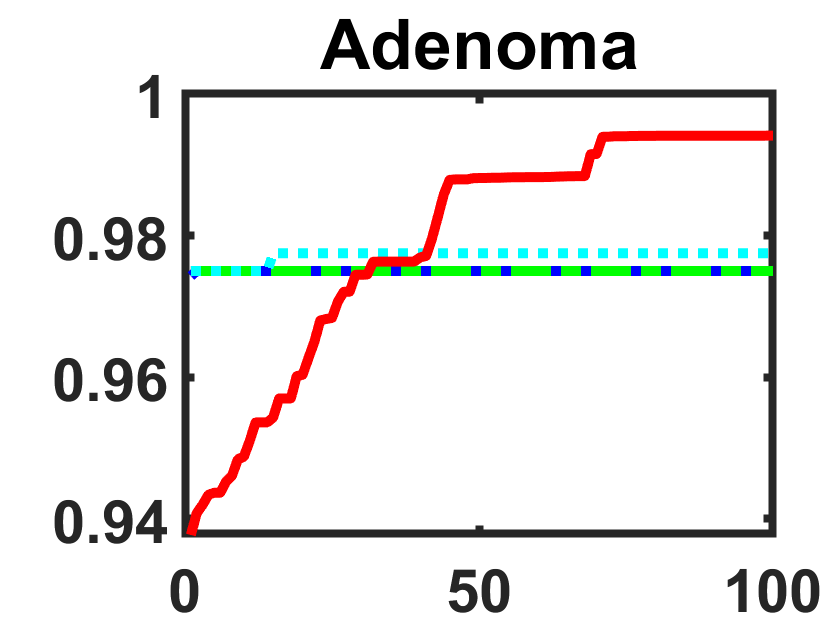}
  \end{minipage}
  \begin{minipage}[b]{0.1612\textwidth}
    \includegraphics[width=\textwidth]{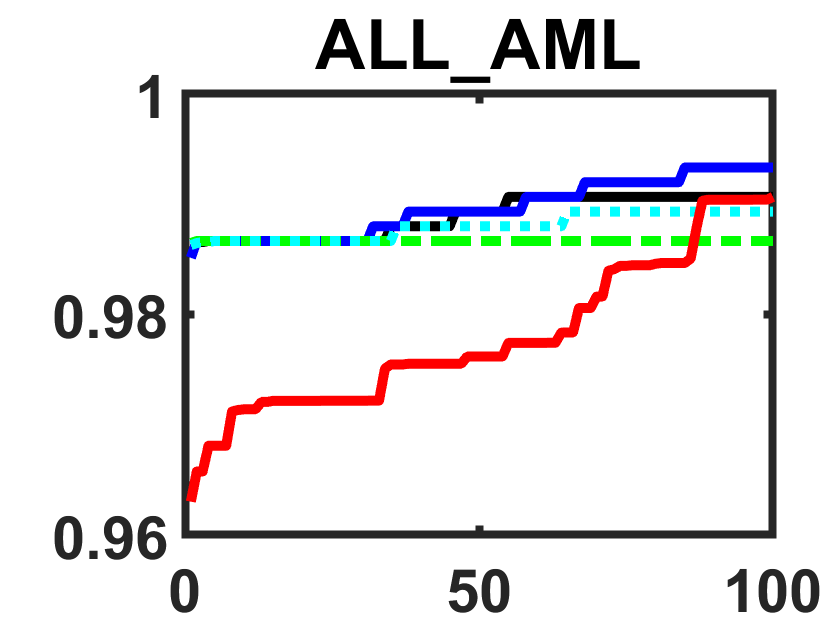}
  \end{minipage}
  \begin{minipage}[b]{0.1612\textwidth}
    \includegraphics[width=\textwidth]{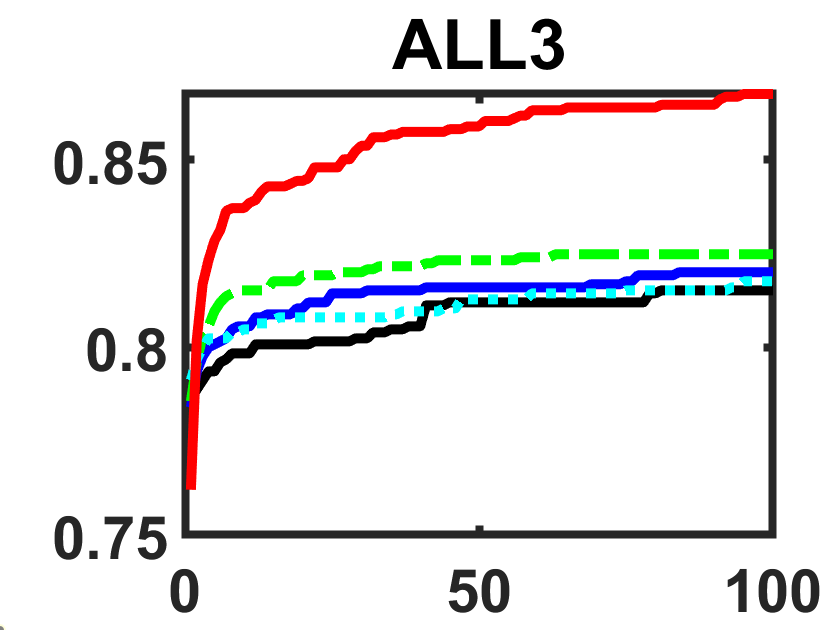}
  \end{minipage}
  \begin{minipage}[b]{0.1612\textwidth}
    \includegraphics[width=\textwidth]{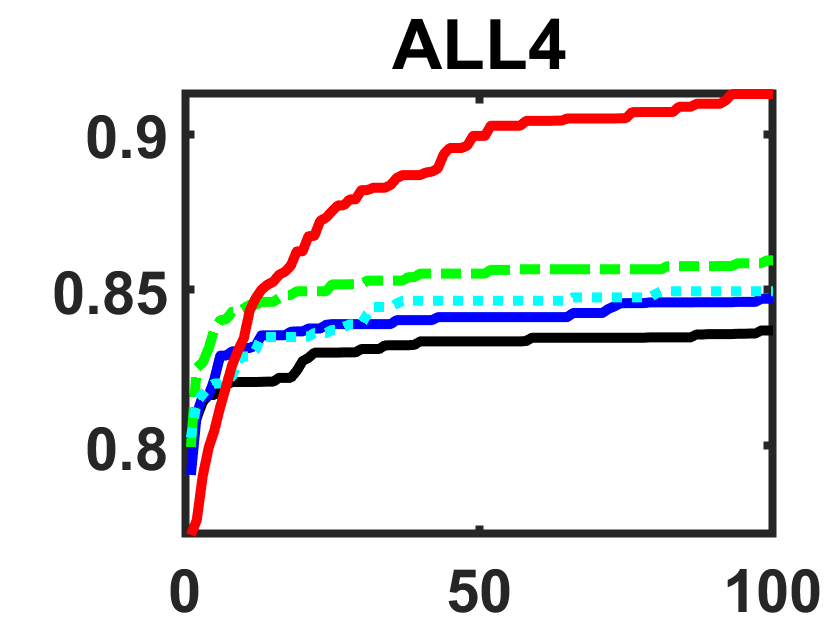}
  \end{minipage}
  \begin{minipage}[b]{0.1612\textwidth}
    \includegraphics[width=\textwidth]{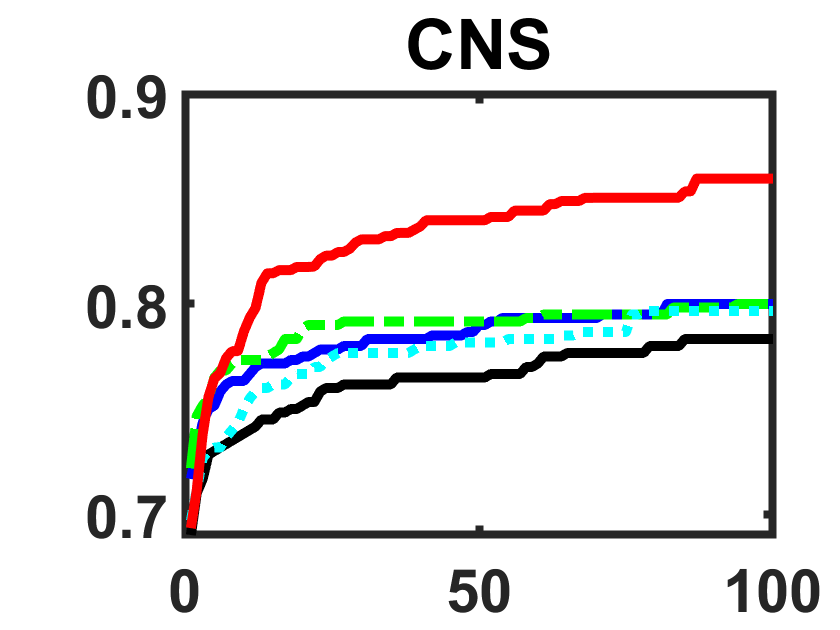}
  \end{minipage}
  \begin{minipage}[b]{0.1612\textwidth}
    \includegraphics[width=\textwidth]{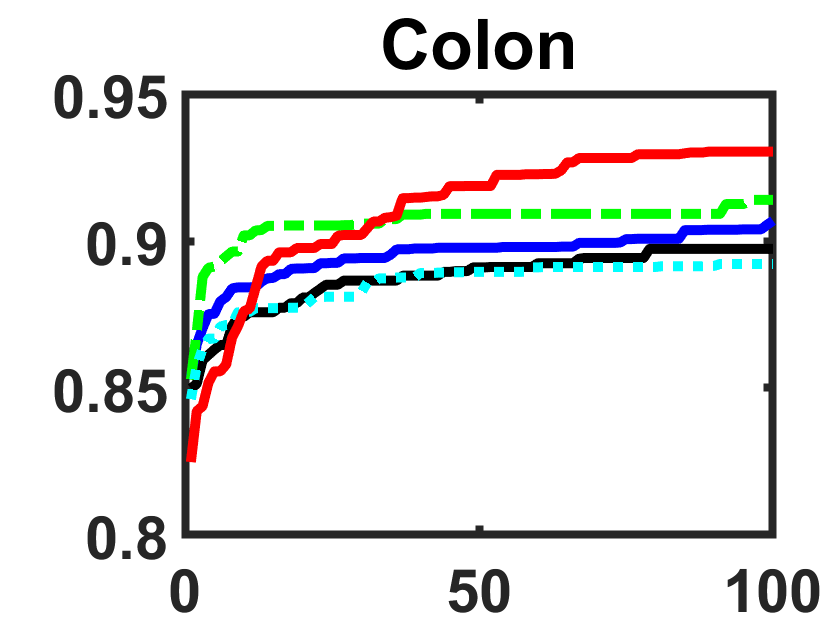}
  \end{minipage}
    \centering
  \begin{minipage}[b]{0.1612\textwidth}
    \includegraphics[width=\textwidth]{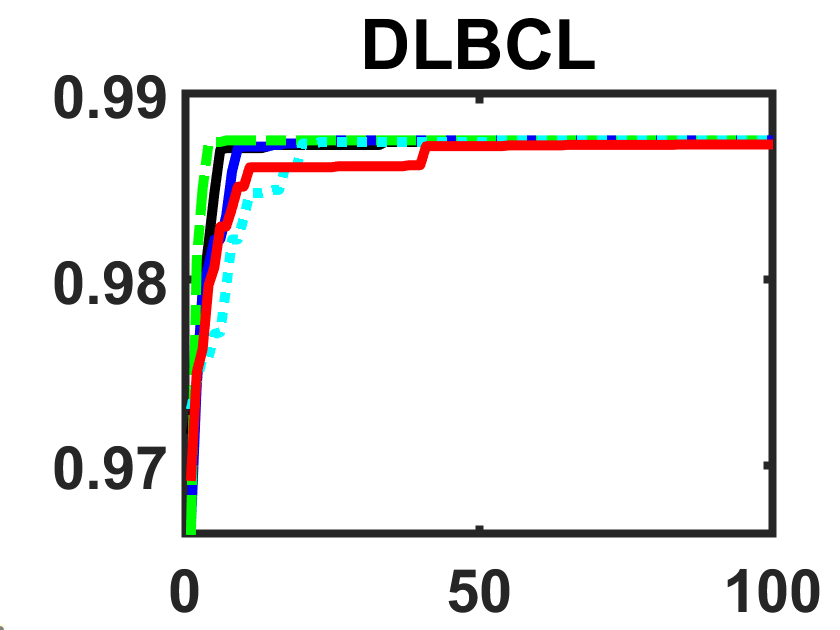}
  \end{minipage}
  \begin{minipage}[b]{0.1612\textwidth}
    \includegraphics[width=\textwidth]{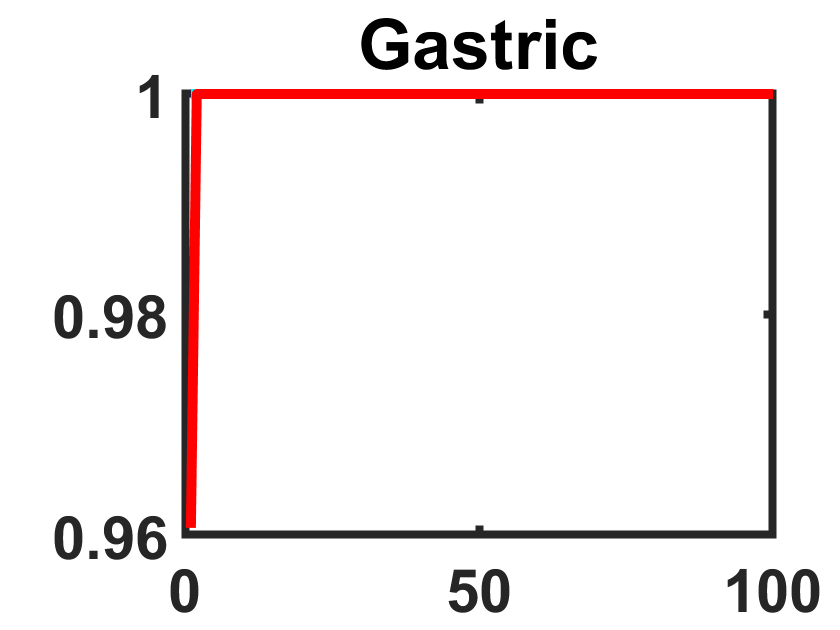}
  \end{minipage}
  \begin{minipage}[b]{0.1612\textwidth}
    \includegraphics[width=\textwidth]{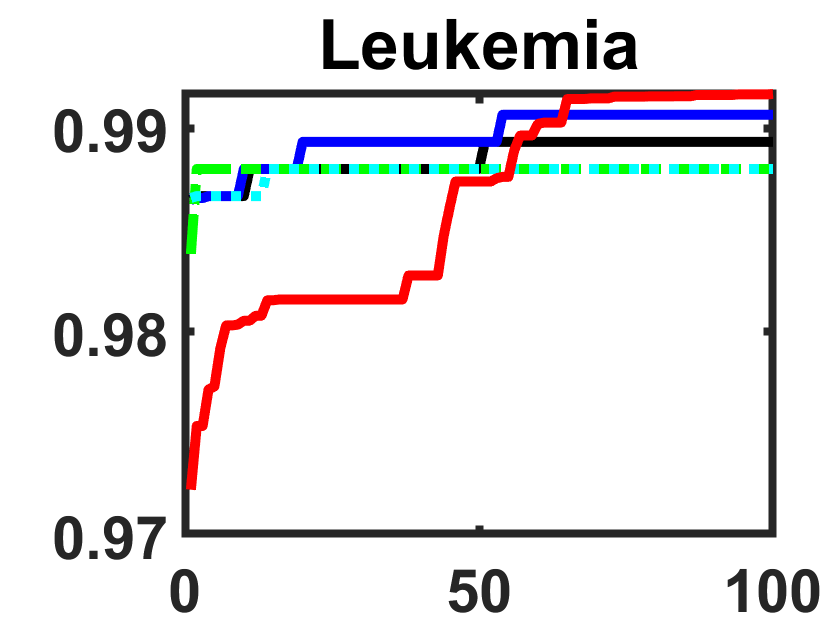}
  \end{minipage}
  \begin{minipage}[b]{0.1612\textwidth}
    \includegraphics[width=\textwidth]{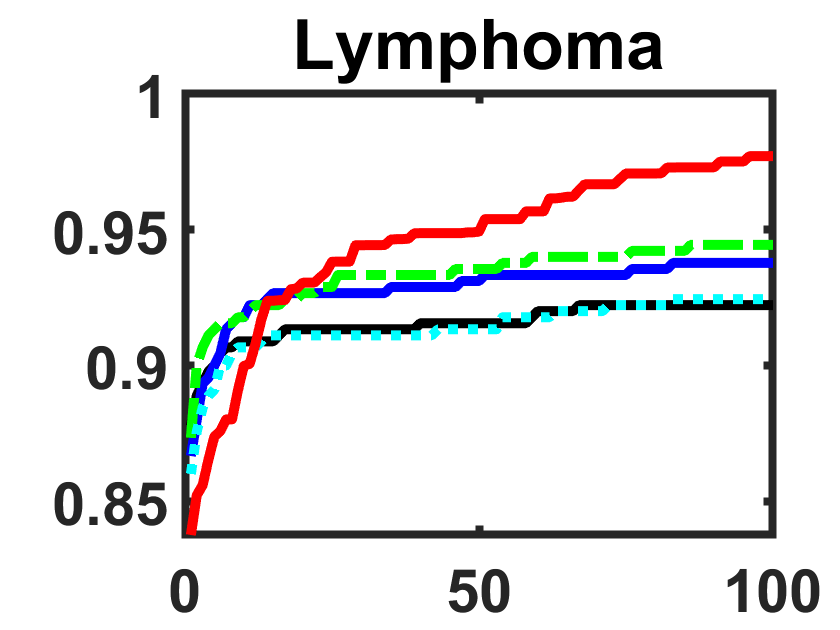}
  \end{minipage}
  \begin{minipage}[b]{0.1612\textwidth}
    \includegraphics[width=\textwidth]{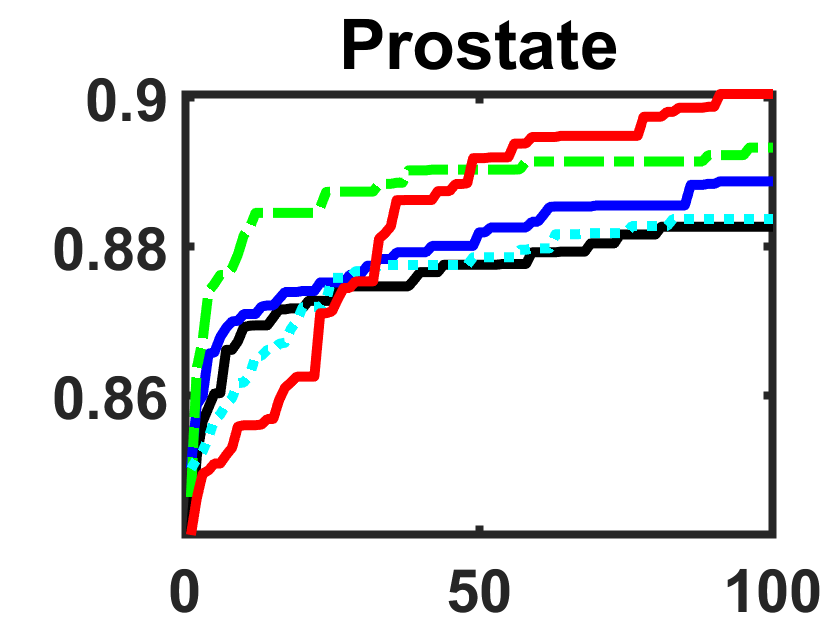}
  \end{minipage}
  \begin{minipage}[b]{0.1612\textwidth}
    \includegraphics[width=\textwidth]{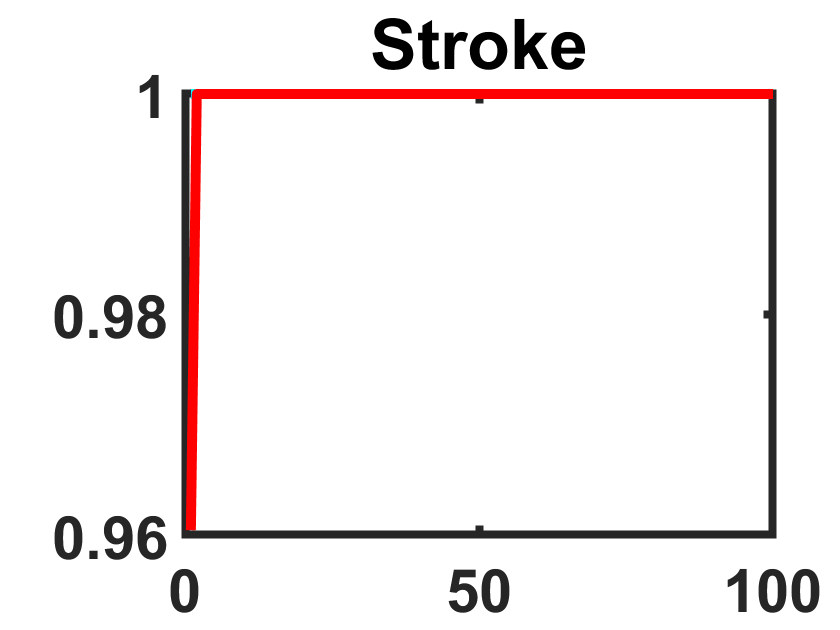}
  \end{minipage}
  \begin{minipage}[b]{0.17\textwidth}
    \includegraphics[width=\textwidth]{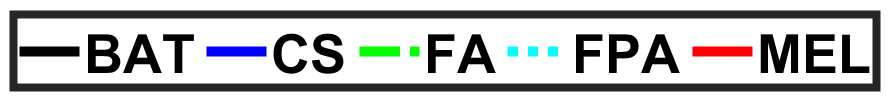}
  \end{minipage}
  \caption{Convergence Curves of Nature-inspired EC Algorithms in Terms of Accuracy}
\end{figure*}

\begin{figure*}[htp!]
  \centering
  \begin{minipage}[b]{0.1612\textwidth}
    \includegraphics[width=\textwidth]{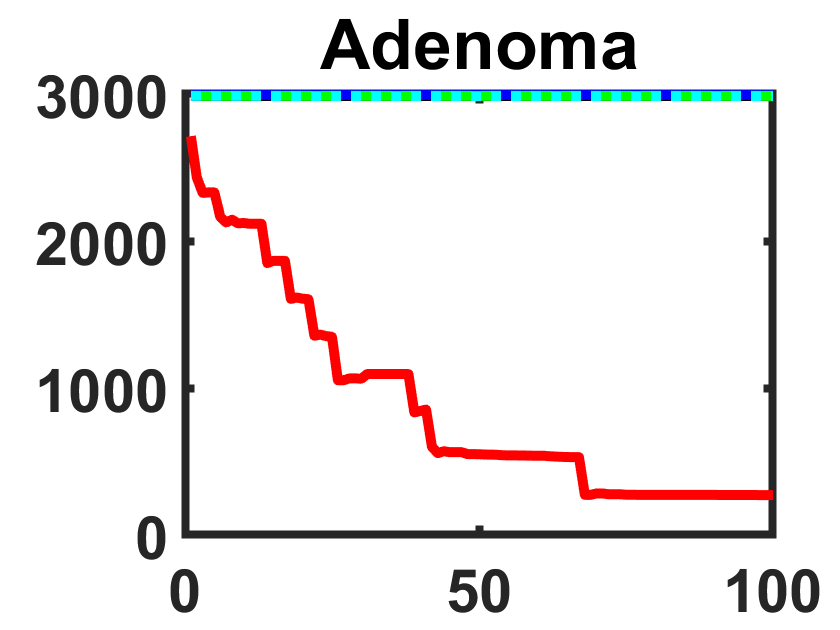}
  \end{minipage}
  \begin{minipage}[b]{0.1612\textwidth}
    \includegraphics[width=\textwidth]{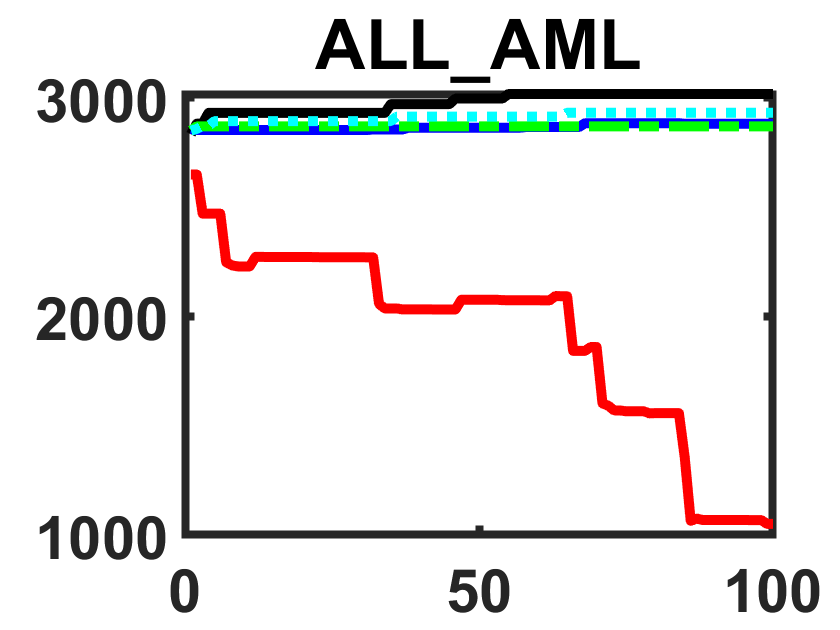}
  \end{minipage}
  \begin{minipage}[b]{0.1612\textwidth}
    \includegraphics[width=\textwidth]{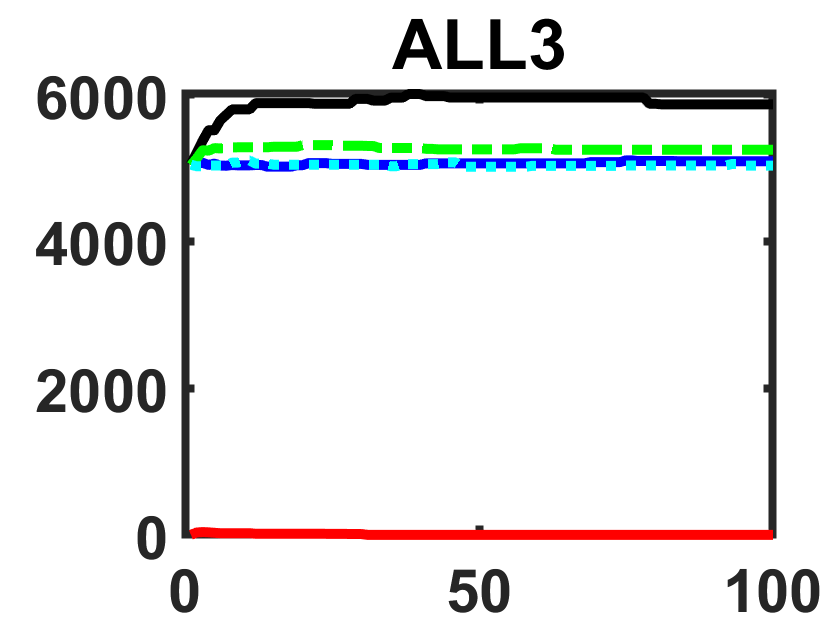}
  \end{minipage}
  \begin{minipage}[b]{0.1612\textwidth}
    \includegraphics[width=\textwidth]{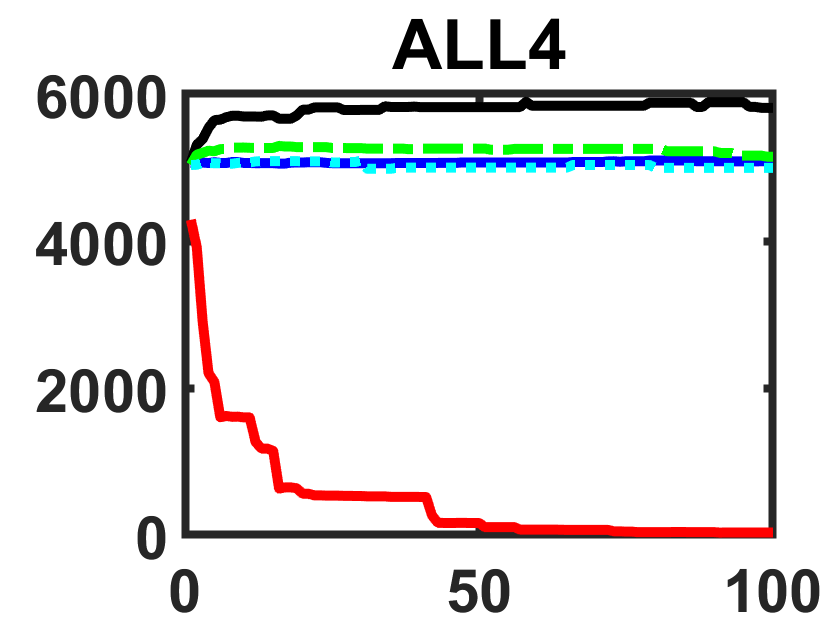}
  \end{minipage}
  \begin{minipage}[b]{0.1612\textwidth}
    \includegraphics[width=\textwidth]{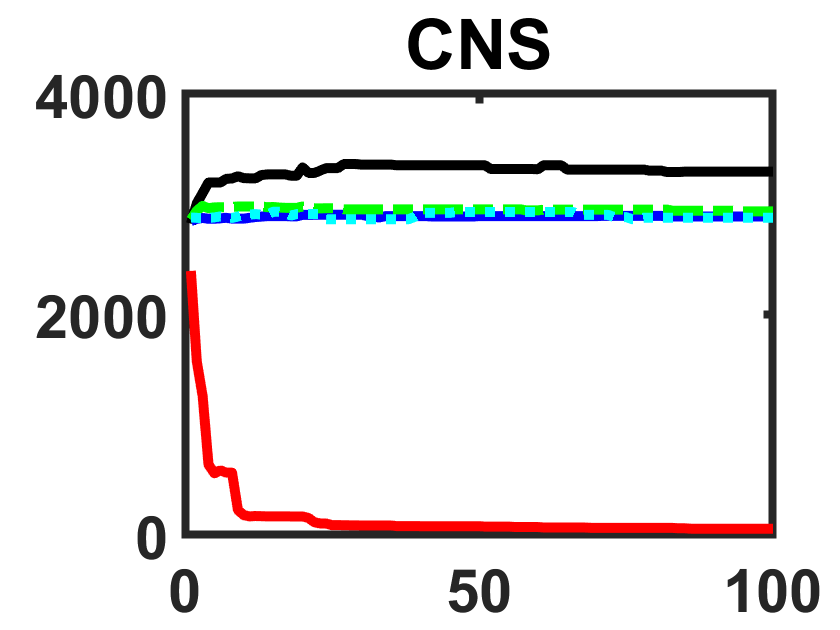}
  \end{minipage}
  \begin{minipage}[b]{0.1612\textwidth}
    \includegraphics[width=\textwidth]{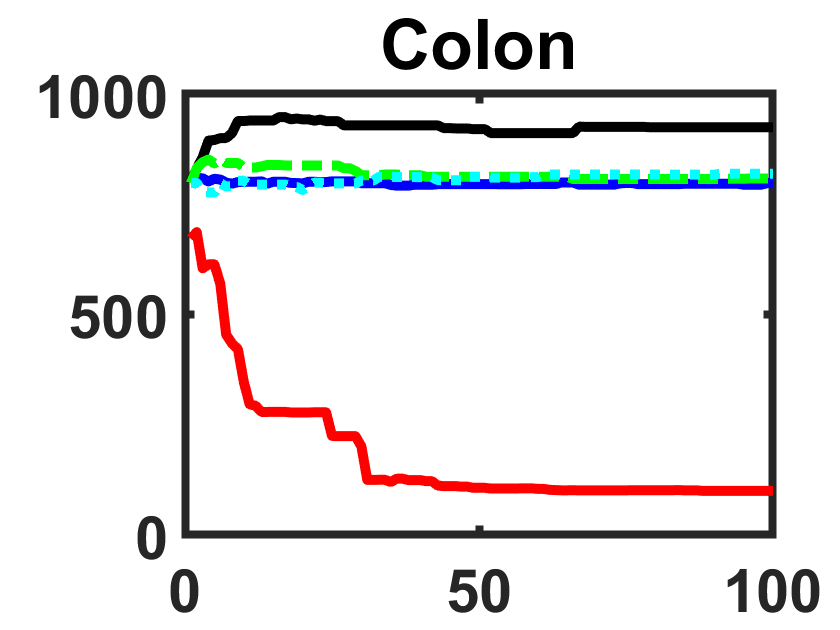}
  \end{minipage}
    \centering
  \begin{minipage}[b]{0.1612\textwidth}
    \includegraphics[width=\textwidth]{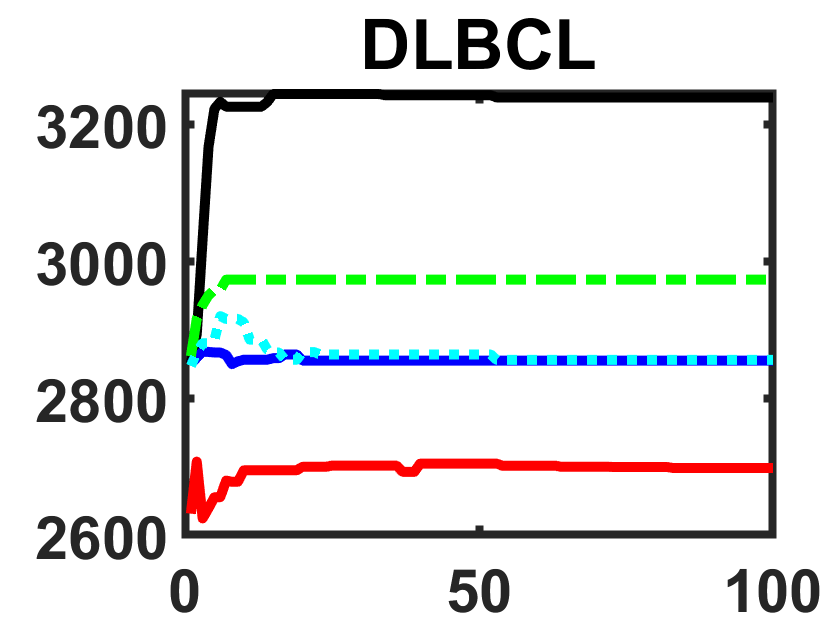}
  \end{minipage}
  \begin{minipage}[b]{0.1612\textwidth}
    \includegraphics[width=\textwidth]{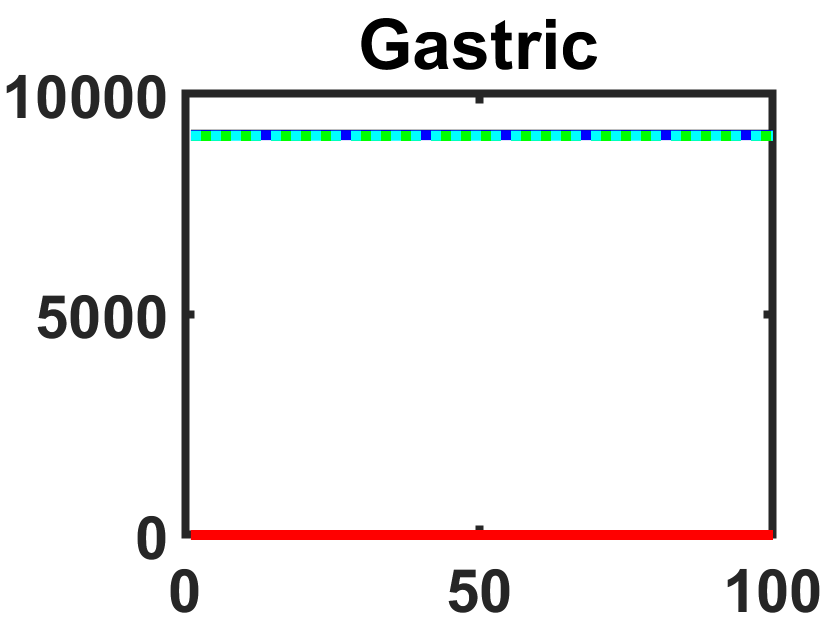}
  \end{minipage}
  \begin{minipage}[b]{0.1612\textwidth}
    \includegraphics[width=\textwidth]{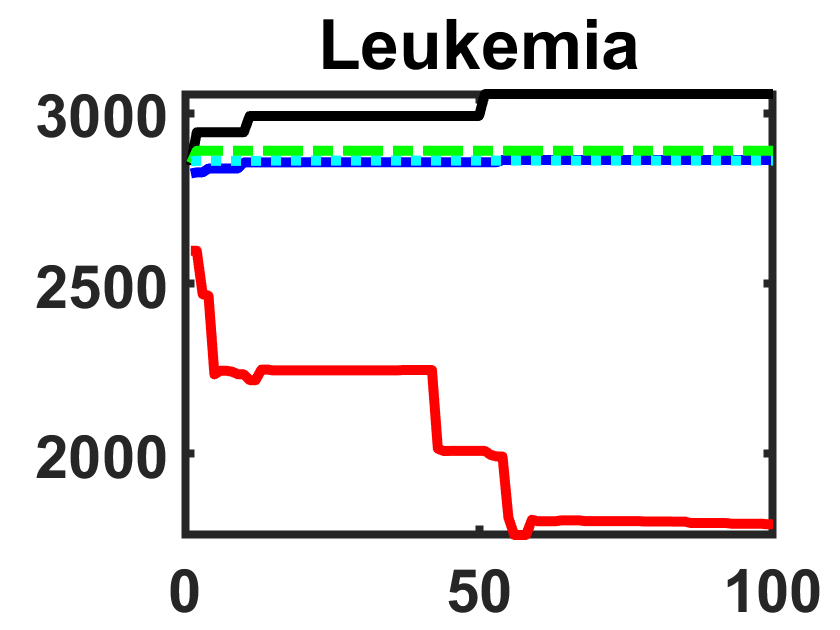}
  \end{minipage}
  \begin{minipage}[b]{0.1612\textwidth}
    \includegraphics[width=\textwidth]{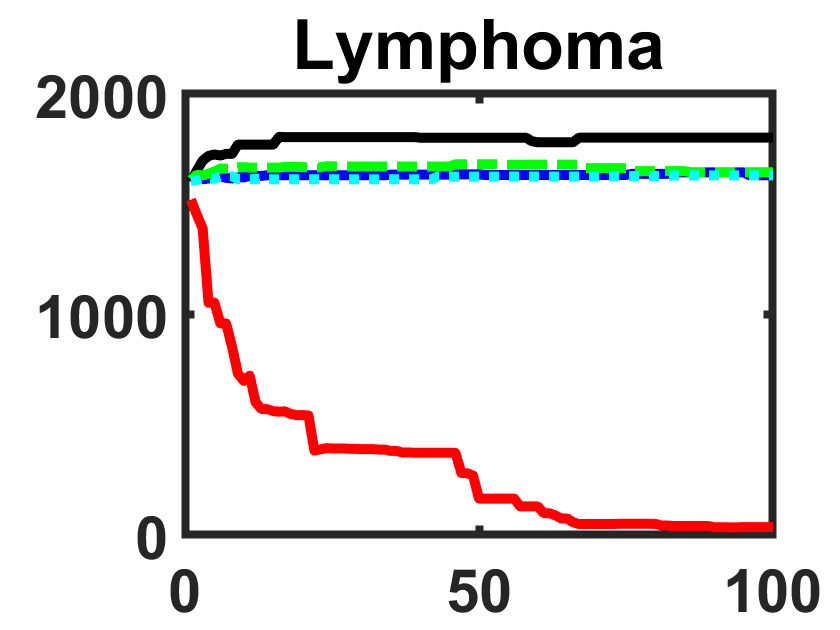}
  \end{minipage}
  \begin{minipage}[b]{0.1612\textwidth}
    \includegraphics[width=\textwidth]{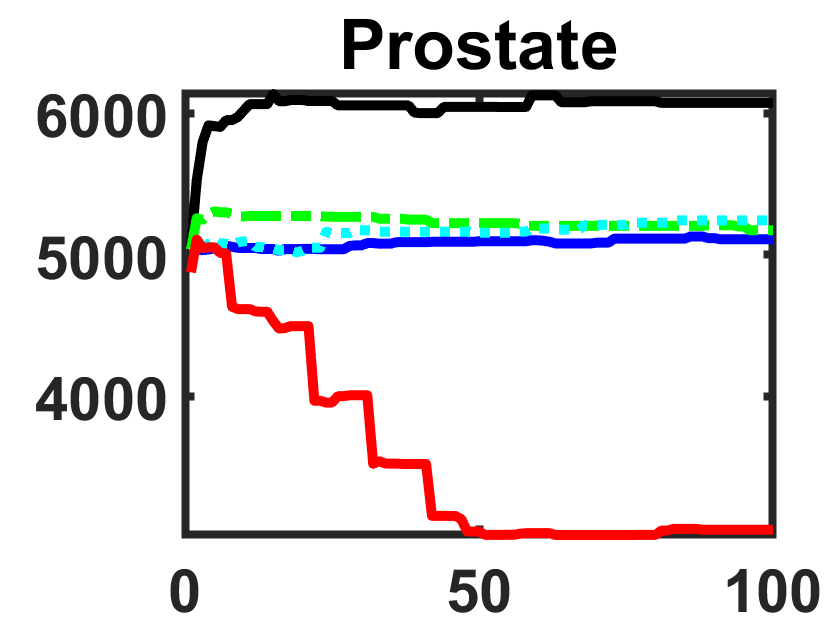}
  \end{minipage}
  \begin{minipage}[b]{0.1612\textwidth}
    \includegraphics[width=\textwidth]{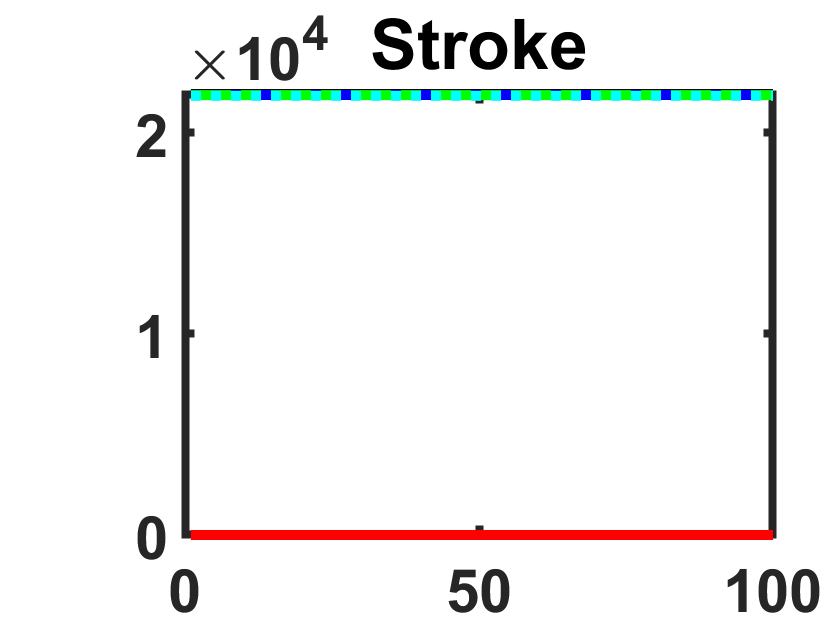}
  \end{minipage}
  \begin{minipage}[b]{0.17\textwidth}
    \includegraphics[width=\textwidth]{nature_inspired/label.png}
  \end{minipage}
  \caption{Convergence Curves of Nature-inspired EC Algorithms in Terms of the Size of Feature Subset}
\end{figure*}

\subsection{Comparison with Nature-inspired Heuristic Methods}
TABLE VI presents the classification accuracy results of MEL and four other nature-inspired heuristic methods. It is evident that MEL achieves the highest classification accuracy on 9 out of 12 datasets and demonstrates the best overall performance. Additionally, TABLE VII highlights that the average feature subset size obtained by MEL is approximately 1/7 of the sizes obtained by the other nature-inspired methods, with the latter all exceeding 5000. Figures 4 and 5 display the convergence curves of these algorithms in terms of classification accuracy and subset size, respectively. From Figure 4, it can be observed that although MEL did not initially achieve the best performance, it quickly outperformed the other methods on most datasets. MEL exhibits slower convergence but demonstrates stronger global search ability. Figure 5 illustrates that while the curves of the other methods stabilize quickly, MEL effectively searches for the optimal feature subset throughout most of the entire cycle.
 
\begin{table}[!htb]
\centering
\resizebox{87mm}{!}{
\begin{tabular}{ l l l l l l }
\hline
\multirow{2}{*}{\textbf{Dataset}} & \textbf{BAT}	&	\textbf{CS}	&	\textbf{FA}	&	\textbf{FPA}	&	\textbf{MEL (Ours)} \\ 
 \cline{2-6}
   \textbf{ } & \textbf{Mean $\pm$ Std} & \textbf{Mean $\pm$ Std}  & \textbf{Mean $\pm$ Std} & \textbf{Mean $\pm$ Std}  & \textbf{Mean $\pm$ Std} \\
\hline
Adenoma	&	0.9750	±	0.0000	&	0.9750	±	0.0000	&	0.9750	±	0.0000	&	0.9775	±	0.0079	&	\textbf{0.9975}	±	0.0079	\\ \hline
ALL$\_$AML	&	0.9907	±	0.0064	&	\textbf{0.9933}	±	0.0070	&	0.9867	±	0.0000	&	0.9893	±	0.0056	&	0.9907	±	0.0064	\\ \hline
ALL3	&	0.8152	±	0.0116	&	0.8200	±	0.0094	&	0.8248	±	0.0059	&	0.8176	±	0.0074	&	\textbf{0.8528}	±	0.0137	\\ \hline
ALL4	&	0.8370	±	0.0105	&	0.8471	±	0.0069	&	0.8595	±	0.0093	&	0.8495	±	0.0101	&	\textbf{0.9035}	±	0.0219	\\ \hline
CNS	&	0.7833	±	0.0222	&	0.8000	±	0.0192	&	0.8000	±	0.0136	&	0.7967	±	0.0070	&	\textbf{0.8450}	±	0.0223	\\ \hline
Colon	&	0.8973	±	0.0112	&	0.9069	±	0.0062	&	0.9140	±	0.0110	&	0.8922	±	0.0065	&	\textbf{0.9282}	±	0.0136	\\ \hline
DLBCL	&	\textbf{0.9875}	±	0.0000	&	\textbf{0.9875}	±	0.0000	&	\textbf{0.9875}	±	0.0000	&	0.8691	±	0.0000	&	0.9873	±	0.0004	\\ \hline
Gastric	&	\textbf{1.0000}	±	0.0000	&	\textbf{1.0000}	±	0.0000	&	\textbf{1.0000}	±	0.0000	&	\textbf{1.0000}	±	0.0000	&	\textbf{1.0000}	±	0.0000	\\ \hline
Leukemia	&	0.9893	±	0.0056	&	0.9907	±	0.0064	&	0.9880	±	0.0042	&	0.9880	±	0.0042	&	\textbf{0.9917}	±	0.0071	\\ \hline
Lymphoma	&	0.9222	±	0.0157	&	0.9378	±	0.0141	&	0.9444	±	0.0240	&	0.9244	±	0.0115	&	\textbf{0.9756}	±	0.0164	\\ \hline
Prostate	&	0.8826	±	0.0045	&	0.8887	±	0.0099	&	\textbf{0.9305}	±	0.0061	&	0.8837	±	0.0084	&	0.9004	±	0.0259	\\ \hline
Stroke	&	\textbf{1.0000}	±	0.0000	&	\textbf{1.0000}	±	0.0000	&	\textbf{1.0000}	±	0.0000	&	\textbf{1.0000}	±	0.0000	&	\textbf{1.0000}	±	0.0000	\\ \hline
\textbf{Average}	&	0.9233	±	0.0073	&	0.9289	±	0.0066	&	0.9342	±	0.0062	&	0.9157	±	0.0057	&	\textbf{0.9477}	±	0.0113	\\ \hline
\end{tabular}}
\caption{Comparison with Nature-inspired Heuristic Methods on Accuracy}
\end{table}

\begin{table}[!htb]
\centering
\resizebox{87mm}{!}{
\begin{tabular}{ l l l l l l }
\hline
\multirow{2}{*}{\textbf{Dataset}} & \textbf{BAT}	&	\textbf{CS}	&	\textbf{FA}	&	\textbf{FPA}	&	\textbf{MEL (Ours)} \\ 
 \cline{2-6}
   \textbf{ } & \textbf{Mean $\pm$ Std} & \textbf{Mean $\pm$ Std}  & \textbf{Mean $\pm$ Std} & \textbf{Mean $\pm$ Std}  & \textbf{Mean $\pm$ Std} \\
\hline
Adenoma	&	2986.4	±	23.4	&	2994.0	±	57.6	&	2987.1	±	41.6	&	2985.5	±	36.6	&	\textbf{271.1}	±	844.3	\\ \hline
ALL$\_$AML	&	3016.9	±	256.3	&	2880.6	±	69.5	&	2868.9	±	48.6	&	2931.2	±	91.6	&	\textbf{1050.8}	±	1281.9	\\ \hline
ALL3	&	5868.0	±	307.3	&	5092.1	±	105.6	&	5249.9	±	72.7	&	5032.7	±	189.2	&	\textbf{1.0}	±	0.0	\\ \hline
ALL4	&	5811.1	±	234.5	&	5085.1	±	83.7	&	5146.8	±	90.5	&	4995.7	±	319.7	&	\textbf{34.1}	±	24.9	\\ \hline
CNS	&	3295.9	±	206.7	&	2889.3	±	54.8	&	2935.0	±	61.1	&	2877.9	±	188.8	&	\textbf{57.2}	±	74.0	\\ \hline
Colon	&	924.6	±	53.2	&	798.0	±	20.6	&	808.3	±	20.5	&	819.1	±	50.7	&	\textbf{100.1}	±	266.1	\\ \hline
DLBCL	&	3239.0	±	212.9	&	2854.6	±	49.2	&	2973.0	±	62.6	&	2855.7	±	61.4	&	\textbf{2697.8}	±	146.6	\\ \hline
Gastric	&	9076.9	±	68.7	&	9067.8	±	84.1	&	9047.2	±	72.9	&	9053.5	±	53.4	&	\textbf{1.0}	±	0.0	\\ \hline
Leukemia	&	3056.9	±	269.1	&	2863.2	±	59.5	&	2890.4	±	79.7	&	2861.1	±	36.1	&	\textbf{1793.4}	±	1236.5	\\ \hline
Lymphoma	&	1802.1	±	158.0	&	1631.3	±	39.6	&	1645.1	±	41.7	&	1630.7	±	41.9	&	\textbf{35.6}	±	46.3	\\ \hline
Prostate	&	6074.9	±	262.7	&	5110.3	±	61.5	&	3729.4	±	25.9	&	5244.2	±	175.5	&	\textbf{3055.8}	±	2579.6	\\ \hline
Stroke	&	21933.6	±	140.0	&	21882.3	±	139.5	&	21874.8	±	144.5	&	21838.3	±	106.3	&	\textbf{1.0}	±	0.0	\\ \hline
\textbf{Average}	&	5590.5	±	182.7	&	5262.4	±	68.8	&	5179.7	±	63.5	&	5260.5	±	112.6	&	\textbf{758.2}	±	541.7	\\ \hline
\end{tabular}}
\caption{Comparison with Nature-inspired Heuristic Methods on the Subset Size}
\end{table}

\begin{table}[!htb]
\centering
\resizebox{87mm}{!}{
\begin{tabular}{ l l l l l l }
\hline
\multirow{2}{*}{\textbf{Dataset}} & \textbf{BAT}	&	\textbf{CS}	&	\textbf{FA}	&	\textbf{FPA}	&	\textbf{MEL (Ours)} \\ 
 \cline{2-6}
   \textbf{ } & \textbf{Mean $\pm$ Std} & \textbf{Mean $\pm$ Std}  & \textbf{Mean $\pm$ Std} & \textbf{Mean $\pm$ Std}  & \textbf{Mean $\pm$ Std} \\
\hline
Adenoma	&	82.0	±	3.4	&	137.1	±	0.9	&	\textbf{8.0}	±	1.4	&	67.4	±	2.1	&	55.5	±	1.9	\\ \hline
ALL$\_$AML	&	101.7	±	2.6	&	165.4	±	4.9	&	\textbf{13.8}	±	3.4	&	90.9	±	0.9	&	66.2	±	5.8	\\ \hline
ALL3	&	219.1	±	4.3	&	351.3	±	2.2	&	1689.7	±	51.7	&	185.9	±	1.3	&	\textbf{62.9}	±	2.3	\\ \hline
ALL4	&	171.8	±	2.8	&	266.5	±	2.0	&	1426.9	±	12.9	&	141.6	±	0.9	&	\textbf{66.3}	±	5.6	\\ \hline
CNS	&	93.7	±	1.8	&	148.7	±	0.6	&	724.8	±	11.8	&	81.8	±	0.5	&	\textbf{53.1}	±	0.9	\\ \hline
Colon	&	59.5	±	1.4	&	103.5	±	1.0	&	536.1	±	5.8	&	59.4	±	0.5	&	\textbf{44.9}	±	1.0	\\ \hline
DLBCL	&	105.1	±	1.9	&	169.9	±	3.5	&	304.4	±	38.5	&	111.9	±	0.4	&	\textbf{72.3}	±	0.8	\\ \hline
Gastric	&	199.8	±	4.5	&	308.7	±	7.0	&	\textbf{1.7}	±	0.1	&	161.3	±	1.1	&	74.4	±	1.0	\\ \hline
Leukemia	&	101.4	±	2.5	&	163.8	±	0.9	&	80.6	±	216.6	&	90.5	±	0.8	&	\textbf{66.8}	±	6.1	\\ \hline
Lymphoma	&	65.8	±	1.2	&	111.1	±	1.3	&	441.2	±	151.9	&	63.0	±	0.7	&	\textbf{48.2}	±	1.3	\\ \hline
Prostate	&	186.9	±	2.2	&	298.7	±	7.0	&	579.9	±	13.2	&	153.5	±	0.5	&	\textbf{98.0}	±	19.0	\\ \hline
Stroke	&	295.2	±	6.0	&	370.3	±	1.7	&	\textbf{2.3}	±	0.1	&	199.7	±	1.8	&	115.7	±	1.3	\\ \hline
\textbf{Average}	&	140.2	±	2.9	&	216.2	±	2.7	&	484.1	±	42.3	&	117.2	±	0.9	&	\textbf{68.7}	±	3.9	\\ \hline
\end{tabular}}
\caption{Comparison with Nature-inspired Heuristic Methods on Running Time}
\end{table}

In Table VIII, we conducted extensive experiments by comparing the running times of the nature-inspired heuristic methods. For the Adenoma, ALL\_AML, Gastric, and Stroke datasets, the FA algorithm achieved running times of 8.0 ± 1.4, 13.8 ± 3.4, 1.7 ± 0.1, and 2.3 ± 0.1, respectively, which are smaller than the other four algorithms. However, considering the algorithm's generality, our proposed algorithm outperforms the other four algorithms on 8 out of 12 datasets while maintaining a lower average running time.

\subsection{Comparison with Evolutionary Algorithms}
We also evaluated classic evolutionary algorithms, including differential evolution (DE) and genetic algorithm (GA). For GA, we considered two mechanisms: Roulette and Tournament. Extensive experiments were conducted, and the results are presented in Table IX, which compares the accuracy achieved by MEL with that of evolutionary algorithms. Remarkably, MEL consistently achieves the highest classification accuracy on nearly all datasets. Furthermore, the results of subset sizes (TABLE X) and running times (TABLE XI) further demonstrate the distinct advantage of MEL over these evolutionary algorithms.

\begin{table}[!htb]
\centering
\resizebox{87mm}{!}{
\begin{tabular}{ l l l l l }
\hline
\multirow{2}{*}{\textbf{Dataset}} & \textbf{DE}	&	\textbf{GA (Roulette)}	&	\textbf{GA (Tournament)} &	\textbf{MEL (Ours)} \\ 
 \cline{2-5}
   \textbf{ } & \textbf{Mean $\pm$ Std} & \textbf{Mean $\pm$ Std}  & \textbf{Mean $\pm$ Std} & \textbf{Mean $\pm$ Std} \\
\hline
Adenoma	&	0.9750	±	0.0000	&	0.9750	±	0.0000	&	0.9750	±	0.0000	&	\textbf{0.9975}	±	0.0079	\\ \hline
ALL$\_$AML	&	0.9879	±	0.0040	&	0.9867	±	0.0000	&	0.9893	±	0.0056	&	\textbf{0.9907}	±	0.0064	\\ \hline
ALL3	&	0.8136	±	0.0039	&	0.8224	±	0.0051	&	0.8264	±	0.0066	&	\textbf{0.8528}	±	0.0137	\\ \hline
ALL4	&	0.8433	±	0.0083	&	0.8502	±	0.0068	&	0.8517	±	0.0111	&	\textbf{0.9035}	±	0.0219	\\ \hline
CNS	&	0.7900	±	0.0141	&	0.8350	±	0.0200	&	0.8283	±	0.0177	&	\textbf{0.8450}	±	0.0223	\\ \hline
Colon	&	0.9004	±	0.0062	&	0.9124	±	0.0076	&	0.9176	±	0.0101	&	\textbf{0.9282}	±	0.0136	\\ \hline
DLBCL	&	0.9875	±	0.0000	&	0.9875	±	0.0000	&	\textbf{0.9888}	±	0.0040	&	0.9873	±	0.0004	\\ \hline
Gastric	&	\textbf{1.0000}	±	0.0000	&	\textbf{1.0000}	±	0.0000	&	\textbf{1.0000}	±	0.0000	&	\textbf{1.0000}	±	0.0000	\\ \hline
Leukemia	&	0.9867	±	0.0000	&	0.9867	±	0.0000	&	0.9880	±	0.0042	&	\textbf{0.9917}	±	0.0071	\\ \hline
Lymphoma	&	0.9222	±	0.0117	&	0.9489	±	0.0150	&	0.9489	±	0.0183	&	\textbf{0.9756}	±	0.0164	\\ \hline
Prostate	&	0.8788	±	0.0045	&	0.8866	±	0.0068	&	0.8897	±	0.0092	&	\textbf{0.9004}	±	0.0259	\\ \hline
Stroke	&	\textbf{1.0000}	±	0.0000	&	\textbf{1.0000}	±	0.0000	&	\textbf{1.0000}	±	0.0000	&	\textbf{1.0000}	±	0.0000	\\ \hline
\textbf{Average}	&	0.9238	±	0.0044	&	0.9326	±	0.0051	&	0.9336	±	0.0072	&	\textbf{0.9477}	±	0.0113	\\ \hline
\end{tabular}}
\caption{Comparison with Evolutionary Algorithms on Accuracy}
\end{table}

\begin{table}[!htb]
\centering
\resizebox{87mm}{!}{
\begin{tabular}{ l l l l l }
\hline
\multirow{2}{*}{\textbf{Dataset}} & \textbf{DE}	&	\textbf{GA (Roulette)}	&	\textbf{GA (Tournament)} &	\textbf{MEL (Ours)} \\ 
 \cline{2-5}
   \textbf{ } & \textbf{Mean $\pm$ Std} & \textbf{Mean $\pm$ Std}  & \textbf{Mean $\pm$ Std} & \textbf{Mean $\pm$ Std} \\
\hline
Adenoma	&	3000.2	±	40.8	&	3744.6	±	47.5	&	3731.4	±	51.6	&	\textbf{271.1}	±	844.3	\\ \hline
ALL$\_$AML	&	3012.3	±	581.8	&	3575.8	±	32.4	&	3563.0	±	50.8	&	\textbf{1050.8}	±	1281.9	\\ \hline
ALL3	&	7776.7	±	677.9	&	6309.3	±	72.7	&	6331.1	±	70.2	&	\textbf{1.0}	±	0.0	\\ \hline
ALL4	&	7736.8	±	737.5	&	6329.5	±	52.3	&	6316.4	±	56.8	&	\textbf{34.1}	±	24.9	\\ \hline
CNS	&	4664.9	±	359.0	&	3520.8	±	45.1	&	3554.9	±	47.0	&	\textbf{57.2}	±	74.0	\\ \hline
Colon	&	1269.1	±	44.7	&	1007.3	±	33.9	&	994.0	±	19.0	&	\textbf{100.1}	±	266.1	\\ \hline
DLBCL	&	4376.7	±	580.5	&	3584.8	±	61.0	&	3555.2	±	29.0	&	\textbf{2697.8}	±	146.6	\\ \hline
Gastric	&	9089.5	±	66.5	&	11336.9	±	77.2	&	11303.4	±	62.0	&	\textbf{1.0}	±	0.0	\\ \hline
Leukemia	&	3065.3	±	303.6	&	3562.9	±	49.0	&	3589.2	±	42.0	&	\textbf{1793.4}	±	1236.5	\\ \hline
Lymphoma	&	2427.4	±	333.8	&	2017.8	±	27.2	&	2004.1	±	29.1	&	\textbf{35.6}	±	46.3	\\ \hline
Prostate	&	8124.3	±	883.5	&	6289.3	±	54.1	&	6305.6	±	55.5	&	\textbf{3055.8}	±	2579.6	\\ \hline
Stroke	&	21835.7	±	82.0	&	27338.7	±	65.2	&	27413.0	±	150.3	&	\textbf{1.0}	±	0.0	\\ \hline
\textbf{Average}	&	6364.9	±	391.0	&	6551.5	±	51.5	&	6555.1	±	55.3	&	\textbf{758.2}	±	541.7	\\ \hline

\end{tabular}}
\caption{Comparison with Evolutionary Algorithms on the Subset Size}
\end{table}

\begin{table}[!htb]
\centering
\resizebox{87mm}{!}{
\begin{tabular}{ l l l l l }
\hline
\multirow{2}{*}{\textbf{Dataset}} & \textbf{DE}	&	\textbf{GA (Roulette)}	&	\textbf{GA (Tournament)} &	\textbf{MEL (Ours)} \\ 
 \cline{2-5}
   \textbf{ } & \textbf{Mean $\pm$ Std} & \textbf{Mean $\pm$ Std}  & \textbf{Mean $\pm$ Std} & \textbf{Mean $\pm$ Std} \\
\hline
Adenoma	&	81.3	±	1.7	&	115.4	±	5.2	&	114.5	±	2.5	&	\textbf{55.5}	±	1.9	\\ \hline
ALL$\_$AML	&	108.8	±	3.2	&	140.9	±	3.1	&	157.9	±	8.1	&	\textbf{66.2}	±	5.8	\\ \hline
ALL3	&	247.2	±	5.5	&	311.0	±	4.4	&	341.0	±	9.5	&	\textbf{62.9}	±	2.3	\\ \hline
ALL4	&	187.2	±	4.6	&	238.5	±	2.8	&	260.6	±	4.4	&	\textbf{66.3}	±	5.6	\\ \hline
CNS	&	95.2	±	1.8	&	127.5	±	3.1	&	143.6	±	8.1	&	\textbf{53.1}	±	0.9	\\ \hline
Colon	&	59.1	±	0.6	&	86.2	±	2.6	&	101.0	±	8.7	&	\textbf{44.9}	±	1.0	\\ \hline
DLBCL	&	107.9	±	1.5	&	145.8	±	2.7	&	162.9	±	3.1	&	\textbf{72.3}	±	0.8	\\ \hline
Gastric	&	241.2	±	4.1	&	264.7	±	2.1	&	284.0	±	2.9	&	\textbf{74.4}	±	1.0	\\ \hline
Leukemia	&	107.0	±	1.4	&	140.8	±	2.0	&	157.4	±	1.6	&	\textbf{66.8}	±	6.1	\\ \hline
Lymphoma	&	394.2	±	0.7	&	90.2	±	1.2	&	104.3	±	0.6	&	\textbf{48.2}	±	1.3	\\ \hline
Prostate	&	202.2	±	3.3	&	259.6	±	3.1	&	281.4	±	2.9	&	\textbf{98.0}	±	19.0	\\ \hline
Stroke	&	303.8	±	5.6	&	318.2	±	3.2	&	350.4	±	4.1	&	\textbf{115.7}	±	1.3	\\ \hline
\textbf{Average}	&	177.9	±	2.8	&	186.6	±	3.0	&	204.9	±	4.7	&	\textbf{68.7}	±	3.9	\\ \hline

\end{tabular}}
\caption{Comparison with Evolutionary Algorithms on Running Time}
\end{table}
 
\begin{figure*}[htp!]
  \centering
  \begin{minipage}[b]{0.1612\textwidth}
    \includegraphics[width=\textwidth]{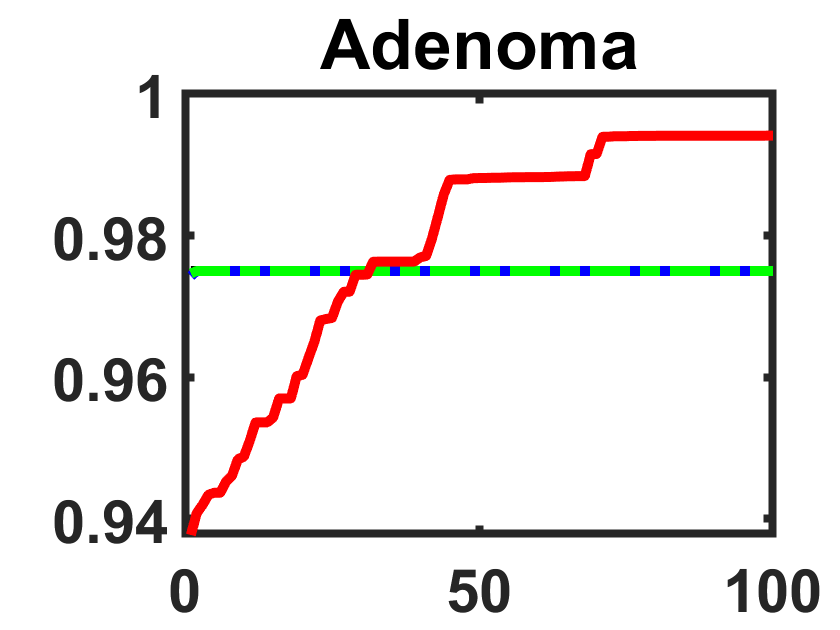}
  \end{minipage}
  \begin{minipage}[b]{0.1612\textwidth}
    \includegraphics[width=\textwidth]{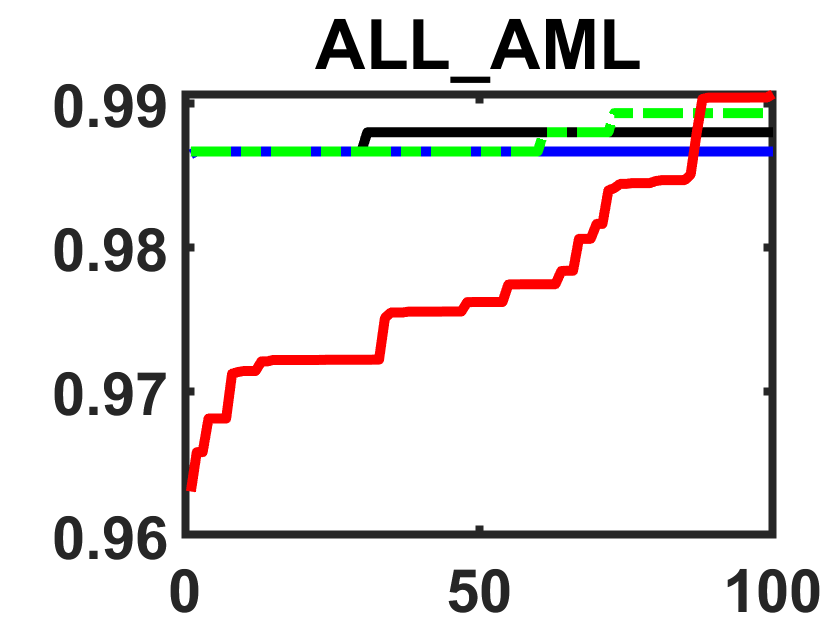}
  \end{minipage}
  \begin{minipage}[b]{0.1612\textwidth}
    \includegraphics[width=\textwidth]{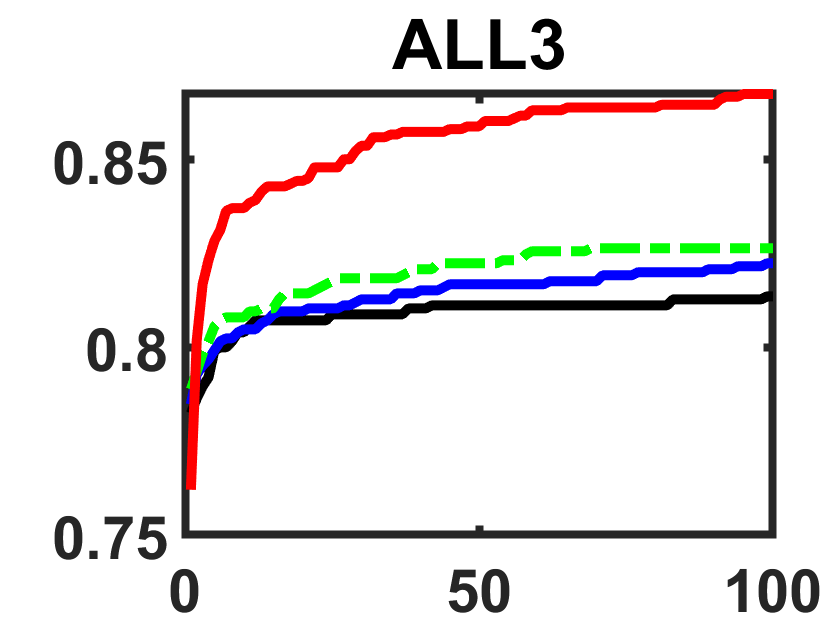}
  \end{minipage}
  \begin{minipage}[b]{0.1612\textwidth}
    \includegraphics[width=\textwidth]{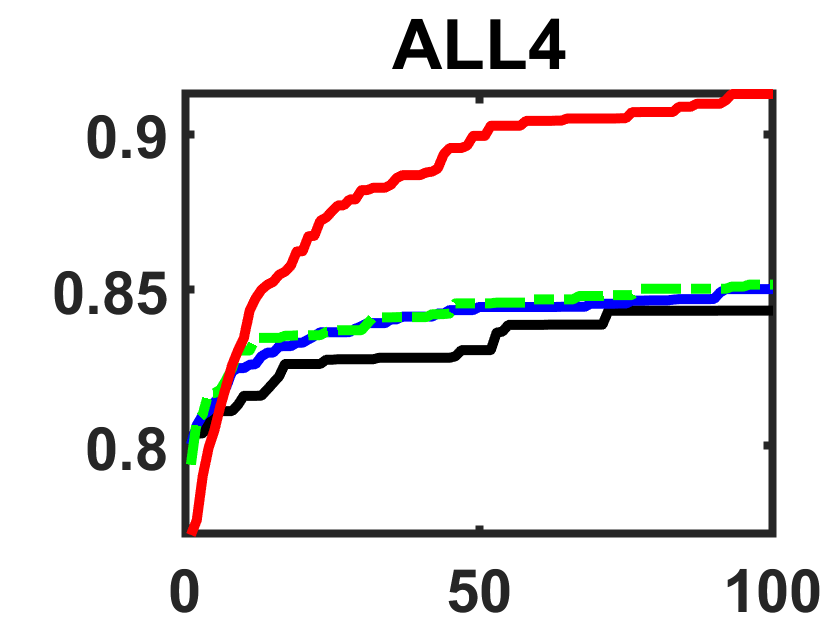}
  \end{minipage}
  \begin{minipage}[b]{0.1612\textwidth}
    \includegraphics[width=\textwidth]{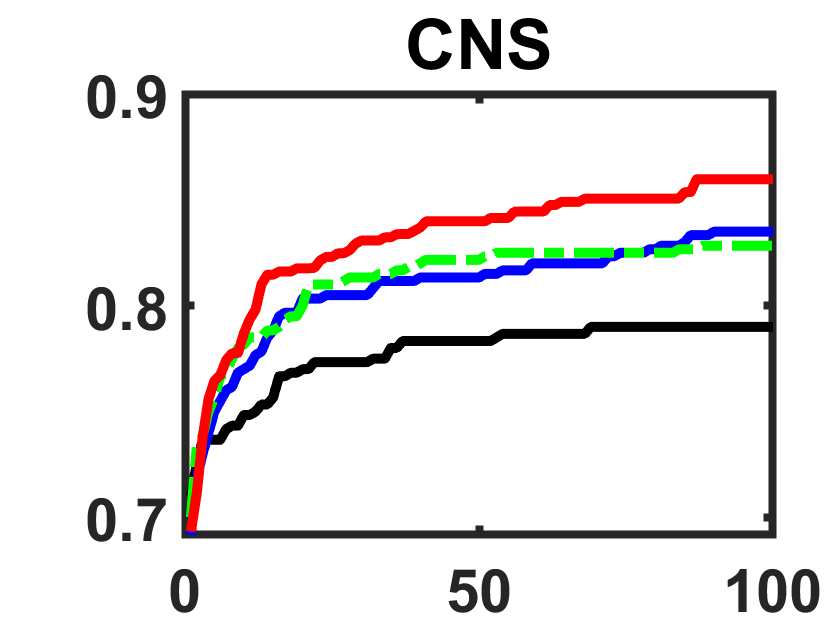}
  \end{minipage}
  \begin{minipage}[b]{0.1612\textwidth}
    \includegraphics[width=\textwidth]{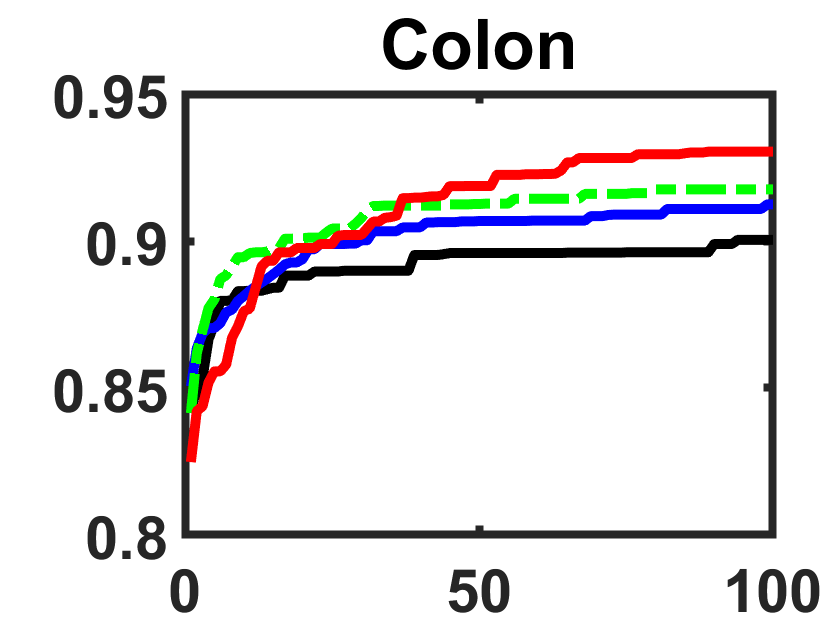}
  \end{minipage}
    \centering
  \begin{minipage}[b]{0.1612\textwidth}
    \includegraphics[width=\textwidth]{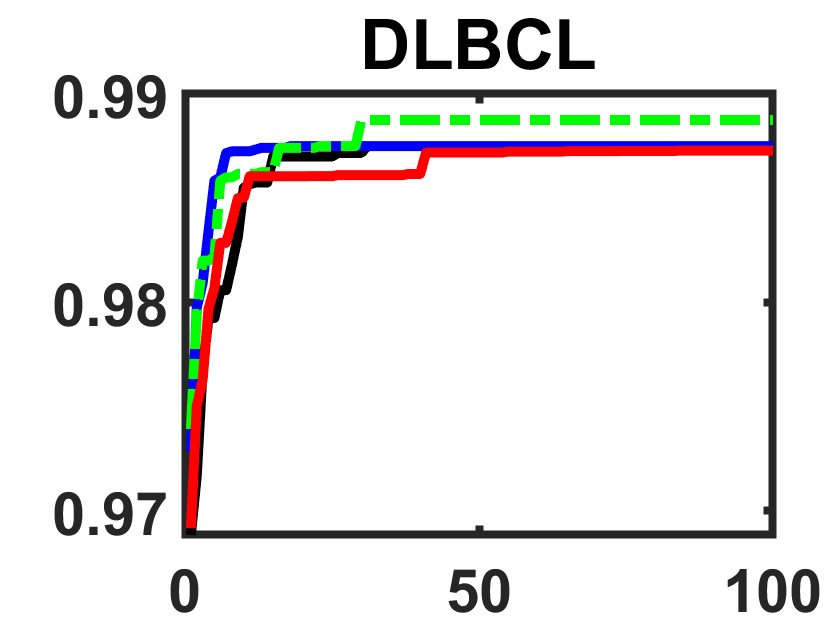}
  \end{minipage}
  \begin{minipage}[b]{0.1612\textwidth}
    \includegraphics[width=\textwidth]{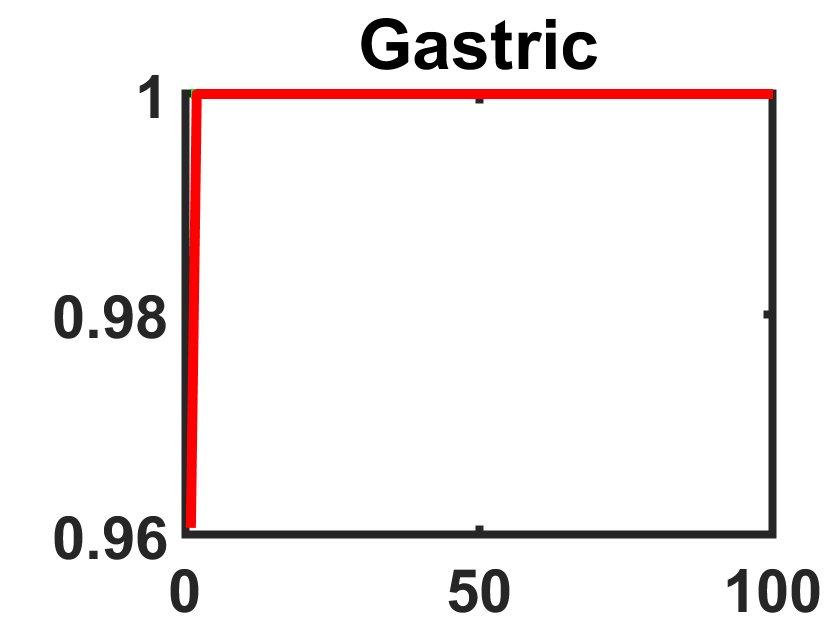}
  \end{minipage}
  \begin{minipage}[b]{0.1612\textwidth}
    \includegraphics[width=\textwidth]{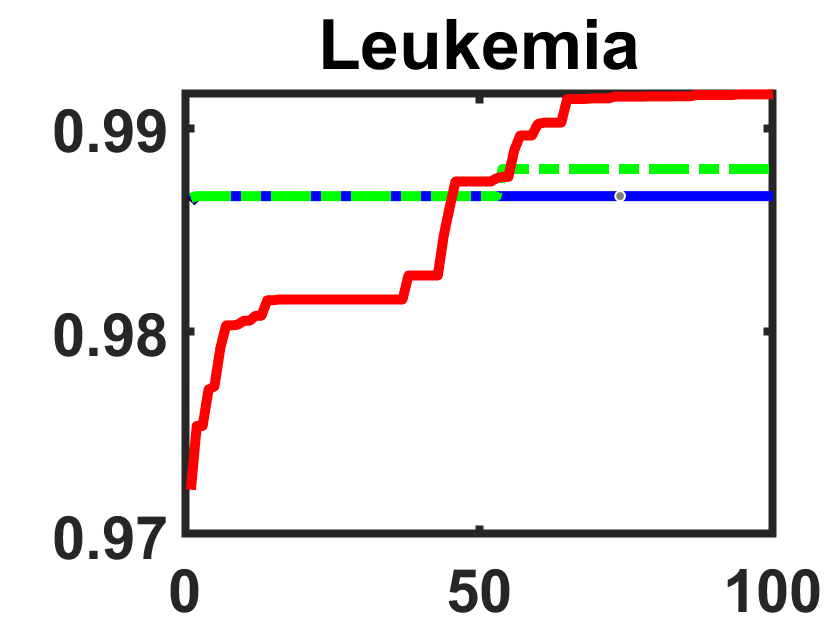}
  \end{minipage}
  \begin{minipage}[b]{0.1612\textwidth}
    \includegraphics[width=\textwidth]{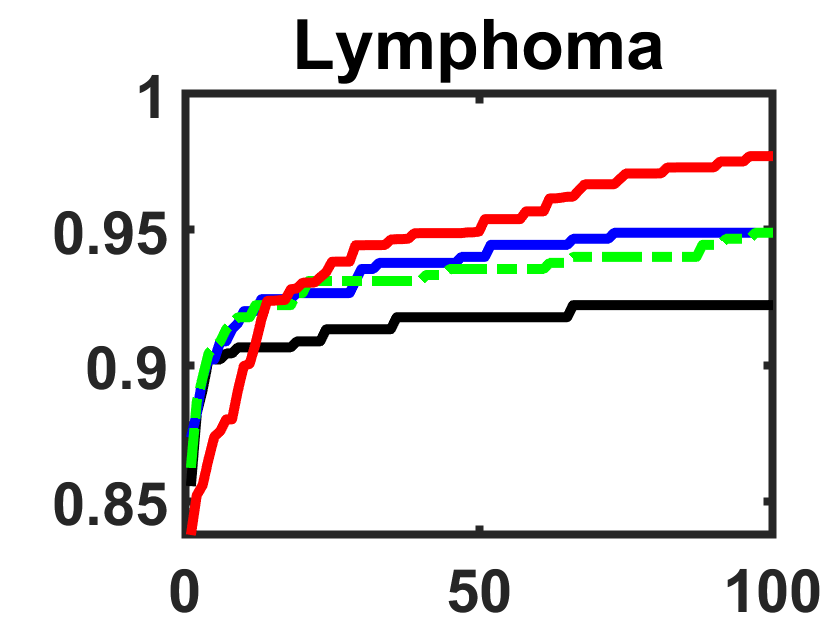}
  \end{minipage}
  \begin{minipage}[b]{0.1612\textwidth}
    \includegraphics[width=\textwidth]{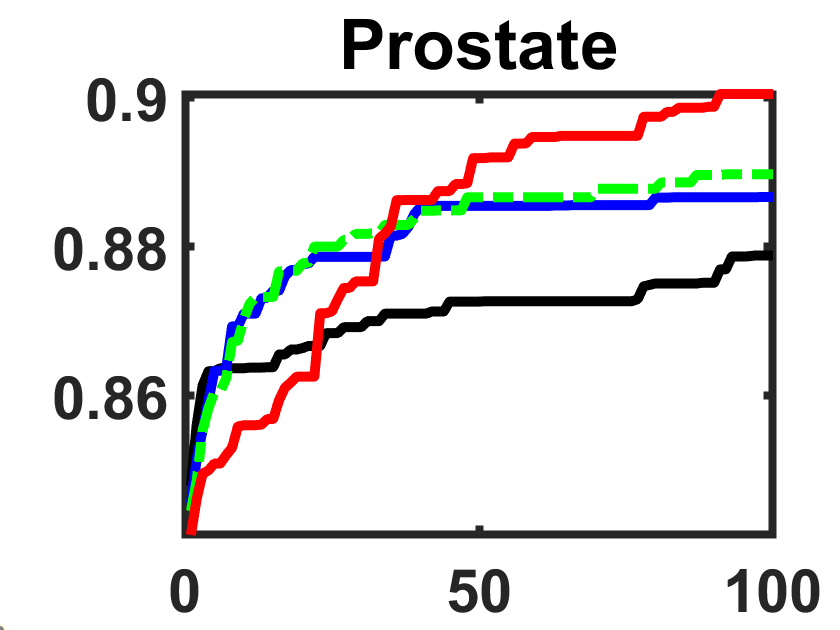}
  \end{minipage}
  \begin{minipage}[b]{0.1612\textwidth}
    \includegraphics[width=\textwidth]{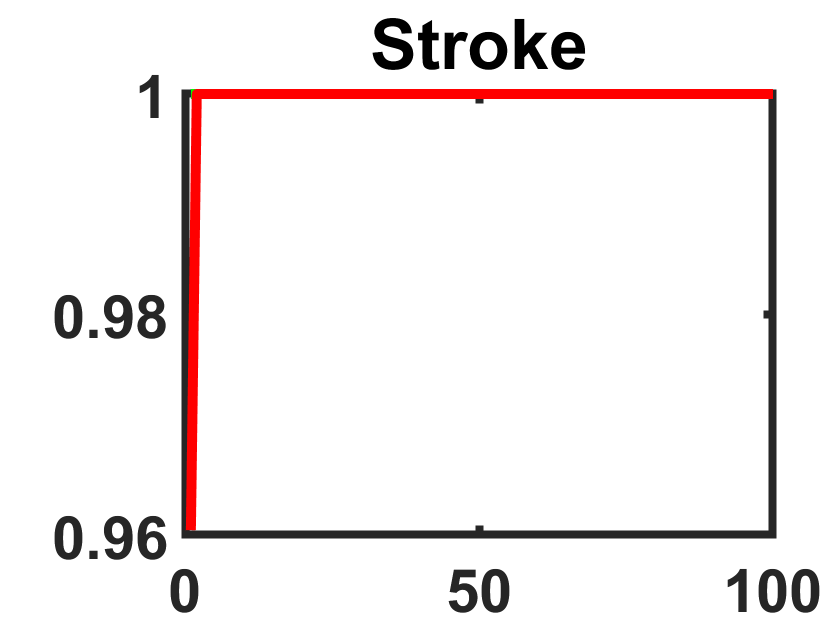}
  \end{minipage}
  \begin{minipage}[b]{0.1612\textwidth}
    \includegraphics[width=\textwidth]{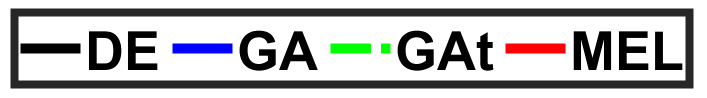}
  \end{minipage}
  \caption{Convergence Curves of Evolutionary Algorithms in Terms of Accuracy}
\end{figure*}

\begin{figure*}[htp!]
  \centering
  \begin{minipage}[b]{0.1612\textwidth}
    \includegraphics[width=\textwidth]{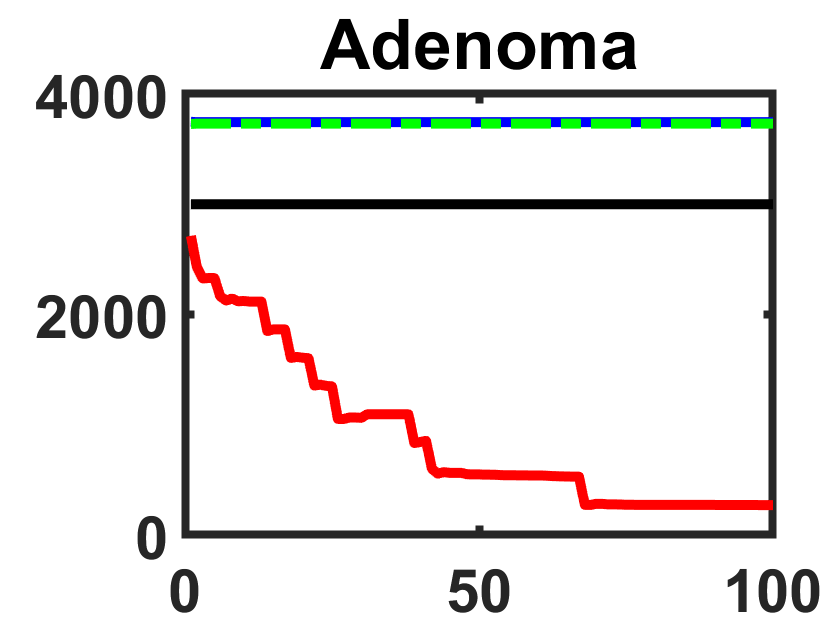}
  \end{minipage}
  \begin{minipage}[b]{0.1612\textwidth}
    \includegraphics[width=\textwidth]{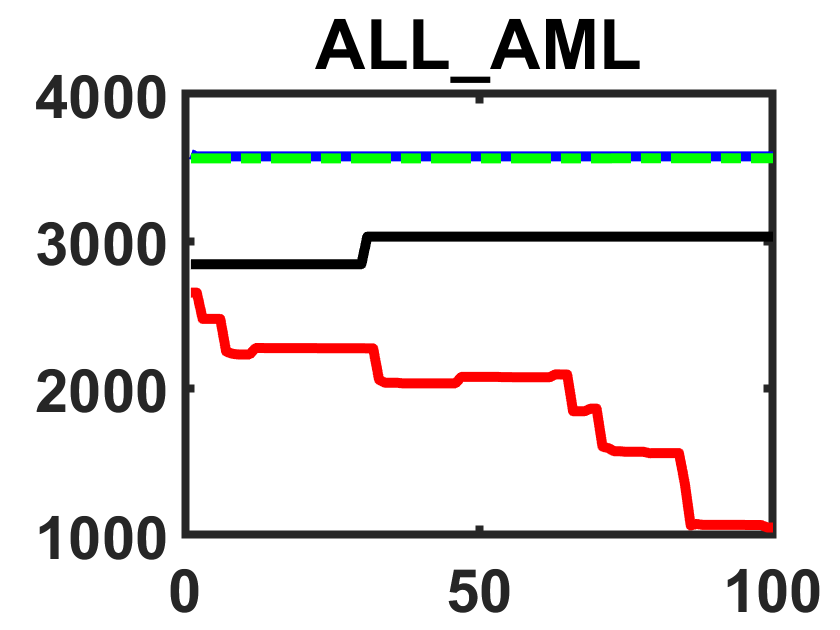}
  \end{minipage}
  \begin{minipage}[b]{0.1612\textwidth}
    \includegraphics[width=\textwidth]{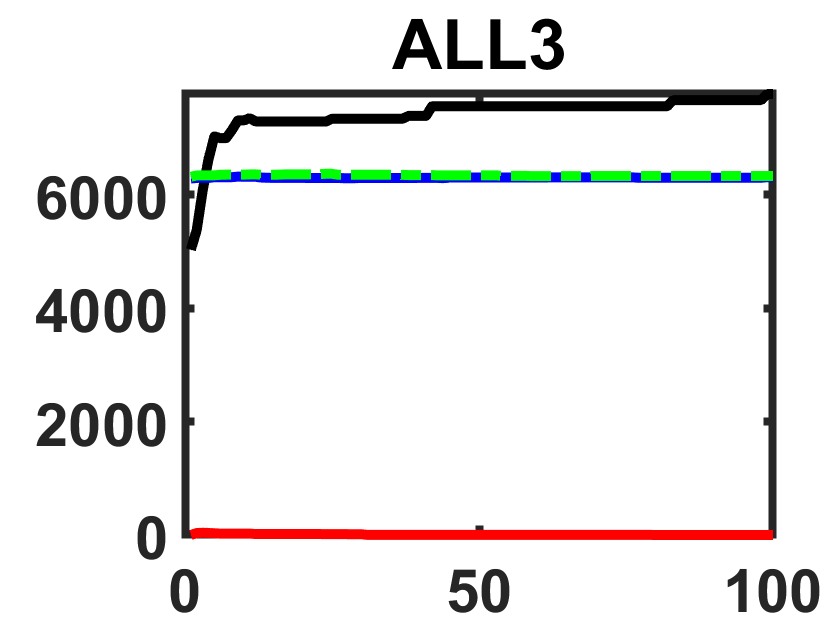}
  \end{minipage}
  \begin{minipage}[b]{0.1612\textwidth}
    \includegraphics[width=\textwidth]{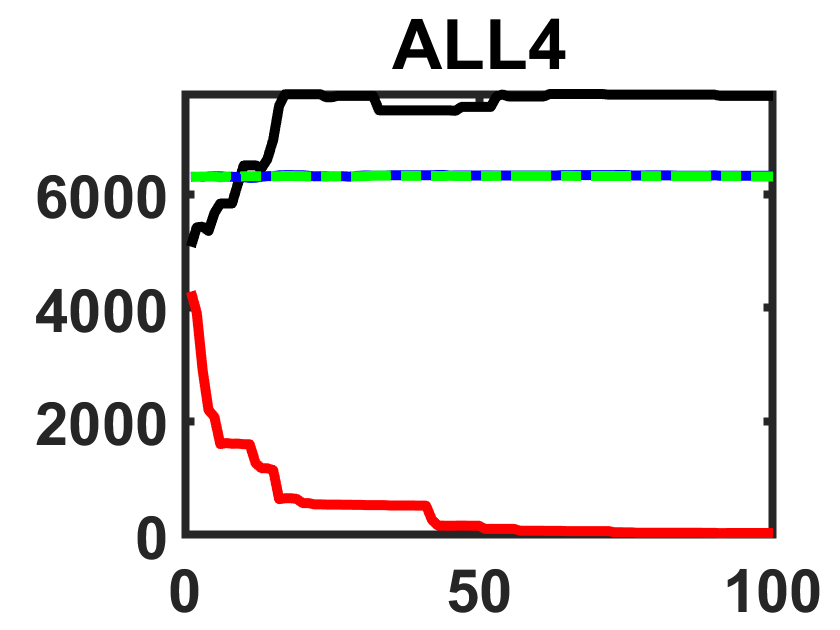}
  \end{minipage}
  \begin{minipage}[b]{0.1612\textwidth}
    \includegraphics[width=\textwidth]{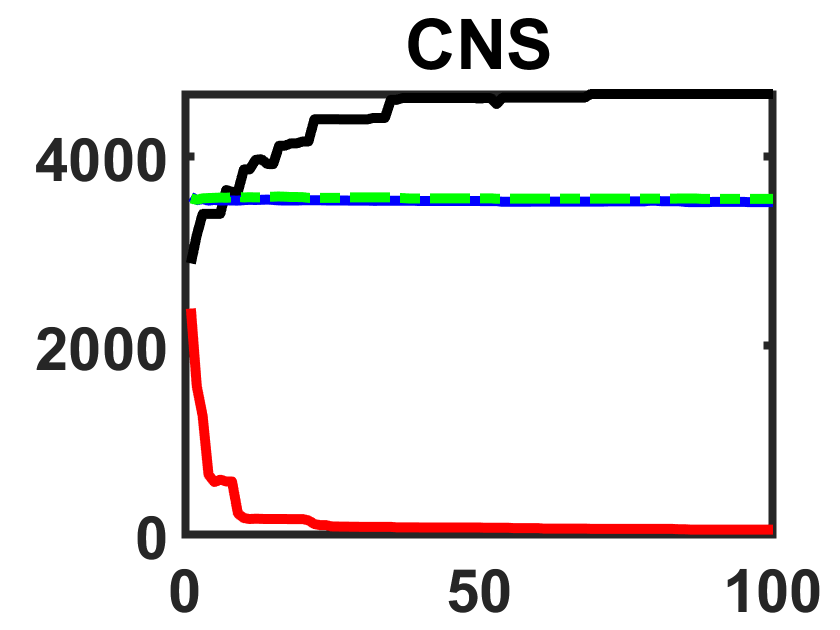}
  \end{minipage}
  \begin{minipage}[b]{0.1612\textwidth}
    \includegraphics[width=\textwidth]{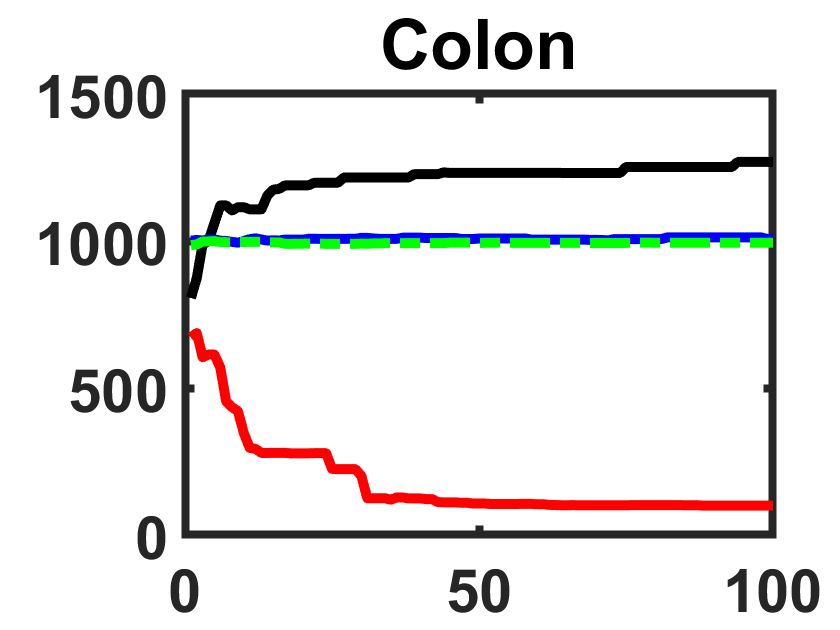}
  \end{minipage}
    \centering
  \begin{minipage}[b]{0.1612\textwidth}
    \includegraphics[width=\textwidth]{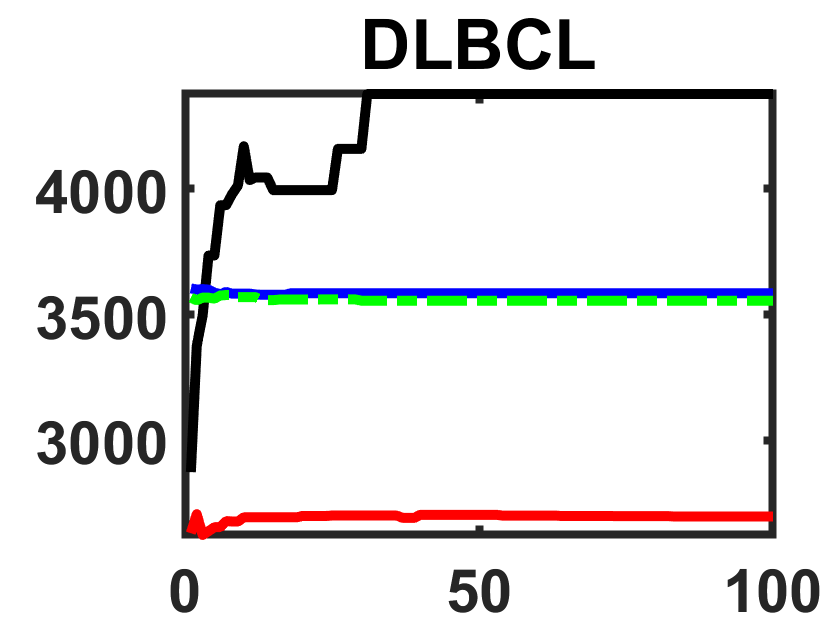}
  \end{minipage}
  \begin{minipage}[b]{0.1612\textwidth}
    \includegraphics[width=\textwidth]{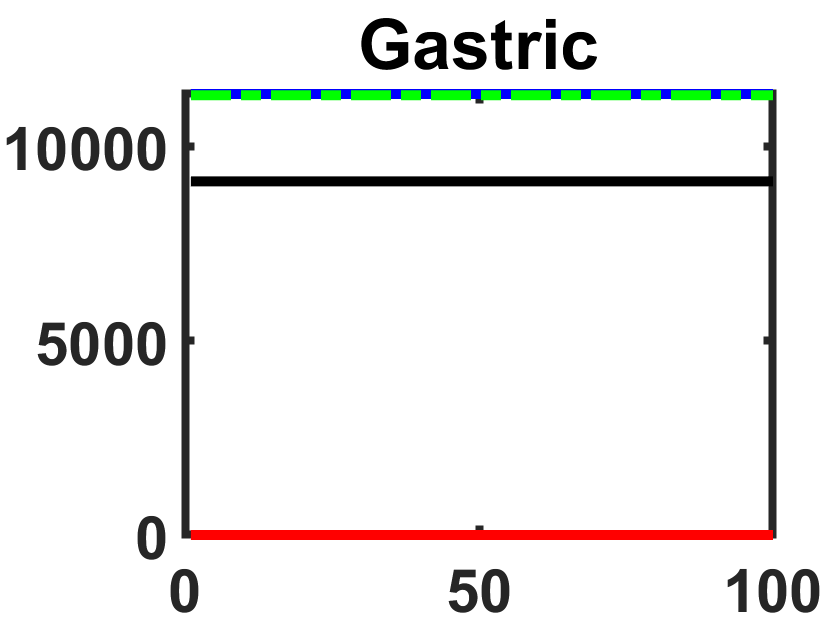}
  \end{minipage}
  \begin{minipage}[b]{0.1612\textwidth}
    \includegraphics[width=\textwidth]{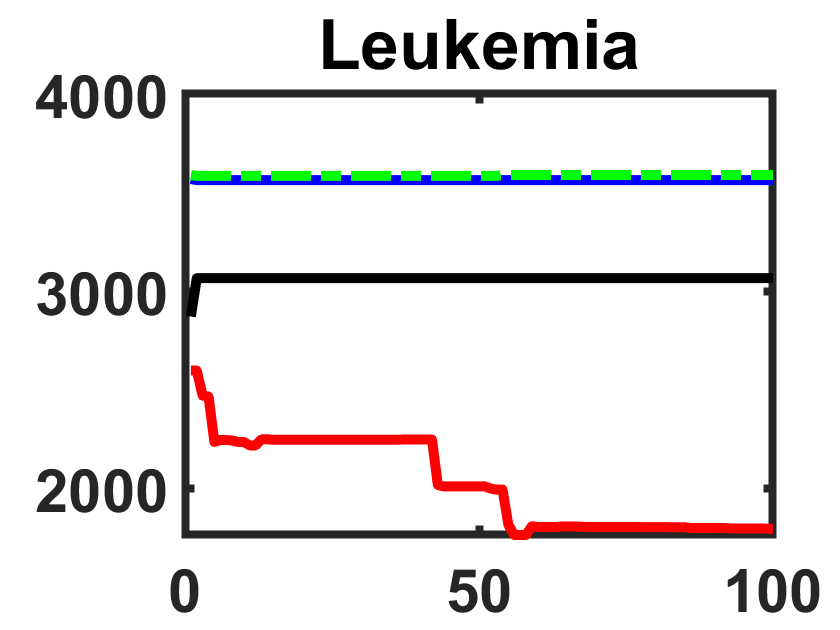}
  \end{minipage}
  \begin{minipage}[b]{0.1612\textwidth}
    \includegraphics[width=\textwidth]{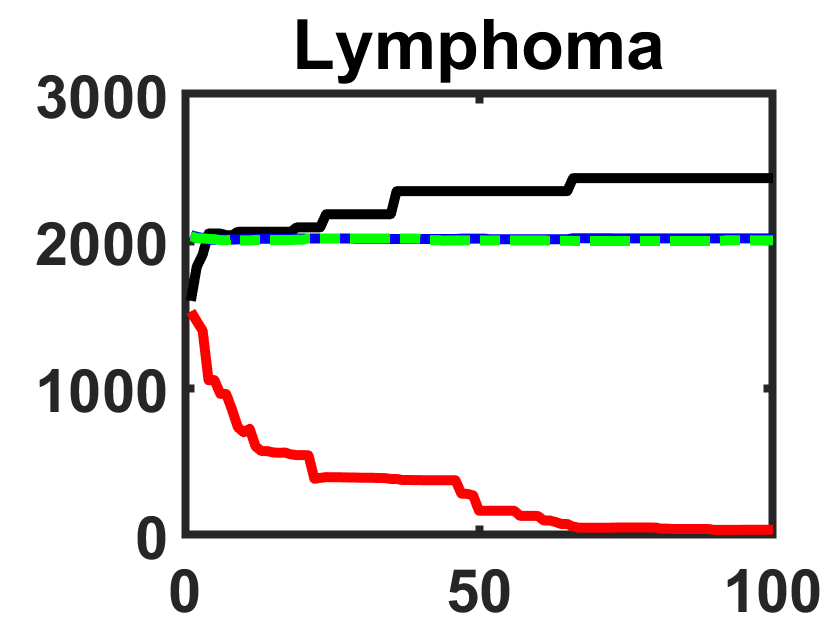}
  \end{minipage}
  \begin{minipage}[b]{0.1612\textwidth}
    \includegraphics[width=\textwidth]{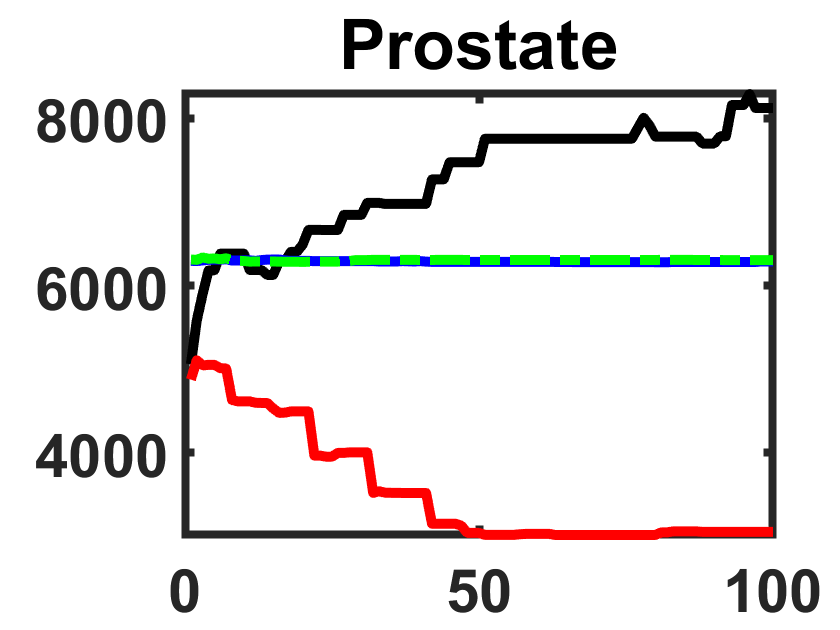}
  \end{minipage}
  \begin{minipage}[b]{0.1612\textwidth}
    \includegraphics[width=\textwidth]{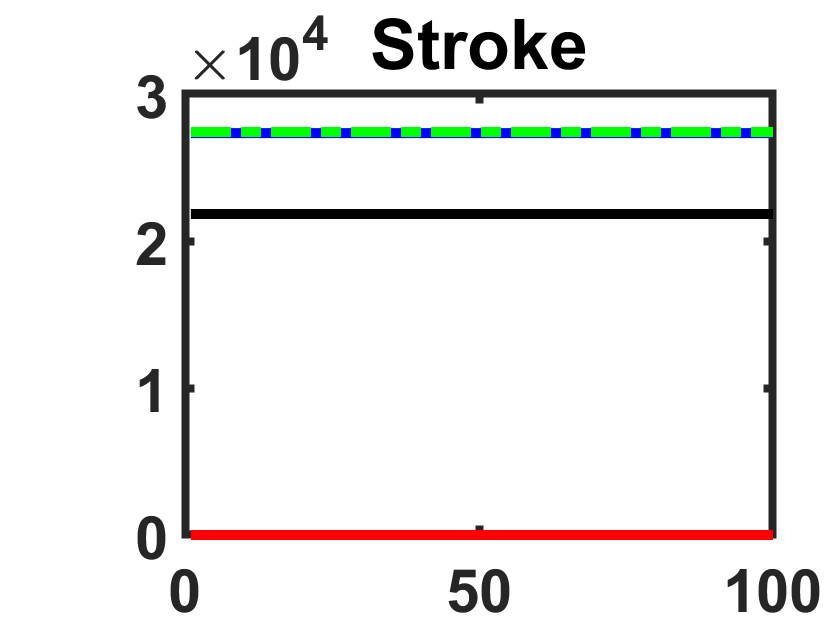}
  \end{minipage}
  \begin{minipage}[b]{0.1612\textwidth}
    \includegraphics[width=\textwidth]{evolutionary/label.png}
  \end{minipage}
  \caption{Convergence Curves of Evolutionary Algorithms in Terms of the Size of Feature Subset}
\end{figure*}

Figures 6 and 7 illustrate the convergence curves of these algorithms during the iterative process. In the figures, ``GA + Roulette" is represented as GA, and ``GA + Tournament" is represented as GAt. It is evident from the figures that MEL exhibits significant advantages over these methods. Additionally, we observed that the difference between GA methods using the Roulette and Tournament mechanisms was not substantial.

\subsection{Comparison with Bio-stimulated Heuristic Methods}
TABLE XII and TABLE XIII present the results of MEL and bio-stimulated methods in terms of classification accuracy and feature subset size. Figures 8 and 9 depict the convergence curves of MEL and bio-stimulated methods in terms of classification accuracy and feature subset size, respectively. Overall, MEL outperforms the bio-stimulated methods in both classification accuracy and subset size. However, we observed that algorithms such as GWO, HHO, and WOA, which were proposed in recent years, exhibit superior search capabilities on certain datasets (e.g., ALL\_AML, DLBCL, Leukemia). It is worth noting that these algorithms demonstrate promising performance in specific scenarios but do not consistently outperform MEL across all datasets. 

\begin{table}[!htb]
\centering
\resizebox{87mm}{!}{
\begin{tabular}{ l l l l l l }
\hline
\multirow{2}{*}{\textbf{Dataset}} & \textbf{FOA}	&	\textbf{GWO}	&	\textbf{HHO}	&	\textbf{WOA}	&	\textbf{MEL (Ours)} \\ 
 \cline{2-6}
   \textbf{ } & \textbf{Mean $\pm$ Std} & \textbf{Mean $\pm$ Std}  & \textbf{Mean $\pm$ Std} & \textbf{Mean $\pm$ Std}  & \textbf{Mean $\pm$ Std} \\
\hline
Adenoma	&	0.9750	±	0.0000	&	0.9750	±	0.0000	&	0.9800	±	0.0105	&	0.9825	±	0.0121	&	\textbf{0.9975}	±	0.0079	\\ \hline
ALL$\_$AML	&	0.9867	±	0.0000	&	\textbf{0.9920}	±	0.0069	&	0.9907	±	0.0064	&	0.9893	±	0.0056	&	0.9907	±	0.0064	\\ \hline
ALL3	&	0.8120	±	0.0057	&	0.8216	±	0.0113	&	0.8504	±	0.0185	&	0.8184	±	0.0131	&	\textbf{0.8528}	±	0.0137	\\ \hline
ALL4	&	0.8422	±	0.0133	&	0.8541	±	0.0173	&	0.8657	±	0.0212	&	0.8489	±	0.0221	&	\textbf{0.9035}	±	0.0219	\\ \hline
CNS	&	0.7833	±	0.0192	&	0.8233	±	0.0161	&	0.8217	±	0.0352	&	0.7850	±	0.0200	&	\textbf{0.8450}	±	0.0223	\\ \hline
Colon	&	0.8979	±	0.0077	&	0.9204	±	0.0147	&	0.9064	±	0.0218	&	0.9017	±	0.0099	&	\textbf{0.9282}	±	0.0136	\\ \hline
DLBCL	&	0.9875	±	0.0000	&	0.9875	±	0.0000	&	\textbf{0.9888}	±	0.0040	&	0.9875	±	0.0000	&	0.9873	±	0.0004	\\ \hline
Gastric	&	\textbf{1.0000}	±	0.0000	&	\textbf{1.0000}	±	0.0000	&	\textbf{1.0000}	±	0.0000	&	\textbf{1.0000}	±	0.0000	&	\textbf{1.0000}	±	0.0000	\\ \hline
Leukemia	&	0.9867	±	0.0000	&	0.9907	±	0.0064	&	0.9893	±	0.0056	&	\textbf{0.9920}	±	0.0069	&	0.9917	±	0.0071	\\ \hline
Lymphoma	&	0.9200	±	0.0115	&	0.9467	±	0.0187	&	0.9267	±	0.0211	&	0.9533	±	0.0195	&	\textbf{0.9756}	±	0.0164	\\ \hline
Prostate	&	0.8845	±	0.0101	&	0.8957	±	0.0178	&	0.8987	±	0.0231	&	0.8908	±	0.0152	&	\textbf{0.9004}	±	0.0259	\\ \hline
Stroke	&	\textbf{1.0000}	±	0.0000	&	\textbf{1.0000}	±	0.0000	&	\textbf{1.0000}	±	0.0000	&	\textbf{1.0000}	±	0.0000	&	\textbf{1.0000}	±	0.0000	\\ \hline
\textbf{Average}	&	0.9230	±	0.0056	&	0.9339	±	0.0091	&	0.9349	±	0.0140	&	0.9291	±	0.0104	&	\textbf{0.9477}	±	0.0113	\\ \hline
\end{tabular}}
\caption{Comparison with Bio-stimulated Heuristic Methods on Accuracy}
\end{table}

\begin{table}[!htb]
\centering
\resizebox{87mm}{!}{
\begin{tabular}{ l l l l l l }
\hline
\multirow{2}{*}{\textbf{Dataset}} & \textbf{FOA}	&	\textbf{GWO}	&	\textbf{HHO}	&	\textbf{WOA}	&	\textbf{MEL (Ours)} \\ 
 \cline{2-6}
   \textbf{ } & \textbf{Mean $\pm$ Std} & \textbf{Mean $\pm$ Std}  & \textbf{Mean $\pm$ Std} & \textbf{Mean $\pm$ Std}  & \textbf{Mean $\pm$ Std} \\
\hline
Adenoma	&	7457.0	±	0.0000	&	2955.3	±	54.5	&	2423.6	±	1237.4	&	2177.5	±	1415.2	&	\textbf{271.1}	±	844.3	\\ \hline
ALL$\_$AML	&	7039.3	±	283.7	&	2625.8	±	202.3	&	2533.5	±	638.0	&	2851.5	±	1476.1	&	\textbf{1050.8}	±	1281.9	\\ \hline
ALL3	&	10240.5	±	1406.2	&	3906.0	±	337.7	&	925.9	±	2666.8	&	3782.8	±	2856.9	&	\textbf{1.0}	±	0.0	\\ \hline
ALL4	&	7035.9	±	2118.1	&	3547.6	±	352.3	&	2760.7	±	2558.7	&	5130.6	±	5148.0	&	\textbf{34.1}	±	24.9	\\ \hline
CNS	&	4080.5	±	1135.9	&	2008.3	±	160.8	&	1884.0	±	1559.4	&	4354.7	±	1140.9	&	\textbf{57.2}	±	74.0	\\ \hline
Colon	&	1269.9	±	333.0	&	548.0	±	37.0	&	460.7	±	321.7	&	1111.7	±	488.6	&	\textbf{100.1}	±	266.1	\\ \hline
DLBCL	&	4768.4	±	843.5	&	2474.3	±	96.7	&	\textbf{2098.0}	±	1304.6	&	2434.8	±	661.7	&	2697.8	±	146.6	\\ \hline
Gastric	&	22645.0	±	0.0	&	9047.4	±	78.9	&	9071.6	±	78.4	&	9030.6	±	81.3	&	\textbf{1.0}	±	0.0	\\ \hline
Leukemia	&	7043.7	±	269.7	&	2719.6	±	225.6	&	2754.8	±	481.2	&	2177.4	±	1766.8	&	\textbf{1793.4}	±	1236.5	\\ \hline
Lymphoma	&	2880.9	±	584.3	&	1293.0	±	73.8	&	1724.5	±	369.6	&	883.8	±	436.2	&	\textbf{35.6}	±	46.3	\\ \hline
Prostate	&	7653.1	±	2180.3	&	3359.2	±	340.4	&	4434.2	±	3382.2	&	4114.3	±	4711.5	&	\textbf{3055.8}	±	2579.6	\\ \hline
Stroke	&	54675.0	±	0.0	&	21931.4	±	96.4	&	21884.1	±	66.5	&	21862.6	±	105.3	&	\textbf{1.0}	±	0.0	\\ \hline
\textbf{Average}	&	11399.1	±	762.9	&	4701.3	±	171.4	&	4413.0	±	1222.0	&	4992.7	±	1690.7	&	\textbf{758.2}	±	541.7	\\ \hline
\end{tabular}}
\caption{Comparison with Bio-stimulated Heuristic Methods on Subset Size}
\end{table}

\begin{table}[!htb]
\centering
\resizebox{87mm}{!}{
\begin{tabular}{ l l l l l l }
\hline
\multirow{2}{*}{\textbf{Dataset}} & \textbf{FOA}	&	\textbf{GWO}	&	\textbf{HHO}	&	\textbf{WOA}	&	\textbf{MEL (Ours)} \\ 
 \cline{2-6}
   \textbf{ } & \textbf{Mean $\pm$ Std} & \textbf{Mean $\pm$ Std}  & \textbf{Mean $\pm$ Std} & \textbf{Mean $\pm$ Std}  & \textbf{Mean $\pm$ Std} \\
\hline
Adenoma	&	130.0	±	6.6	&	66.8	±	1.0	&	104.9	±	8.3 &		60.8	±	7.5	&	\textbf{55.5}	±	1.9	\\ \hline
ALL$\_$AML	&	273.2	±	24.3	&	80.3	±	1.6	&	126.8	±	12.1	&	79.7	±	12.9	&	\textbf{66.2}	±	5.8	\\ \hline
ALL3	&	170.6	±	23.2	&	150.7	±	6.0	&	138.2	±	92.3	&	134.5	±	54.6	&	\textbf{62.9}	±	2.3	\\ \hline
ALL4	&	89.0	±	7.3	&	115.3	±	4.1	&	154.1	±	55.1	&	126.0	±	74.0	&	\textbf{66.3}	±	5.6	\\ \hline
CNS	&	58.3	±	2.7	&	70.0	±	1.3	&	106.7	±	16.3	&	82.5	±	10.1	&	\textbf{53.1}	±	0.9	\\ \hline
Colon	&	106.5	±	7.6	&	53.3	±	0.4	&	80.8	±	6.8	&	53.1	±	4.9	&	\textbf{44.9}	±	1.0	\\ \hline
DLBCL	&	267.3	±	2.2	&	81.5	±	0.8	&	123.8	±	27.0	&	74.9	±	8.3	&	\textbf{72.3}	±	0.8	\\ \hline
Gastric	&	129.6	±	6.0	&	131.8	±	1.4	&	236.6	±	12.2	&	139.7	±	3.4	&	\textbf{74.4}	±	1.0	\\ \hline
Leukemia	&	\textbf{65.0}	±	3.0	&	79.6	±	1.2	&	131.2	±	9.1	&	72.3	±	15.5	&	66.8	±	6.1	\\ \hline
Lymphoma	&	195.8	±	25.7	&	57.6	±	0.4	&	96.1	±	8.7	&	49.3	±	5.3	&	\textbf{48.2}	±	1.3	\\ \hline
Prostate	&	346.2	±	4.4	&	118.8	±	3.9	&	208.9	±	77.1	&	119.6	±	70.9	&	\textbf{98.0}	±	19.0	\\ \hline
Stroke	&	\textbf{91.2}	±	0.8	&	171.4	±	0.6	&	279.7	±	8.0	&	170.3	±	4.2	&	115.7	±	1.3	\\ \hline
\textbf{Average}	&	160.2	±	9.5	&	98.1	±	1.9	&	149.0	±	27.8	&	96.9	±	22.6	&	\textbf{68.7}	±	3.9	\\ \hline
\end{tabular}}
\caption{Comparison with Bio-stimulated Heuristic Methods on Running Time}
\end{table}

\begin{figure*}[htp!]
  \centering
  \begin{minipage}[b]{0.1612\textwidth}
    \includegraphics[width=\textwidth]{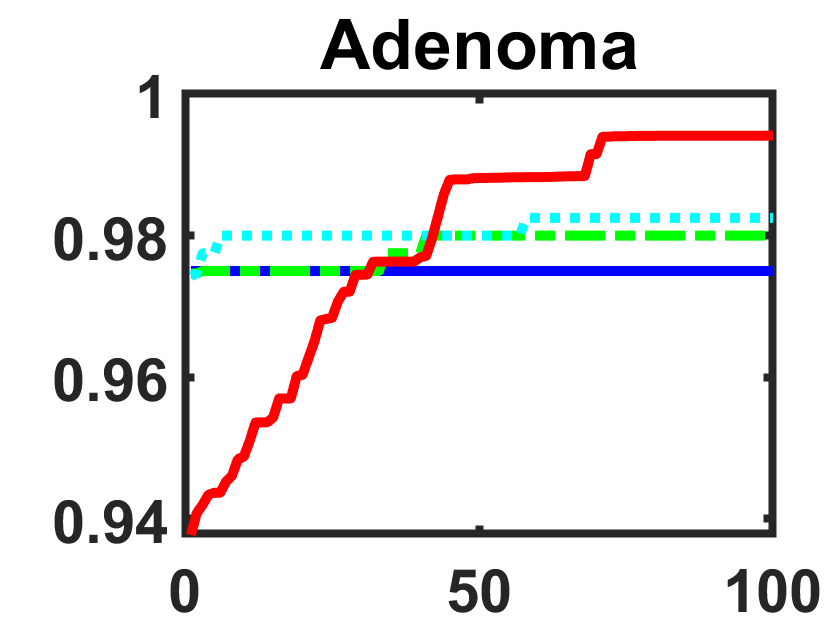}
  \end{minipage}
  \begin{minipage}[b]{0.1612\textwidth}
    \includegraphics[width=\textwidth]{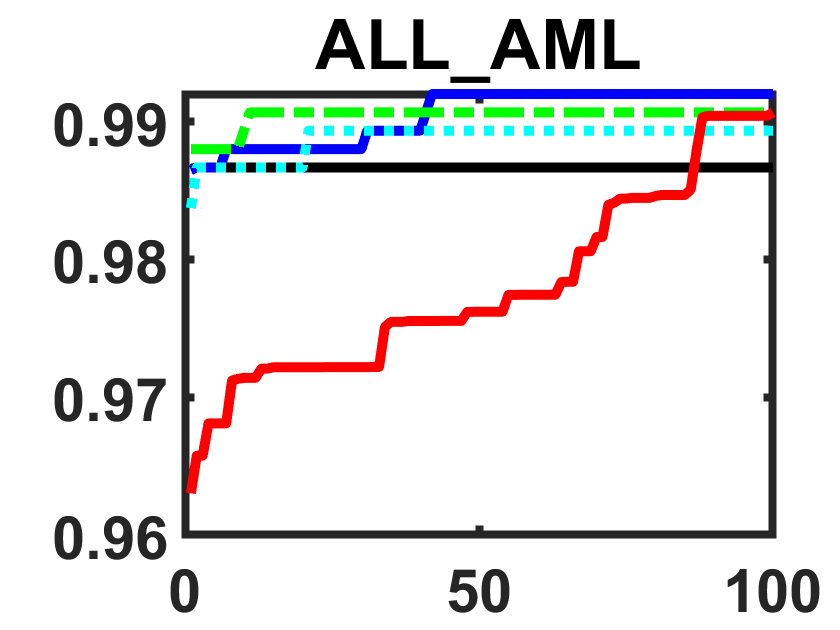}
  \end{minipage}
  \begin{minipage}[b]{0.1612\textwidth}
    \includegraphics[width=\textwidth]{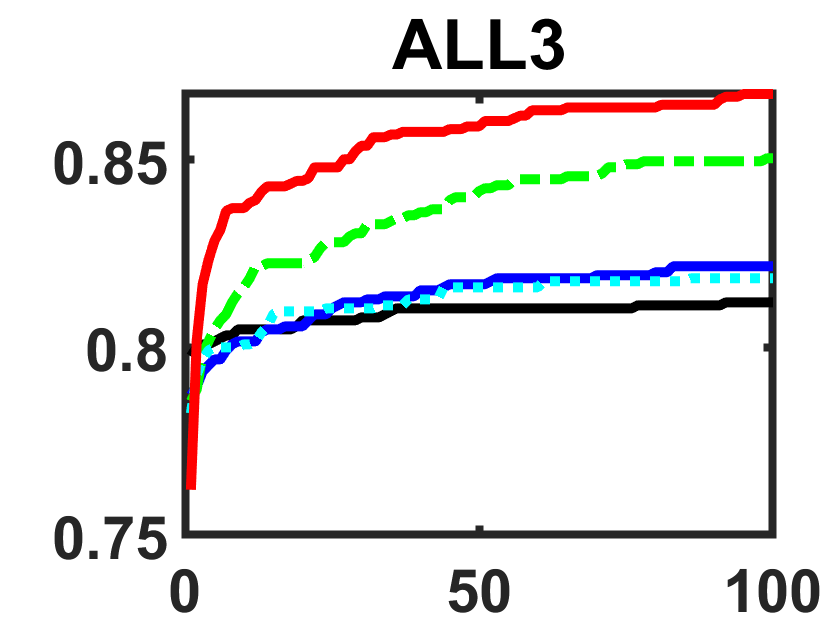}
  \end{minipage}
  \begin{minipage}[b]{0.1612\textwidth}
    \includegraphics[width=\textwidth]{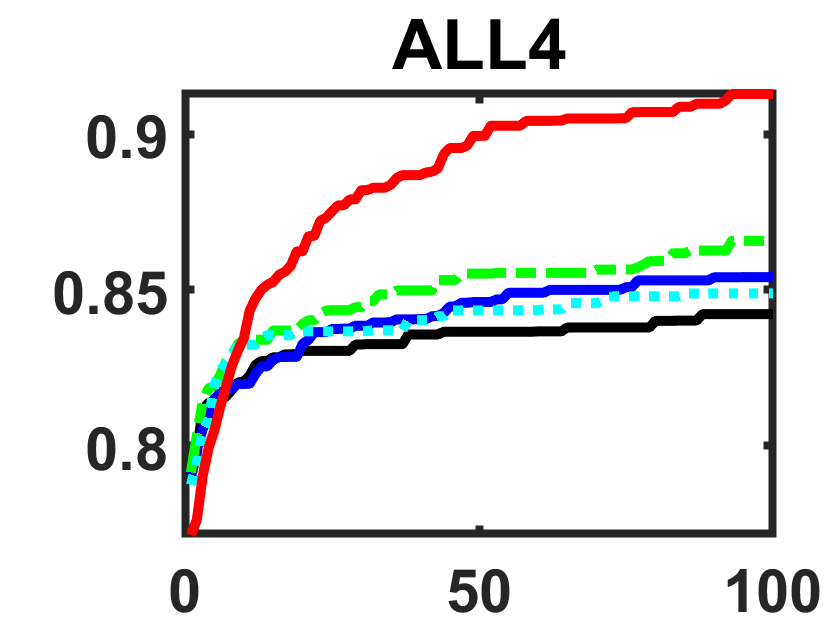}
  \end{minipage}
  \begin{minipage}[b]{0.1612\textwidth}
    \includegraphics[width=\textwidth]{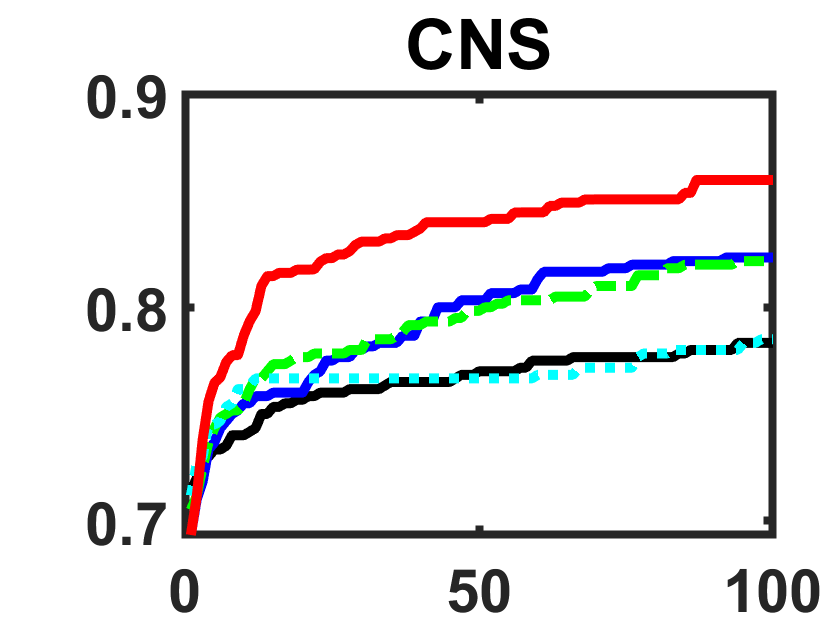}
  \end{minipage}
  \begin{minipage}[b]{0.1612\textwidth}
    \includegraphics[width=\textwidth]{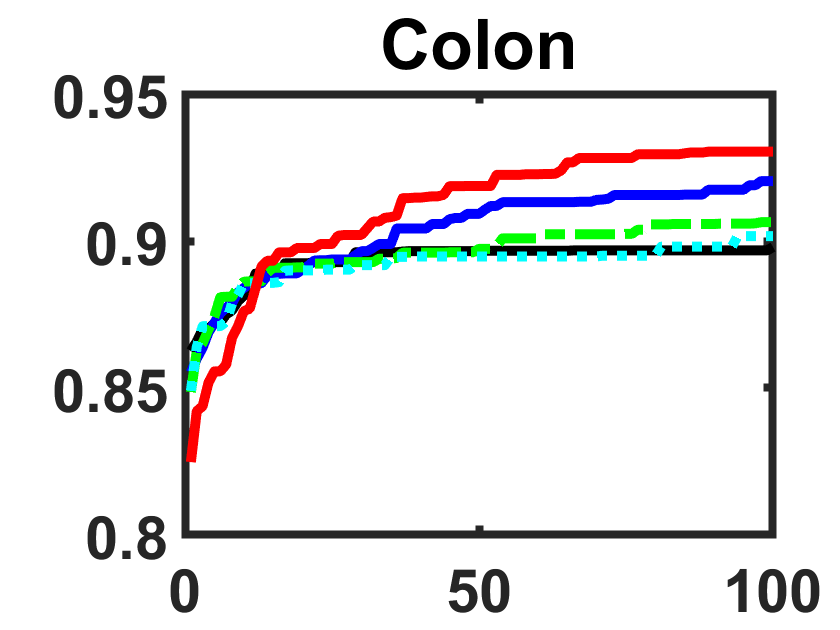}
  \end{minipage}
    \centering
  \begin{minipage}[b]{0.1612\textwidth}
    \includegraphics[width=\textwidth]{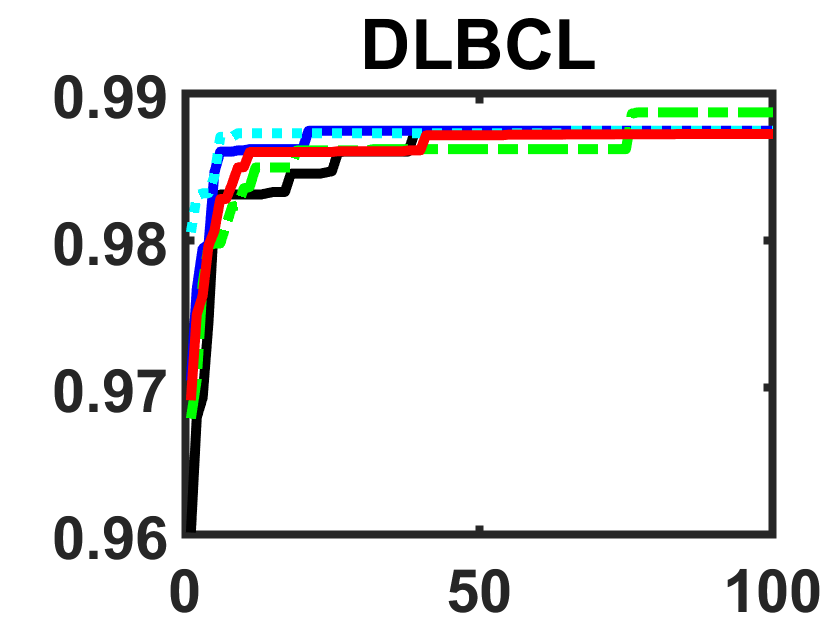}
  \end{minipage}
  \begin{minipage}[b]{0.1612\textwidth}
    \includegraphics[width=\textwidth]{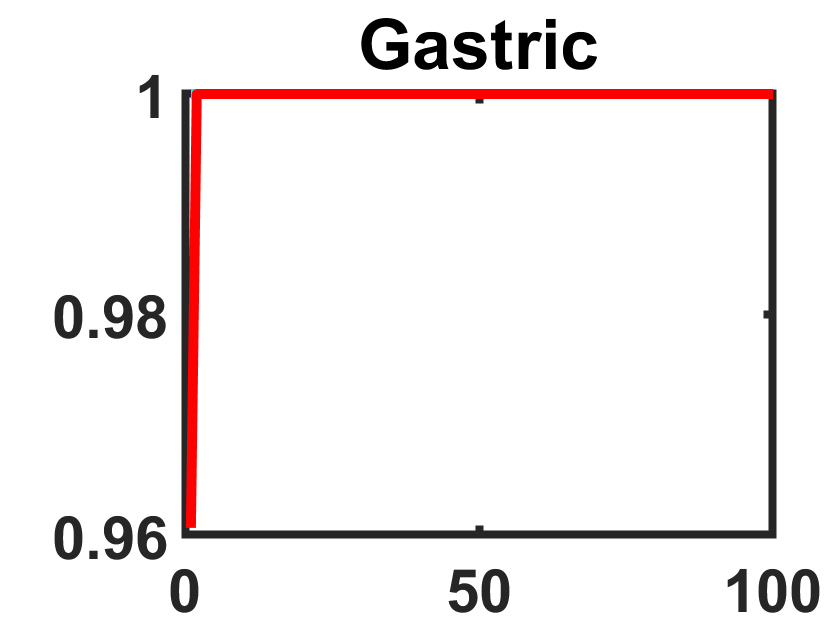}
  \end{minipage}
  \begin{minipage}[b]{0.1612\textwidth}
    \includegraphics[width=\textwidth]{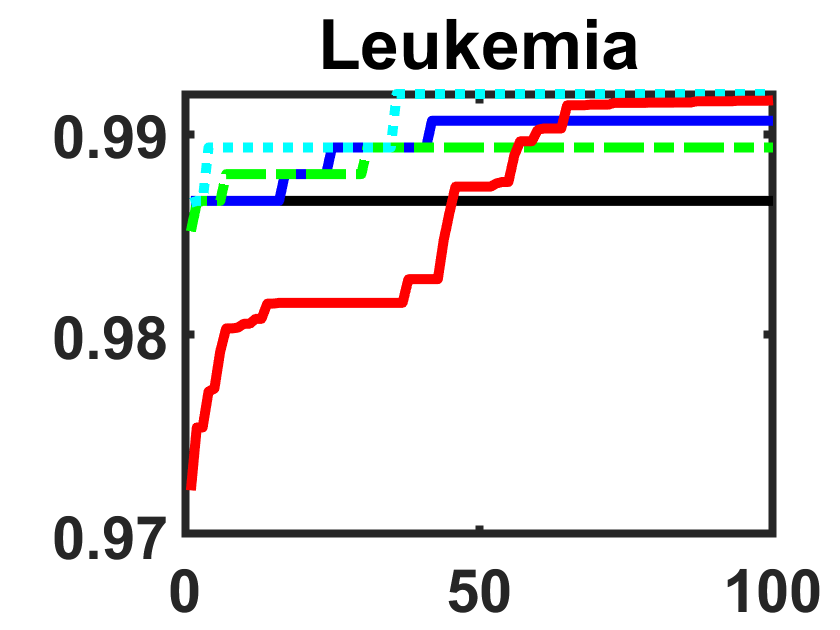}
  \end{minipage}
  \begin{minipage}[b]{0.1612\textwidth}
    \includegraphics[width=\textwidth]{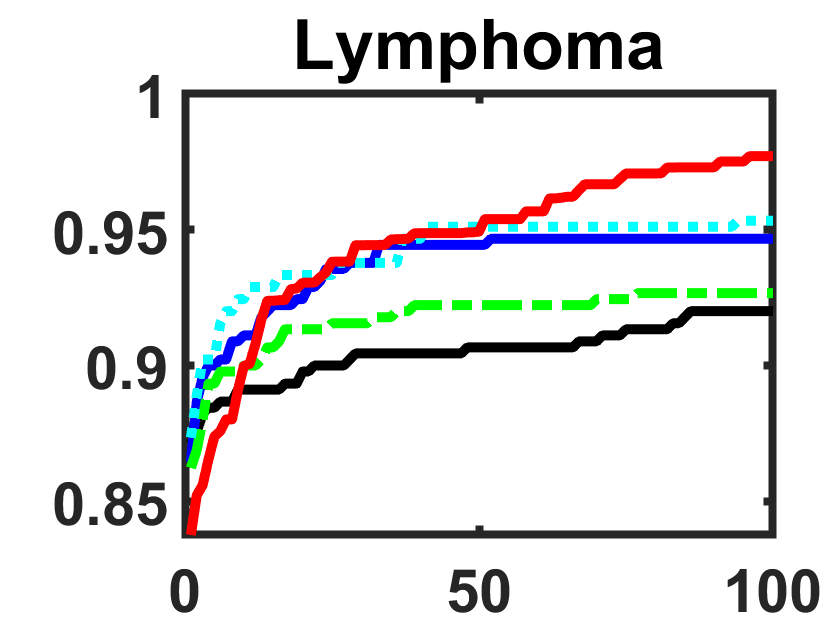}
  \end{minipage}
  \begin{minipage}[b]{0.1612\textwidth}
    \includegraphics[width=\textwidth]{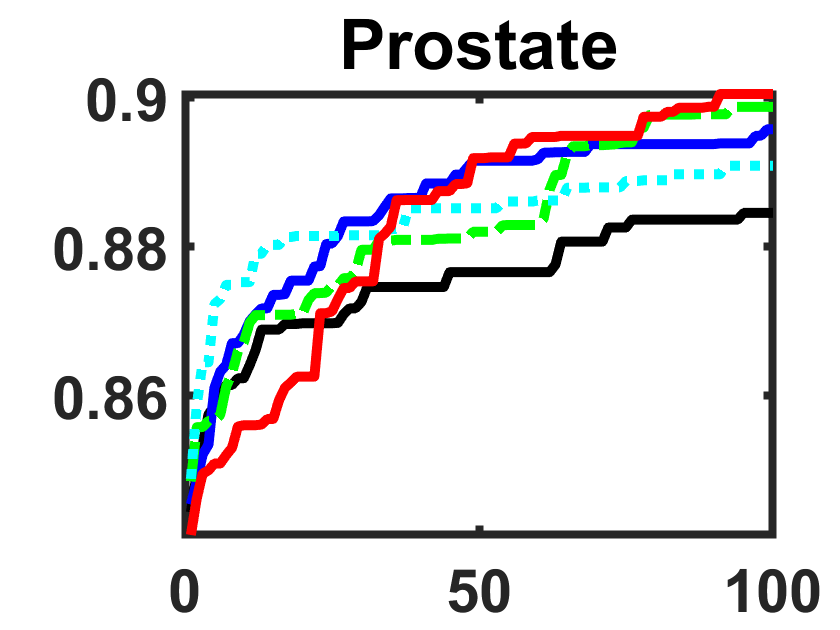}
  \end{minipage}
  \begin{minipage}[b]{0.1612\textwidth}
    \includegraphics[width=\textwidth]{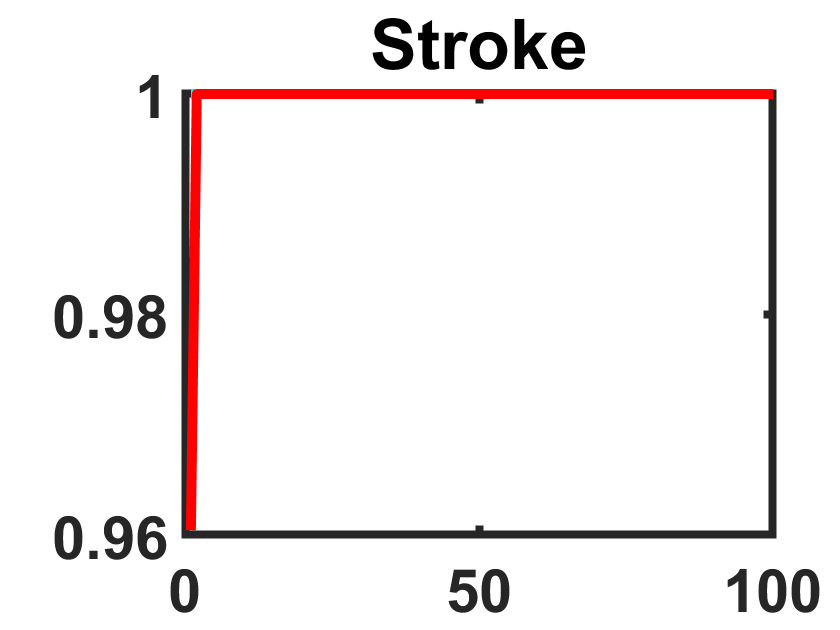}
  \end{minipage}
   \begin{minipage}[b]{0.20\textwidth}
    \includegraphics[width=\textwidth]{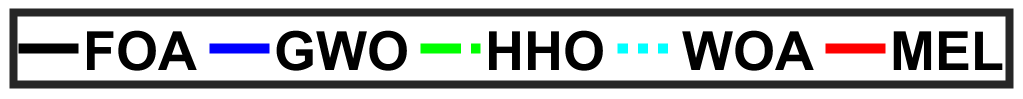}
  \end{minipage}
  \caption{Convergence Curves of Bio-stimulated EC Algorithms in Terms of Accuracy}
\end{figure*}

\begin{figure*}[htp!]
  \centering
  \begin{minipage}[b]{0.1612\textwidth}
    \includegraphics[width=\textwidth]{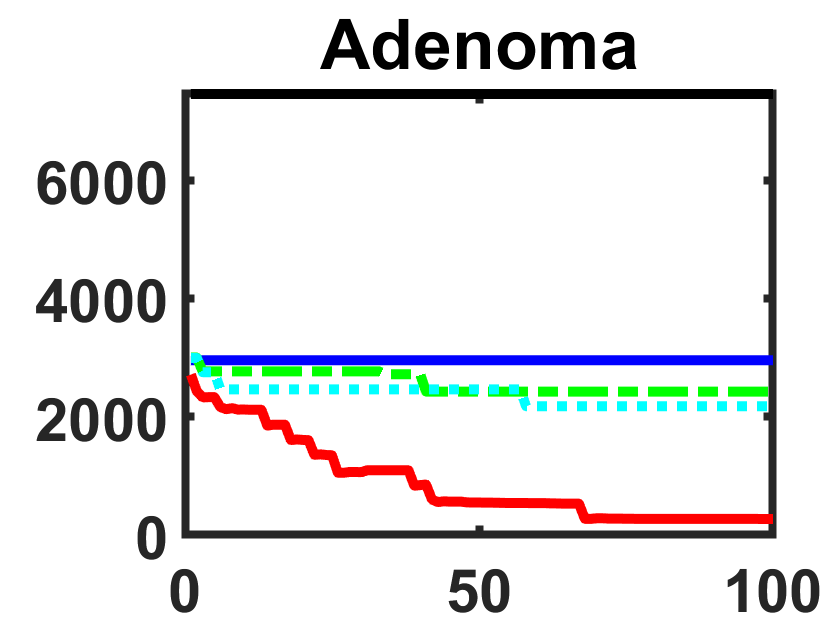}
  \end{minipage}
  \begin{minipage}[b]{0.1612\textwidth}
    \includegraphics[width=\textwidth]{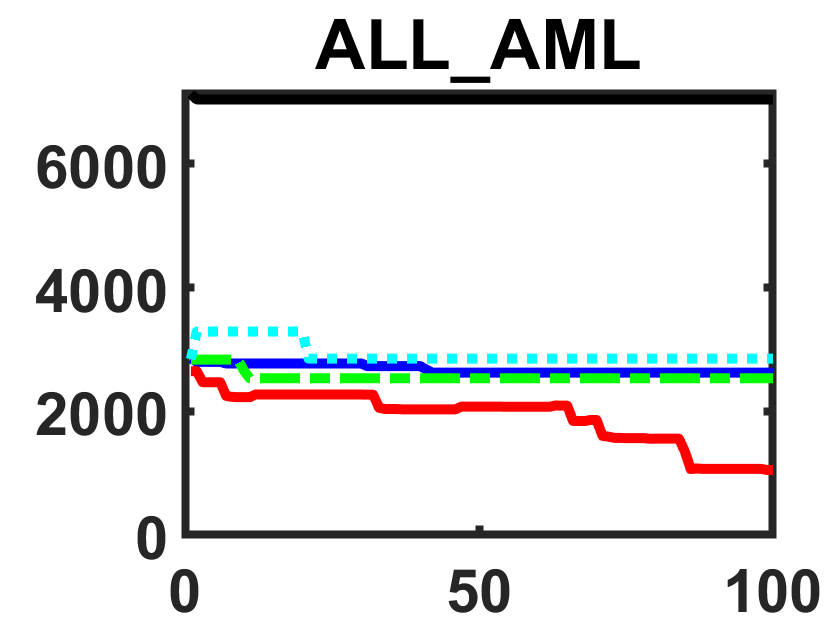}
  \end{minipage}
  \begin{minipage}[b]{0.1612\textwidth}
    \includegraphics[width=\textwidth]{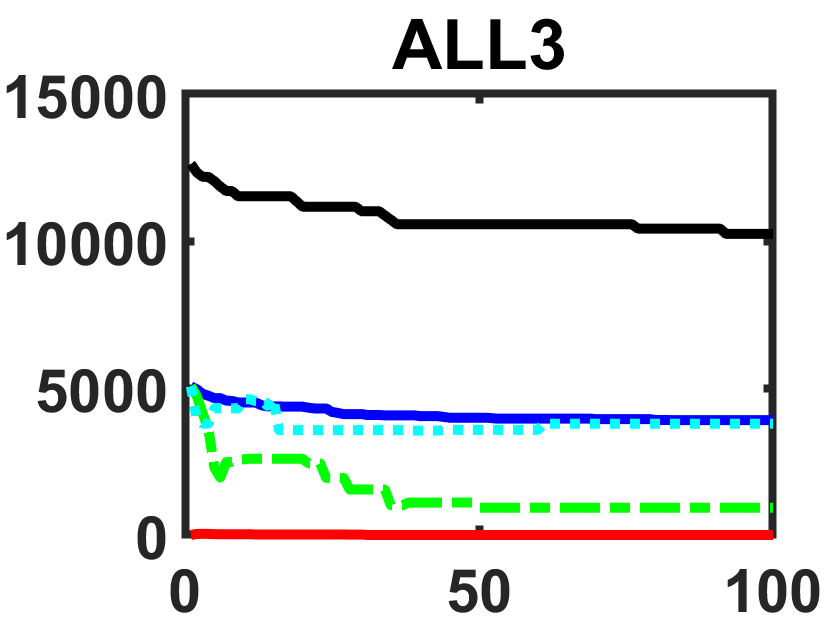}
  \end{minipage}
  \begin{minipage}[b]{0.1612\textwidth}
    \includegraphics[width=\textwidth]{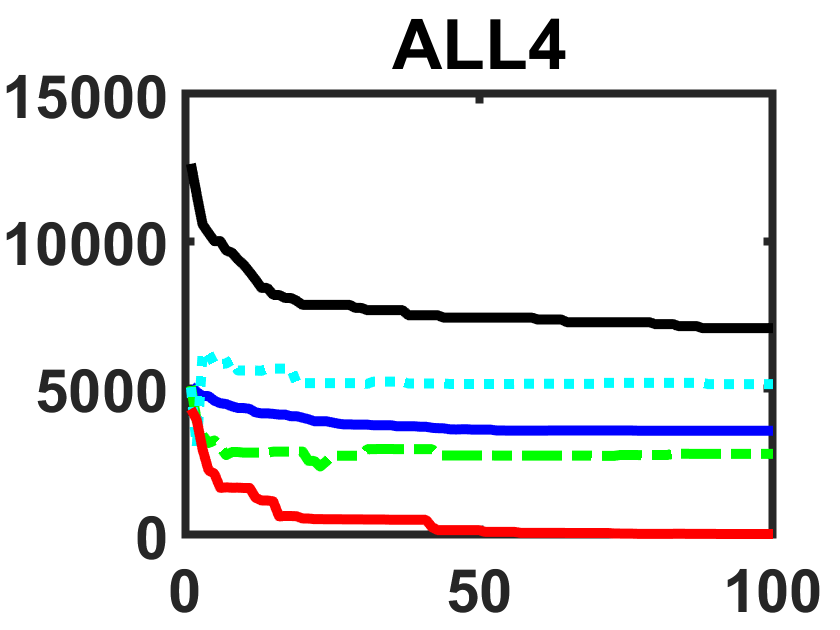}
  \end{minipage}
  \begin{minipage}[b]{0.1612\textwidth}
    \includegraphics[width=\textwidth]{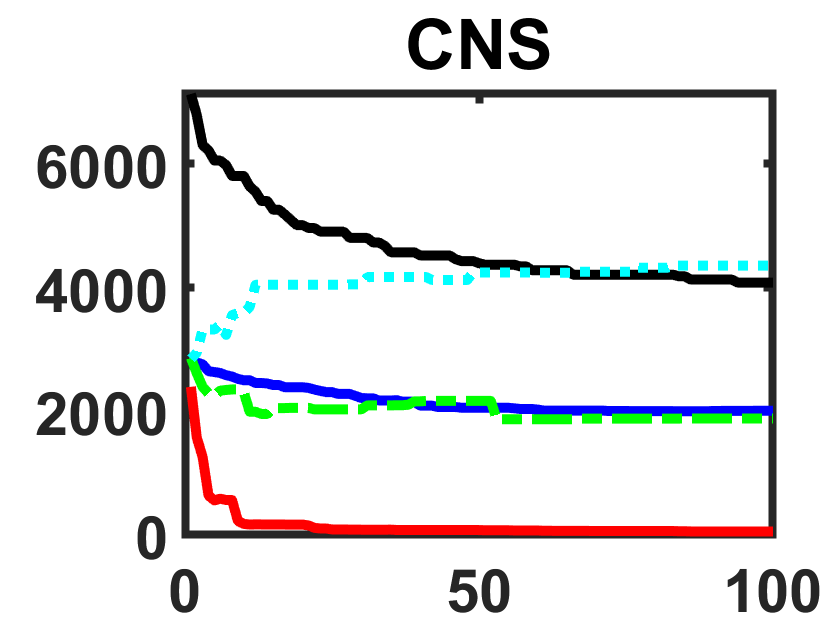}
  \end{minipage}
  \begin{minipage}[b]{0.1612\textwidth}
    \includegraphics[width=\textwidth]{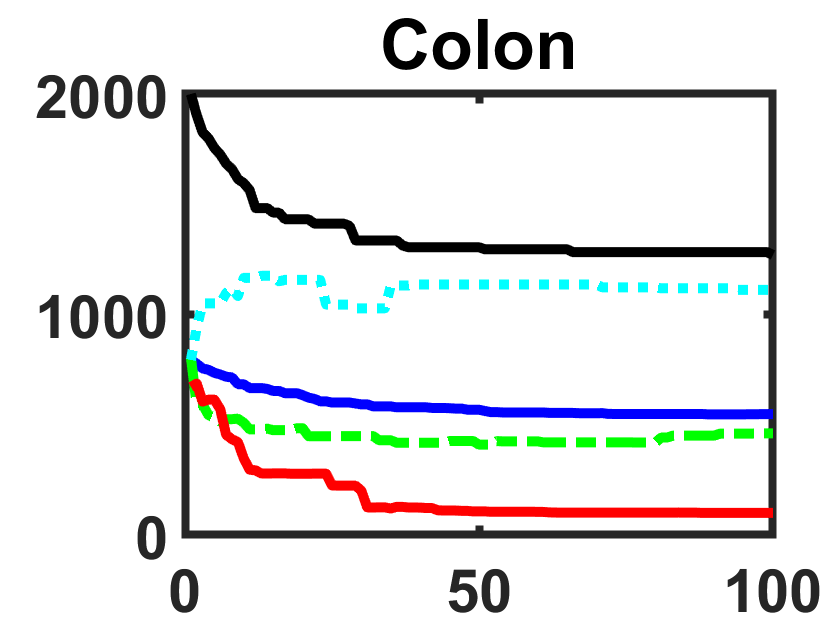}
  \end{minipage}
    \centering
  \begin{minipage}[b]{0.1612\textwidth}
    \includegraphics[width=\textwidth]{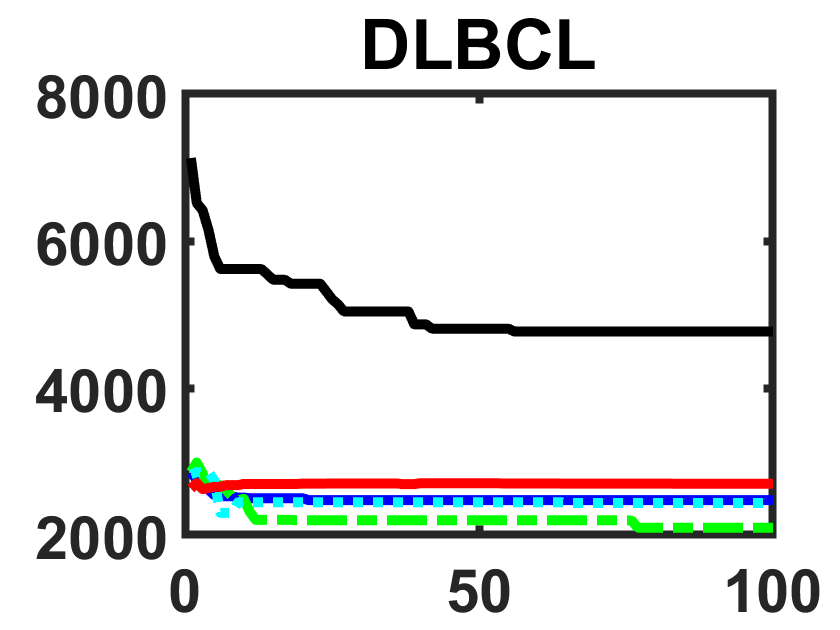}
  \end{minipage}
  \begin{minipage}[b]{0.1612\textwidth}
    \includegraphics[width=\textwidth]{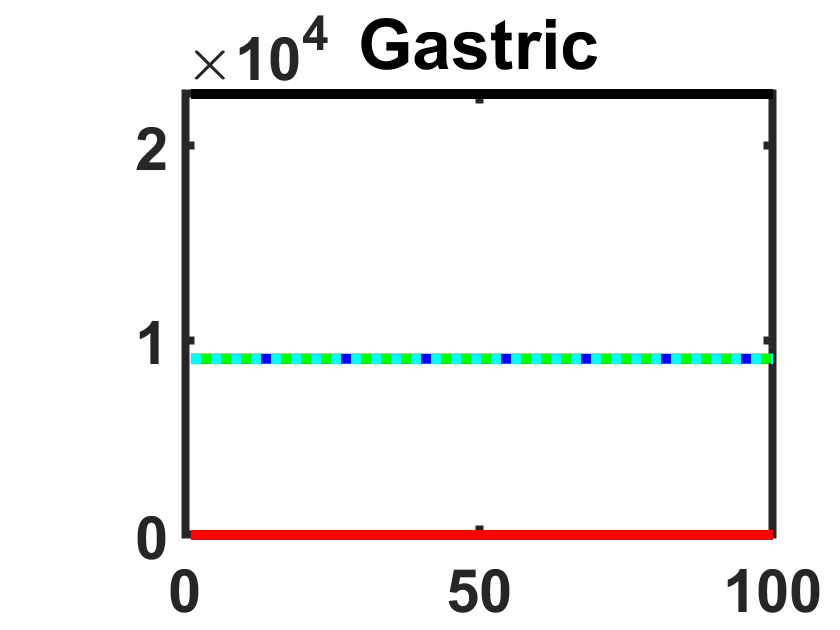}
  \end{minipage}
  \begin{minipage}[b]{0.1612\textwidth}
    \includegraphics[width=\textwidth]{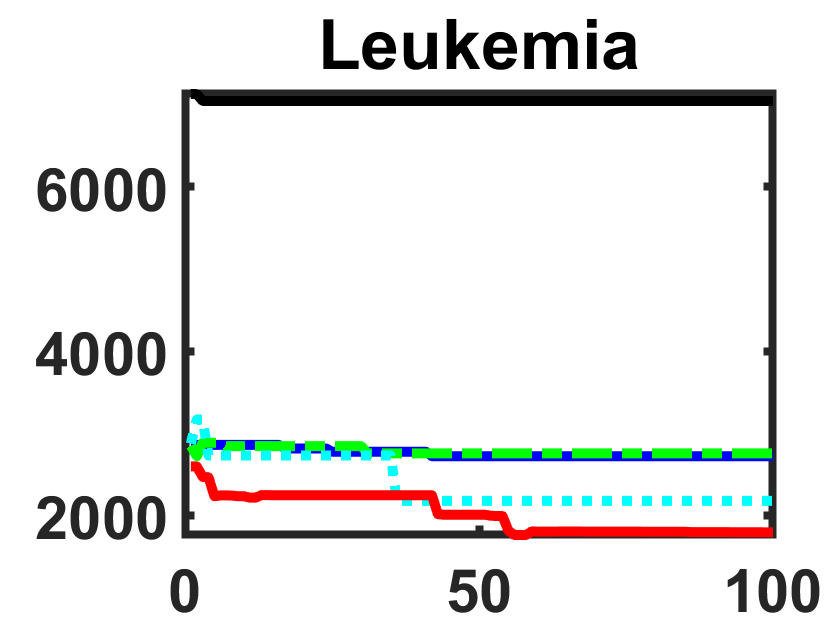}
  \end{minipage}
  \begin{minipage}[b]{0.1612\textwidth}
    \includegraphics[width=\textwidth]{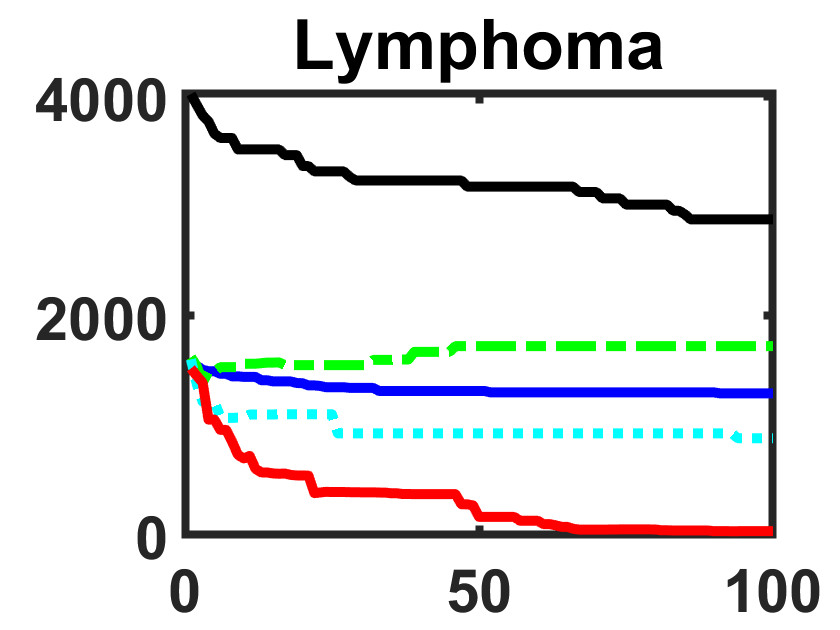}
  \end{minipage}
  \begin{minipage}[b]{0.1612\textwidth}
    \includegraphics[width=\textwidth]{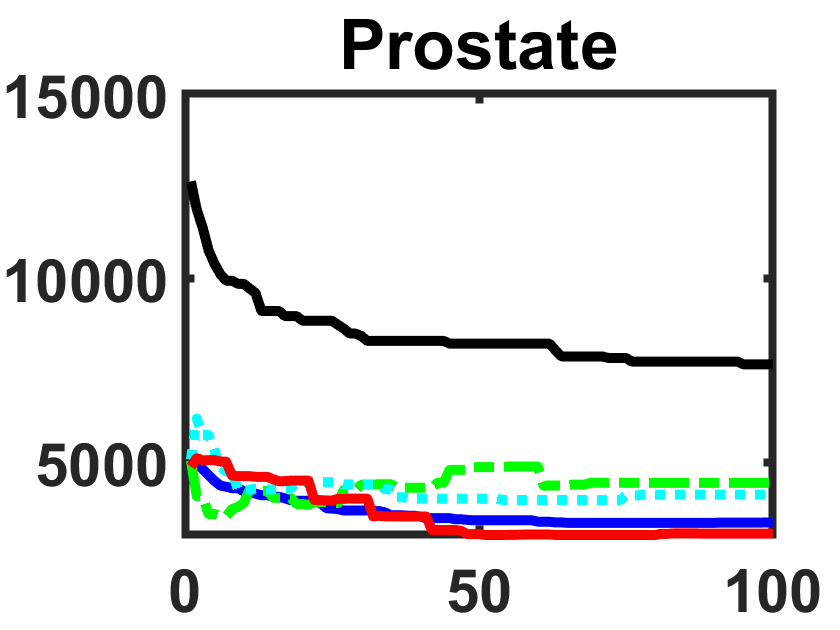}
  \end{minipage}
  \begin{minipage}[b]{0.1612\textwidth}
    \includegraphics[width=\textwidth]{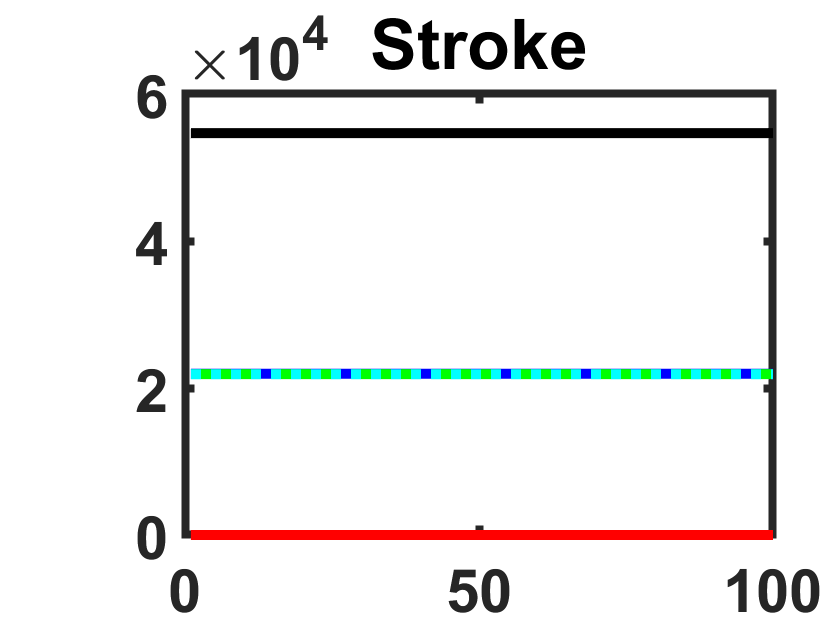}
  \end{minipage}
  \begin{minipage}[b]{0.20\textwidth}
    \includegraphics[width=\textwidth]{bio/label.png}
  \end{minipage}
  \caption{Convergence Curves of Bio-stimulated EC Algorithms in Terms of the Size of Feature Subset}
\end{figure*}

Furthermore, our proposed method demonstrates superior accuracy and computational efficiency compared to bio-inspired heuristic methods, as indicated in Table XIV. It is evident that the MEL algorithm exhibits slightly lower runtimes than the FOA algorithm for the Leukemia and Stroke datasets. On average, the MEL algorithm requires only half the time to run compared to the FOA algorithm across all 12 datasets. Additionally, when compared to the second-ranked WOA algorithm, the MEL algorithm improves runtime efficiency by approximately 29.1\%. In summary, the MEL algorithm not only achieves higher accuracy but also demonstrates improved computational efficiency compared to bio-inspired heuristic methods.

\subsection{Comparison with Physics-based Heuristic Methods}
TABLE XV presents the classification accuracy results of the physics-based method, and Figure 11 in Appendix-C displays the convergence curves for each dataset. It is evident that MEL achieved the best performance on 10 out of the 12 datasets. However, we also observed that the GSA method showed competitiveness, as its convergence curves on the DLBCL, Leukemia, and Prostate datasets exhibited search abilities comparable to those of MEL. TABLE XVI reports the results of the physics-based method on subset size, and Figure 12 in Appendix-C depicts the convergence curves for each dataset. Notably, GSA achieved the smallest feature subset size on the DLBCL and Prostate datasets.

\begin{table}[!htb]
\centering
\resizebox{87mm}{!}{
\begin{tabular}{ l l l l l l }
\hline
\multirow{2}{*}{\textbf{Dataset}} & \textbf{SA}	&	\textbf{HS}	&	\textbf{GSA}	&	\textbf{MVO}	&	\textbf{MEL (Ours)} \\ 
 \cline{2-6}
   \textbf{ } & \textbf{Mean $\pm$ Std} & \textbf{Mean $\pm$ Std}  & \textbf{Mean $\pm$ Std} & \textbf{Mean $\pm$ Std}  & \textbf{Mean $\pm$ Std} \\
\hline
Adenoma	&	0.9750	±	0.0000	&	0.9750	±	0.0000	&	0.9750	±	0.0000	&	0.9750	±	0.0000	&	\textbf{0.9975}	±	0.0079	\\ \hline
ALL$\_$AML	&	0.9867	±	0.0000	&	0.9893	±	0.0056	&	0.9880	±	0.0042	&	0.9867	±	0.0042	&	\textbf{0.9907}	±	0.0064	\\ \hline
ALL3	&	0.8000	±	0.0113	&	0.8200	±	0.0057	&	0.8272	±	0.0077	&	0.8176	±	0.0063	&	\textbf{0.8528}	±	0.0137	\\ \hline
ALL4	&	0.8109	±	0.0091	&	0.8472	±	0.0063	&	0.8601	±	0.0240	&	0.8456	±	0.0149	&	\textbf{0.9035}	±	0.0219	\\ \hline
CNS	&	0.7383	±	0.0261	&	0.7967	±	0.0105	&	0.8233	±	0.0251	&	0.7900	±	0.0263	&	\textbf{0.8450}	±	0.0223	\\ \hline
Colon	&	0.8697	±	0.0095	&	0.9117	±	0.0104	&	0.9067	±	0.0124	&	0.8992	±	0.0078	&	\textbf{0.9282}	±	0.0136	\\ \hline
DLBCL	&	0.9806	±	0.0066	&	0.9900	±	0.0053	&	\textbf{0.9913}	±	0.0060	&	0.9875	±	0.0000	&	0.9873	±	0.0004	\\ \hline
Gastric	&	\textbf{1.0000}	±	0.0000	&	\textbf{1.0000}	±	0.0000	&	\textbf{1.0000}	±	0.0000	&	\textbf{1.0000}	±	0.0000	&	\textbf{1.0000}	±	0.0000	\\ \hline
Leukemia	&	0.9880	±	0.0042	&	0.9920	±	0.0069	&	\textbf{0.9933}	±	0.0070	&	0.9893	±	0.0056	&	0.9917	±	0.0071	\\ \hline
Lymphoma	&	0.8822	±	0.0150	&	0.9444	±	0.0157	&	0.9511	±	0.0230	&	0.9378	±	0.0175	&	\textbf{0.9756}	±	0.0164	\\ \hline
Prostate	&	0.8605	±	0.0110	&	0.8873	±	0.0063	&	0.8967	±	0.0105	&	0.8840	±	0.0094	&	\textbf{0.9004}	±	0.0259	\\ \hline
Stroke	&	\textbf{1.0000}	±	0.0000	&	\textbf{1.0000}	±	0.0000	&	\textbf{1.0000}	±	0.0000	&	\textbf{1.0000}	±	0.0000	&	\textbf{1.0000}	±	0.0000	\\ \hline
\textbf{Average}	&	0.9077	±	0.0077	&	0.9295	±	0.0061	&	0.9344	±	0.0100	&	0.9261	±	0.0077	&	\textbf{0.9477}	±	0.0113	\\ \hline
\end{tabular}}
\caption{Comparison with Physics-based Heuristic Methods on Accuracy}
\end{table}

\begin{table}[!htb]
\centering
\resizebox{87mm}{!}{
\begin{tabular}{ l l l l l l }
\hline
\multirow{2}{*}{\textbf{Dataset}} & \textbf{SA}	&	\textbf{HS}	&	\textbf{GSA}	&	\textbf{MVO}	&	\textbf{MEL (Ours)} \\ 
 \cline{2-6}
   \textbf{ } & \textbf{Mean $\pm$ Std} & \textbf{Mean $\pm$ Std}  & \textbf{Mean $\pm$ Std} & \textbf{Mean $\pm$ Std}  & \textbf{Mean $\pm$ Std} \\
\hline
Adenoma	&	3739.4	±	37.1	&	2986.4	±	38.7	&	2760.0	±	604.0	&	2986.4	±	36.0	&	\textbf{271.1}	±	844.3	\\ \hline
ALL$\_$AML	&	3560.6	±	35.5	&	2849.7	±	21.2	&	2302.4	±	723.7	&	2802	±	48.9	&	\textbf{1050.8}	±	1281.9	\\ \hline
ALL3	&	6309.8	±	55.3	&	5017.1	±	57.3	&	3005.3	±	279.8	&	5089.2	±	73.0	&	\textbf{1.0}	±	0.0	\\ \hline
ALL4	&	6292.0	±	50.9	&	5059.7	±	48.8	&	3056.9	±	264.4	&	5151.5	±	83.8	&	\textbf{34.1}	±	24.9	\\ \hline
CNS	&	3561.8	±	57.2	&	2845.3	±	27.7	&	1996.7	±	54.4	&	2873.2	±	38.8	&	\textbf{57.2}	±	74.0	\\ \hline
Colon	&	1000.9	±	9.4	&	799.1	±	31.5	&	756.1	±	18.2	&	808.2	±	25.3	&	\textbf{100.1}	±	266.1	\\ \hline
DLBCL	&	3571.9	±	33.9	&	2845.6	±	48.0	&	\textbf{1916.4}	±	156.9	&	2860.5	±	39.3	&	2697.8	±	146.6	\\ \hline
Gastric	&	11345.3	±	52.1	&	9063.7	±	59.1	&	9045.2	±	65.7	&	9061.5	±	73.4	&	\textbf{1.0}	±	0.0	\\ \hline
Leukemia	&	3549.8	±	23.9	&	2835.4	±	32.7	&	2280.1	±	627.8	&	2852.9	±	43.5	&	\textbf{1793.4}	±	1236.5	\\ \hline
Lymphoma	&	2016.9	±	27.5	&	1606.7	±	28.0	&	1349.7	±	56.4	&	1637.2	±	29.7	&	\textbf{35.6}	±	46.3	\\ \hline
Prostate	&	6335.7	±	60.5	&	5058.0	±	50.4	&	\textbf{2879.2}	±	202.8	&	5173.4	±	108.7	&	3055.8	±	2579.6	\\ \hline
Stroke	&	27360.8	±	58.3	&	21818.8	±	120.1	&	21877.2	±	96.1	&	21889.7	±	126.5	&	\textbf{1.0}	±	0.0	\\ \hline
\textbf{Average}	&	6553.7	±	41.8	&	5232.1	±	47.0	&	4435.4	±	262.5	&	5265.5	±	60.6	&	\textbf{758.2}	±	541.7	\\ \hline
\end{tabular}}
\caption{Comparison with Physics-based Heuristic Methods on the Subset Size}
\end{table}

\begin{table}[!htb]
\centering
\resizebox{87mm}{!}{
\begin{tabular}{ l l l l l l }
\hline
\multirow{2}{*}{\textbf{Dataset}} & \textbf{SA}	&	\textbf{HS}	&	\textbf{GSA}	&	\textbf{MVO}	&	\textbf{MEL (Ours)} \\ 
 \cline{2-6}
   \textbf{ } & \textbf{Mean $\pm$ Std} & \textbf{Mean $\pm$ Std}  & \textbf{Mean $\pm$ Std} & \textbf{Mean $\pm$ Std}  & \textbf{Mean $\pm$ Std} \\
\hline
Adenoma	&	\textbf{4.2}	±	1.0	&	75.6	±	1.6	&	70.9	±	1.8	&	70.0	±	1.0	&	55.5	±	1.9	\\ \hline
ALL$\_$AML	&	\textbf{5.4}	±	1.6	&	95.9	±	21.2	&	81.6	±	2.6	&	86.9	±	2.3	&	66.2	±	5.8	\\ \hline
ALL3	&	\textbf{10.7}	±	0.5	&	194.2	±	57.3	&	144.0	±	6.5	&	183.3	±	2.8	&	62.9	±	2.3	\\ \hline
ALL4	&	\textbf{8.2}	±	0.6	&	151.3	±	48.8	&	116.5	±	5.7	&	143.3	±	1.3	&	66.3	±	5.6	\\ \hline
CNS	&	\textbf{4.4}	±	0.1	&	86.8	±	27.7	&	75.7	±	1.1	&	82.0	±	1.3	&	53.1	±	0.9	\\ \hline
Colon	&	\textbf{3.1}	±	0.2	&	60.8	±	31.5	&	58.2	±	1.0	&	58.4	±	1.0	&	44.9	±	1.0	\\ \hline
DLBCL	&	\textbf{5.4}	±	0.4	&	97.5	±	48.0	&	82.9	±	2.4	&	92.6	±	1.8	&	72.3	±	0.8	\\ \hline
Gastric	&	\textbf{8.9}	±	0.5	&	176.0	±	59.1	&	30.2	±	0.4	&	153.1	±	2.2	&	74.4	±	1.0	\\ \hline
Leukemia	&	\textbf{4.9}	±	0.2	&	95.4	±	32.7	&	80.2	±	1.5	&	89.3	±	1.7	&	66.8	±	6.1	\\ \hline
Lymphoma	&	\textbf{3.3}	±	0.1	&	65.7	±	28.0	&	61.4	±	1.1	&	62.0	±	1.5	&	48.2	±	1.3	\\ \hline
Prostate	&	\textbf{8.7}	±	0.4	&	162.0	±	50.4	&	127.4	±	5.7	&	154.4	±	3.0	&	98.0	±	19.0	\\ \hline
Stroke	&	\textbf{10.9}	±	0.6	&	244.5	±	120.1	&	117.8	±	2.2	&	194.1	±	2.2	&	115.7	±	1.3	\\ \hline
\textbf{Average}	&	\textbf{6.5}	±	0.5	&	125.5	±	43.9	&	87.2	±	2.7	&	114.1	±	1.9	&	68.7	±	3.9	\\ \hline
\end{tabular}}
\caption{Comparison with Physics-based Heuristic Methods on Running Time.}
\end{table}

The running times of physics-based heuristic methods are compared in TABLE XVII. Overall, the physics-based approach demonstrates fast execution. Notably, SA stands out as it runs approximately one-tenth the time of MEL (6.5 ± 0.5). However, it is important to note that SA exhibits the lowest average classification accuracy among all the algorithms, indicating that its fast search strategy sacrifices some classification performance. Taking into account factors such as algorithm complexity, accuracy, and runtime, we can conclude that MEL is an efficient and effective method. It strikes a balance between accuracy and computational efficiency, making it a favorable choice in practice.
 
\subsection{Comparison with Recently Published Evolutionary Methods}
The provided tables (TABLE XVIII - XX) present a detailed comparison of the performance of several recently published evolutionary methods, including SaWDE \cite{wang2022self}, FWPSO \cite{wang2022feature}, DENCA \cite{hancer2023evolutionary}, VGS-MOEA \cite{cheng2022variable}, PSO-EMT \cite{chen2020evolutionary} and the proposed MEL in this study. The analysis focuses on three key evaluation metrics: accuracy, the size of the selected feature subset and the running time across various datasets. For a fair comparison with other algorithms, we re-run the experiment of our MEL algorithm on MATLAB R2023b with a 13th i7-13700KF CPU and 64GB of memory. It can be seen that the proposed MEL algorithm achieves the highest average accuracy of 94.55\% across all datasets, outperforming the other state-of-the-art algorithms. MEL obtains the best accuracy on 10 out of the 12 datasets. This demonstrates the effectiveness of MEL in selecting an optimal subset of features that leads to high classification performance. Regarding the size of the selected feature subset (TABLE XIX), it can be observed that FWPSO generally finds the smallest subsets, with an average size of just 2.8 features. Although small subset is preferable as it simplifies the model and improves understandability, MEL achieves a favorable trade-off between accuracy and subset size. In addition, MEL's search efficiency is also competitive, second only to SaWDE.

In summary, MEL achieves the best balance of high classification accuracy, optimal subset size selection and efficient running time compared to other evolutionary algorithms. While some methods may get higher accuracy on a few datasets, MEL performs consistently well on most datasets. The subset search of MEL effectively finds representative features to classify samples accurately within a reasonable time limit.

\begin{table}[!htb]
\centering
\resizebox{87mm}{!}{
\begin{tabular}{ l l l l l l l }
\hline \textbf{Dataset}& \textbf{SaWDE}	&	\textbf{FWPSO}& \textbf{DENCA} & \textbf{VGS-MOEA} & \textbf{PSO-EMT} & \textbf{MEL (Ours)} \\  
\hline
Adenoma	&	0.9243	&	0.8361	&	0.9492	&	0.9960	&	0.9803	&	 \textbf{1.0000} 	\\ \hline
ALL$\_$AML	&	0.9541	&	0.8465	&	0.9828	&	\textbf{0.9957}	&	0.9655	&	 0.9890 	\\ \hline
ALL3	&	0.7544	&	0.8240	&	0.7774	&	0.7997	&	0.8136	&	\textbf{0.8480 }	\\ \hline
ALL4	&	0.6977	&	0.8038	&	0.7703	&	0.8295	&	0.8670	&	 \textbf{0.8846} 	\\ \hline
CNS	&	0.6750	&	0.7417	&	0.6109	&	0.7879	&	0.7615	&	 \textbf{0.8300} 	\\ \hline
Colon	&	0.7997	&	0.8123	&	0.8514	&	0.9183	&	0.8706	&	 \textbf{0.9265} 	\\ \hline
DLBCL	&	0.8990	&	0.8725	&	0.9354	&	0.9707	&	0.9685	&	 \textbf{0.9874} 	\\ \hline
Gastric	&	\textbf{1.0000}	&	\textbf{1.0000}	&	\textbf{1.0000} 	&	-	&	\textbf{1.0000}	&	 \textbf{1.0000} 	\\ \hline
Leukemia	&	0.9429	&	0.8663	&	0.9721	&	\textbf{0.9935}	&	0.9667	&	 0.9917 	\\ \hline
Lymphoma	&	0.7933	&	0.8111	&	0.8366	&	0.9591	&	0.9344	&	 \textbf{0.9800} 	\\ \hline
Prostate	&	0.7688	&	0.7955	&	0.8316	&	0.9065	&	0.9272	&	 \textbf{0.9092} 	\\ \hline
Stroke	&	\textbf{1.0000}	&	\textbf{1.0000}	&	\textbf{1.0000} 	&	-	&	\textbf{1.0000}	&	 \textbf{1.0000} 	\\ \hline
\textbf{Average}	&	0.8508	&	0.8508	&	0.8765 	&	0.9157	&	0.9213	&	 \textbf{0.9455} 	\\ \hline
\end{tabular}}
\caption{Comparison with Recently Published Evolutionary Methods on Accuracy}
\end{table}

\begin{table}[!htb]
\centering
\resizebox{87mm}{!}{
\begin{tabular}{ l l l l l l l }
\hline \textbf{Dataset}& \textbf{SaWDE}	&	\textbf{FWPSO}& \textbf{DENCA} & \textbf{VGS-MOEA} & \textbf{PSO-EMT} & \textbf{MEL (Ours)} \\  
\hline
Adenoma	&	2710	&	\textbf{2.6}	&	3699.8	&	136.8	&	54.6	&	6.3	\\ \hline
ALL$\_$AML	&	2867.3	&	\textbf{3.3}	&	2170.5	&	149.0	&	44.2	&	2040.9	\\ \hline
ALL3	&	3889.6	&	\textbf{1.7}	&	3701.3	&	470.6	&	257.5	&	5.4	\\ \hline
ALL4	&	4532.6	&	\textbf{4.0}	&	3746.0 	&	498.6	&	232.9	&	74.0	\\ \hline
CNS	&	773.3	&	\textbf{2.3}	&	3564.3	&	348.3	&	189.8	&	29.8	\\ \hline
Colon	&	797.8	&	\textbf{1.8}	&	605.3	&	57.0	&	48.5	&	21.0	\\ \hline
DLBCL	&	2673.9	&	\textbf{2.7}	&	2154.6	&	182.4	&	68.4	&	2252.0	\\ \hline
Gastric	&	7510.6	&	\textbf{1.0}	&	11321.7	&	-	&	\textbf{1.0}	&	\textbf{1.0}	\\ \hline
Leukemia	&	2791.6	&	\textbf{3.2}	&	2403.9	&	202.3	&	42.7	&	1310.6	\\ \hline
Lymphoma	&	1518.8	&	\textbf{2.9}	&	1231.9	&	163.7	&	36.0	&	187.8	\\ \hline
Prostate	&	5103.2	&	\textbf{6.5}	&	3976.8	&	470.1	&	216.2	&	2064.5	\\ \hline
Stroke	&	21983.9	&	\textbf{1.0}	&	27336.7	&	-	&	\textbf{1.0}	&	\textbf{1.0}	\\ \hline
\textbf{Average}	&	4762.7	&	\textbf{2.8}	&	5492.7 	&	267.9	&	99.4	&	666.2	\\ \hline
\end{tabular}}
\caption{Comparison with Recently Published Evolutionary Methods on the Subset Size}
\end{table}

\begin{table}[!htb]
\centering
\resizebox{87mm}{!}{
\begin{tabular}{ l l l l l l l }
\hline \textbf{Dataset}& \textbf{SaWDE}	&	\textbf{FWPSO}& \textbf{DENCA} & \textbf{VGS-MOEA} & \textbf{PSO-EMT} & \textbf{MEL (Ours)} \\  
\hline
Adenoma	&	\textbf{4.6}	&	66.0	&	59.7	&	48.2	&	3749.6	&	40.4 	\\ \hline
ALL$\_$AML	&	\textbf{5.1}	&	67.7	&	158.5	&	51.1	&	5305.9	&	48.7 	\\ \hline
ALL3	&	\textbf{9.1}	&	86.6	&	195.2	&	110.3	&	17406.8	&	42.9 	\\ \hline
ALL4	&	\textbf{7.4}	&	79.6	&	198.9	&	95.7	&	10887.3	&	46.1 	\\ \hline
CNS	&	\textbf{4.9}	&	66.8	&	220.3	&	45.1	&	4628.6	&	40.9 	\\ \hline
Colon	&	\textbf{4.0}	&	58.5	&	202.5	&	14.7	&	3137.3	&	35.3 	\\ \hline
DLBCL	&	\textbf{5.2}	&	68.3	&	222.7	&	53.5	&	5308.7	&	51.0 	\\ \hline
Gastric	&	\textbf{8.5}	&	89.2	&	279.4	&	-	&	7817.3	&	54.5 	\\ \hline
Leukemia	&	\textbf{5.1}	&	67.9	&	246.3	&	51.5	&	5082.7	&	48.4 	\\ \hline
Lymphoma	&	\textbf{4.2}	&	61.8	&	251.6	&	28.4	&	3241.5	&	38.0 	\\ \hline
Prostate	&	\textbf{8.0}	&	69.5	&	297.7	&	107.4	&	11480.2	&	61.0 	\\ \hline
Stroke	&	\textbf{4.3}	&	109.2	&	515.4	&	-	&	13111.5	&	80.0 	\\ \hline
\textbf{Average}	&	\textbf{5.9}	&	74.3	&	237.4 	&	60.6	&	7596.5	&	48.9 	\\ \hline
\end{tabular}}
\caption{Comparison with Recently Published Evolutionary Methods on Running Time}
\end{table}

\subsection{Experiments with Larger Data Samples}
This section analyze how MEL performs on larger datasets compared to other methods. Table XXI shows MEL achieved the highest average classification accuracy of 88.64\%, outperforming other methods on 6 out of 10 datasets. This demonstrates MEL's effectiveness in selecting informative features, even for datasets with thousands of samples and features. As shown in Table XXIII in Appendix-D, MEL selects medium-sized feature subsets compared to other algorithms. FWPSO selects the smallest subsets but attains much lower accuracy. MEL achieves an optimal balance between subset size and performance. Table XXIV in Appendix-D reveals MEL has the second best average runtime of 486.1 seconds, significantly faster than all methods except SaWDE. PSO-EMT incurs extremely high computational costs, making it impractical for real-world applications. In conclusion, MEL demonstrates excellent scalability for feature selection on large, high-dimensional datasets. It efficiently produces compact subsets that yield high accuracy. The balanced optimization of effectiveness, efficiency and subset quality exhibited by MEL makes it well-suited for real-world applications involving big data.

\begin{table}[!htb]
\centering
\resizebox{87mm}{!}{
\begin{tabular}{ l l l l l l l }
\hline \textbf{Dataset}& \textbf{SaWDE}	&	\textbf{FWPSO}& \textbf{DENCA} & \textbf{PSO-EMT } & \textbf{MTPSO} & \textbf{MEL (Ours)} \\  
\hline
BASEHOCK	&	0.8260 	&	0.5164 	&	0.8324	&	0.6751	&	0.7675 	&	\textbf{0.8839} 	\\ \hline
COIL20	&	0.9660 	&	0.6720 	&	0.9769	&	\textbf{0.9981}	&	0.9780 	&	0.9950 	\\ \hline
HAPTDataSet	&	0.8875 	&	0.5452 	&	0.9012	&	0.8731	&	0.8269 	&	\textbf{0.9315} 	\\ \hline
Isolet	&	0.8323 	&	0.1122 	&	0.8763	&	0.6947	&	0.8942 	&	\textbf{0.8790} 	\\ \hline
madelon	&	0.7087 	&	0.5548 	&	0.7347	&	0.8901	&	\textbf{0.9157} 	&	0.7661 	\\ \hline
MultipleFeaturesDigit	&	0.9220 	&	0.4254 	&	0.9115	&	0.9084	&	 0.9523 &	\textbf{0.9525} 	\\ \hline
Pancancer	&	0.8350 	&	0.7809 	&	\textbf{0.8484 }	&	-	&	0.7587 	&	0.8453 	\\ \hline
PCMAC	&	0.7615 	&	0.5065 	&	0.7738	&	0.6696	&	0.7776 	&	\textbf{0.8128} 	\\ \hline
RELATHE	&	0.7621 	&	0.5505 	&	0.7931	&	0.6626	&	\textbf{0.8314} 	&	0.8307 	\\ \hline
USPS	&	0.9630 	&	0.3029 	&	0.9647	&	-	&	0.9473 	&	\textbf{0.9674}	\\ \hline
Average	&	0.8464 	&	0.4967 	&	0.8613 	&	0.7965 	&	0.8650 	&	\textbf{0.8864} 	\\ \hline
\end{tabular}}
\caption{Accuracy Comparison on Datasets with Large Sample Size}
\end{table}

\section{Conclusion and Future Work}
In this paper, we propose a PSO-based multi-task learning method called MEL for high-dimensional feature selection. MEL evolves by continuously learning the importance of individual features through the entire population. Specifically, two subpopulations are constructed in MEL and each independently searches for the optimal feature subset using different methods. The first subpopulation ($\Vec{Sub_1}$) realizes knowledge transfer by incorporating the influence of the optimal solution from the other subpopulation ($\Vec{Sub_2}$) during its search process. $\Vec{Sub_2}$ narrows the search scope and improves its ability to identify potentially valuable features, with the help of feature importance information learned by the whole population. Knowledge is shared between the two subpopulations through multi-task learning to enhance the search ability and efficiency of the algorithm. 

We evaluate the effectiveness of our method using three metrics: accuracy, feature subset size, and algorithm running time. Extensive experiments on 12 high-dimensional genetic datasets showed that MEL can effectively improve classification accuracy while obtaining a small feature subset. MEL also showed highly competitive running time compared to 18 state-of-the-art meta-heuristic optimization algorithms and five recently published evolutionary feature selection methods. Additionally, we provided further experiments on a separate set of 10 larger sample size datasets, comparing MEL against five representative algorithms. The results demonstrated MEL's superior overall performance in classification metrics, validating its effectiveness on high-dimensional data with both few and many samples. Despite the encouraging results of our study, some directions are still worth pursuing. For example, the data discussed in this study are basically class balanced. However, class imbalance data is also very common in real life, and how to address it better still needs a lot of efforts.

\section{Acknowledgments}
This work is supported in part by the Chinese National Research Fund (NSFC) under Grant 62272050; in part by the Guangdong Key Lab of AI and Multi-modal Data Processing, United International College (UIC), Zhuhai under Grant 2020KSYS007 sponsored by Guangdong Provincial Department of Education, and in part by the Interdisciplinary Intelligence SuperComputer Center of Beijing Normal University (Zhuhai).

\bibliographystyle{IEEEtran}
\bibliography{reference}




\section*{Appendix}
\renewcommand{\thesubsection}{\Alph{subsection}}

\subsection{Representative Classic Meta-heuristic Optimization Algorithms}
To demonstrate the effectiveness of our approach, we employed a total of 18 evolutionary computation algorithms in the first phase of the experiment. Following the classification methods outlined in the literature \cite{kumar2020comparative}, we carefully selected four swarm-based methods, four nature-inspired methods, two evolutionary algorithms, four bio-stimulated methods, and four physics-based methods. These choices are depicted in Figure 10. These meta-heuristic optimization methods that we have chosen are considered to be the most classical, representative, and extensively utilized techniques in the field. In this section, we will provide a brief introduction to each of these methods.
\begin{figure*}[htp!]
  \centering
    \includegraphics[width=18cm]{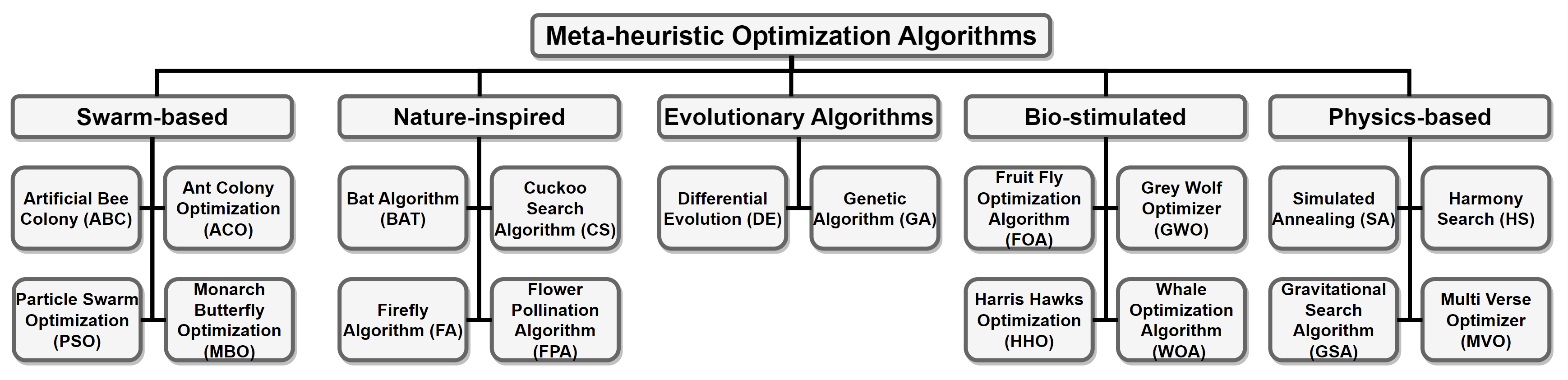}
    \caption{Representative Classic Methods: 18 Meta-heuristic Optimization Algorithms.}
    \label{fig1}
\end{figure*}
\begin{enumerate}
    \item  \textbf{Artificial Bee Colony (ABC)} \cite{karaboga2007powerful}: It simulates the biological behaviour of honeybees cooperate with each other to collect honey through individual division of labor and information exchange.
    \item  \textbf{Ant Colony Optimization (ACO)} \cite{aghdam2009text}: ACO is inspired by the foraging behavior of ant colonies to find their way around food.
    \item  \textbf{Particle Swarm Optimization (PSO)} \cite{shi1998modified}: The concept of PSO arose from the study of bird feeding behavior. By simulating the behavior of bird flocks flying for food, the birds collaborate with one another to achieve the group's optimal goal.
    \item \textbf{Monarch Butterfly Optimization (MBO)} \cite{wang2019monarch}: MBO algorithm simulates the migration and adaptation behavior of monarch butterfly.
    \item  \textbf{Bat Algorithm (BAT)} \cite{yang2010new}: BAT algorithm is a random search algorithm that simulates bats in nature using a kind of sonar to detect prey and avoid obstacles.
    \item  \textbf{Cuckoo Search Algorithm (CS)} \cite{yang2009cuckoo}: CS algorithm is an optimization algorithm by simulating the incubation parasitism of cuckoos and Levy flight mechanism.
    \item  \textbf{Firefly Algorithm (FA)} \cite{yang2010firefly}: FA is a heuristic algorithm for information exchange, mutual attraction and danger warning based on flashing behavior of fireflies.
    \item \textbf{Flower Pollination Algorithm (FPA)} \cite{yang2014flower}: FPA simulates the process of plant cross-pollination by birds and bees using Levy's flight mechanism.
    \item  \textbf{Differential Evolution (DE)} \cite{storn1997differential}: DE is an intelligent optimization algorithm generated by the cooperation and competition between individuals within a group.
    \item  \textbf{Genetic Algorithm (GA)} \cite{huang2006ga}: GA is an optimization algorithm that simulates the natural selection and genetic mechanism of biological evolution.
    \item  \textbf{Fruit Fly Optimization Algorithm (FOA)} \cite{pan2012new}: FOA mimics the process of fruit flies that use their keen sense of smell and vision to hunt.
    \item  \textbf{Grey Wolf Optimizer (GWO)} \cite{mirjalili2014grey}: GWO is an optimization search method inspired by the prey hunting activities of grey wolves, which has strong convergence performance, few parameters and easy implementation. 
    \item  \textbf{Harris Hawks Optimization (HHO)} \cite{heidari2019harris}: HHO is an intelligent optimization algorithm that simulates the predatory behavior of the Harris Hawk.
    \item  \textbf{Whale Optimization Algorithm (WOA)} \cite{mirjalili2016whale}: WOA mimics the hunting behavior of whales in nature and has the advantages of being easy to implement and having fewer parameters.
    \item  \textbf{Simulated Annealing (SA)} \cite{kirkpatrick1983optimization}: Simulated annealing algorithm comes from solid annealing principle and is based on Monte-Carlo iterative solution strategy. To avoid falling into local optimality, the search process is endowed with a time-varying probability of jumping to zero.
    \item \textbf{Harmony Search (HS)} \cite{geem2001new}: HS algorithm is a simulation of the process that musicians achieve the most beautiful harmony by repeatedly adjusting the tones of different instruments, so as to achieve the purpose of global optimization.
    \item \textbf{Gravitational Search Algorithm (GSA)} \cite{rashedi2009gsa}: GSA is an optimization method based on the law of gravitation and Newton's second law, which seeks the optimal solution through the continuous movement of gravitation between particles within a population.
    \item \textbf{Multi Verse Optimizer (MVO)} \cite{mirjalili2016multi}: MVO simulates the motion behavior of objects with high expansion rate tends to low expansion rate under the combined action of white holes, black holes and wormholes in the multiverse population, and tends to the optimal position in the search space by means of gravity.
\end{enumerate}

\subsection{Parameter Settings}
We used the toolkit\footnote{https://github.com/JingweiToo/Wrapper-Feature-Selection-Toolbox} for our experiments, and we followed the default settings provided in the toolbox for each method. In this section, we provide a detailed description of the parameters for each method, which can be referred to in Table XXII.

\begin{table}[!htb]
\centering
\resizebox{83mm}{!}{
\begin{tabular}{| l | m{5.7cm} |}
\hline
\textbf{Methods} & \textbf{Parameters}  \\
\hline
ABC & lb = 0, ub = 1, $\theta =0.6$, max\_limit = 5.\\
\hline 
ACO &  tau = 1, eta = 1, alpha = 1, beta = 0.1, rho = 0.2.\\
\hline
PSO & lb = 0, ub = 1, $\theta =0.6$, c1 = 2, c2 = 2, w = 0.9, Vmax =(ub - lb)/2.\\
\hline
MBO &  lb = 0, ub = 1, $\theta =0.6$, peri = 1.2, p = 5/12, Smax = 1, BAR = 5/12, num\_land1 = 4, beta = 1.5.\\
\hline
BAT & lb = 0, ub = 1, $\theta =0.6$, fmax = 2, fmin = 0, alpha = 0.9, gamma = 0.9, A\_max = 2, r0\_max = 1.\\ \hline
CS & lb = 0, ub = 1, $\theta = 0.6$, Pa = 0.25, alpha = 1, beta = 1.5.\\
\hline
FA & lb = 0, ub = 1, $\theta =0.6$, alpha0 = 1, beta0 = 1, gamma = 1, calpha = 0.97. \\ \hline
FPA & lb = 0, ub = 1, $\theta =0.6$, beta = 1.5, P = 0.8. \\\hline
DE & lb = 0, ub = 1, $\theta = 0.6$, CR = 0.9, F = 0.5.\\
\hline
GA (Roulette) & CR = 0.8, MR = 0.01.\\
\hline
GA (Tournament) & CR = 0.8, MR = 0.01, Tour\_size = 3.\\
\hline
FOA & lb = 0, ub = 1, $\theta =0.6$.\\ \hline
GWO & lb = 0, ub = 1, $\theta =0.6$.\\ 
\hline
HHO & lb = 0, ub = 1, $\theta =0.6$, beta = 1.5.\\\hline
WOA & lb = 0, ub = 1, $\theta =0.6$, b = 1.\\\hline
SA & c = 0.93, t0 = 100.\\\hline
HS & lb = 0, ub = 1, $\theta =0.6$, PAR = 0.05, HMCR = 0.7, bw = 0.2, NP = 20.\\\hline
GSA &  lb = 0, ub = 1, $\theta =0.6$, G0 = 100, alpha = 20.\\\hline
MVO & lb = 0, ub = 1, $\theta =0.6$, p = 6, Wmax = 1, Wmin = 0.2.\\\hline
\end{tabular}}
\caption{Parameters of Different Evolutionary Algorithms}
\end{table}

\begin{figure*}[htp!]
  \centering
  \begin{minipage}[b]{0.1612\textwidth}
    \includegraphics[width=\textwidth]{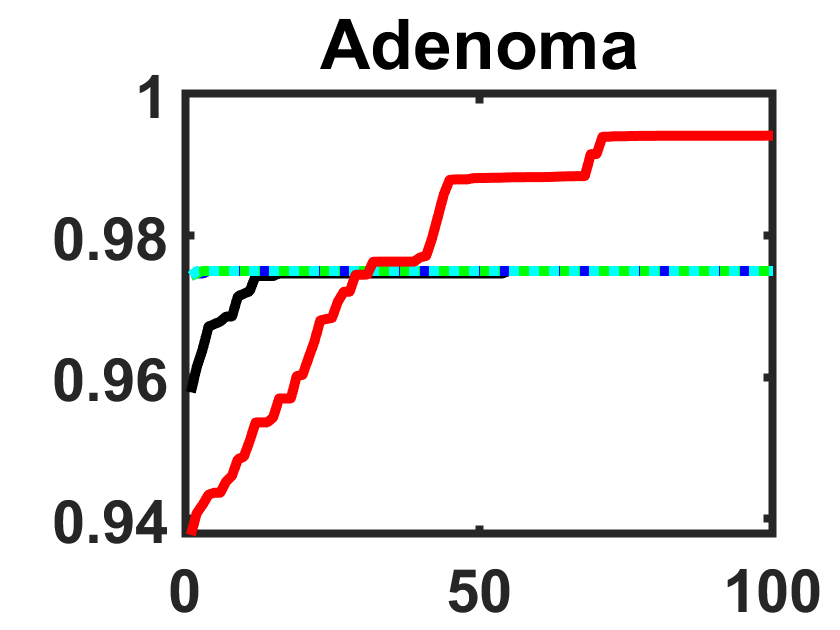}
  \end{minipage}
  \begin{minipage}[b]{0.1612\textwidth}
    \includegraphics[width=\textwidth]{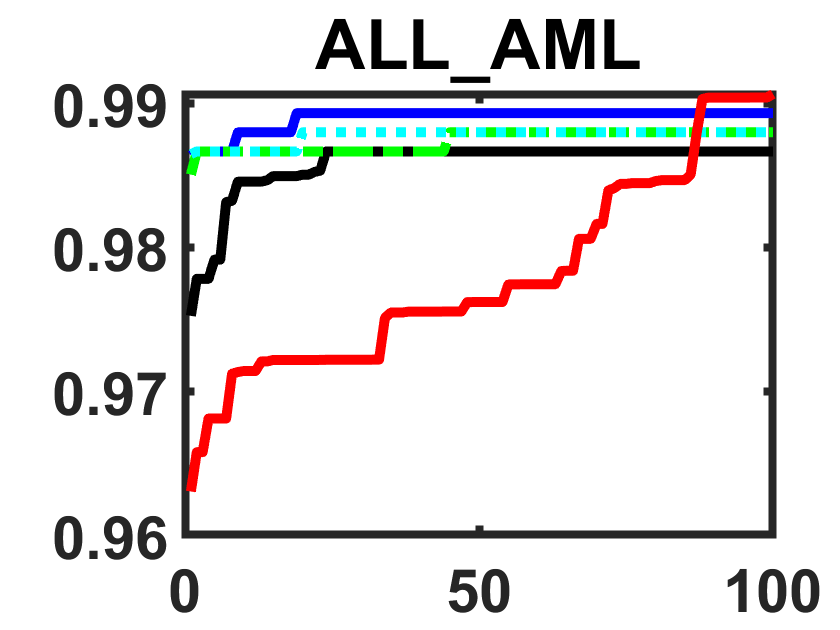}
  \end{minipage}
  \begin{minipage}[b]{0.1612\textwidth}
    \includegraphics[width=\textwidth]{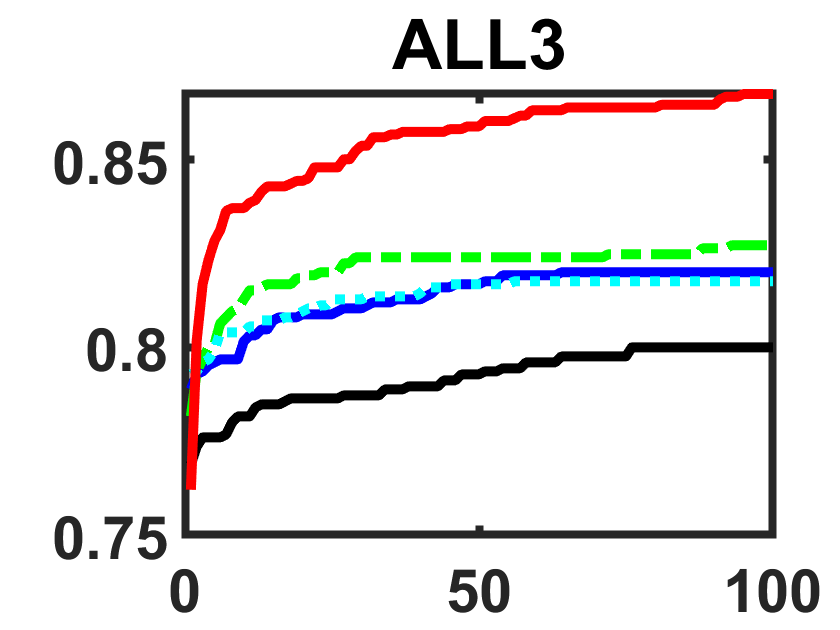}
  \end{minipage}
  \begin{minipage}[b]{0.1612\textwidth}
    \includegraphics[width=\textwidth]{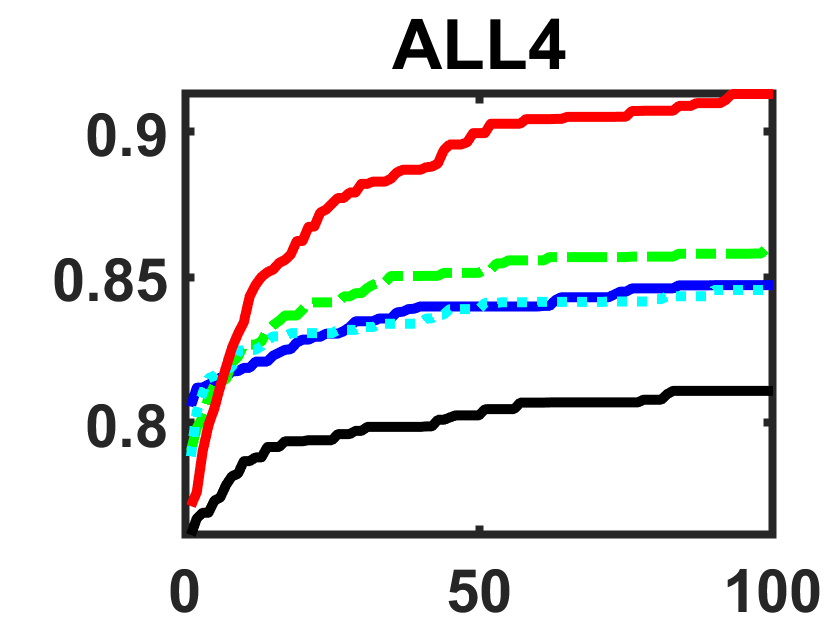}
  \end{minipage}
  \begin{minipage}[b]{0.1612\textwidth}
    \includegraphics[width=\textwidth]{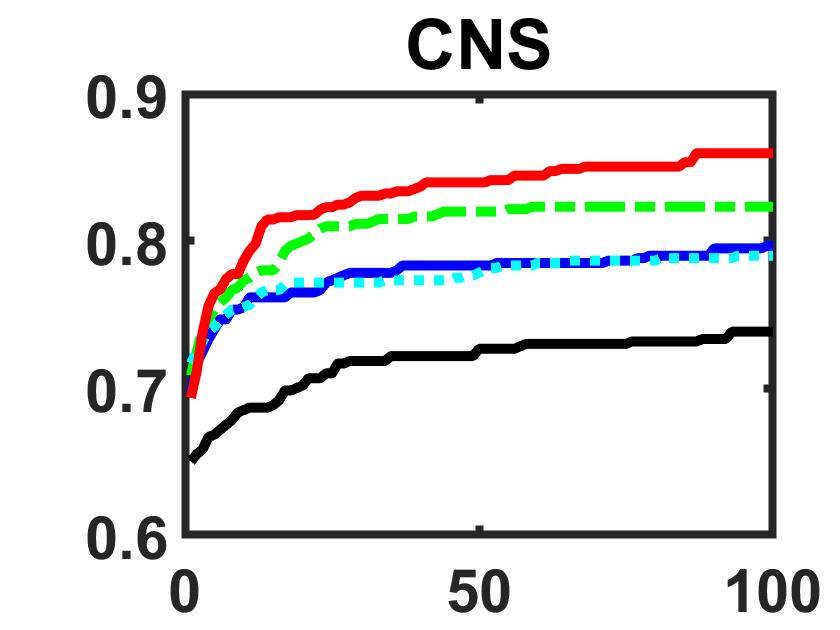}
  \end{minipage}
  \begin{minipage}[b]{0.1612\textwidth}
    \includegraphics[width=\textwidth]{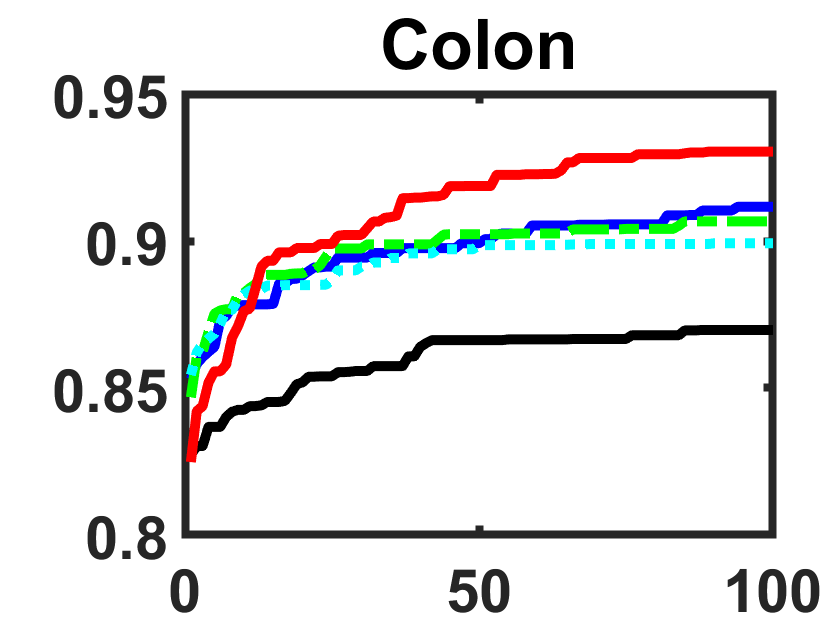}
  \end{minipage}
    \centering
  \begin{minipage}[b]{0.1612\textwidth}
    \includegraphics[width=\textwidth]{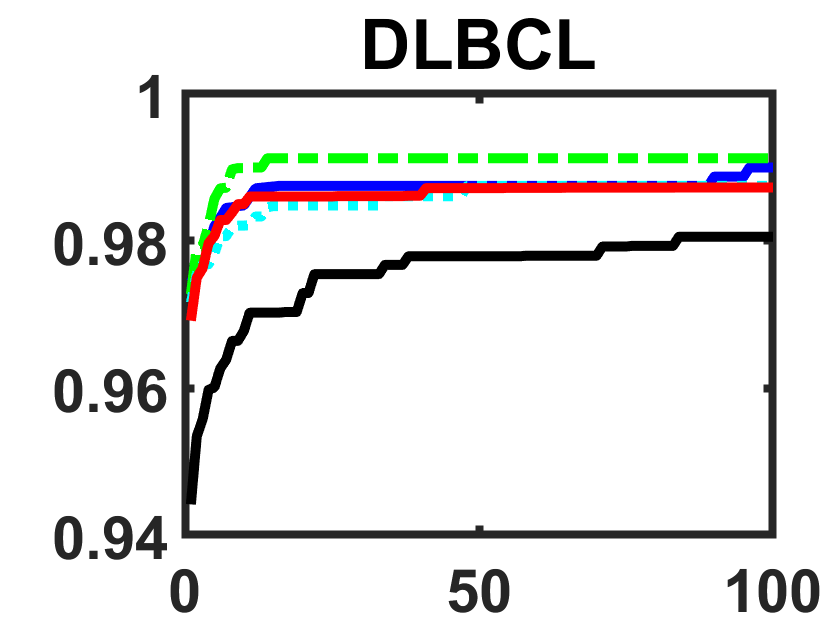}
  \end{minipage}
  \begin{minipage}[b]{0.1612\textwidth}
    \includegraphics[width=\textwidth]{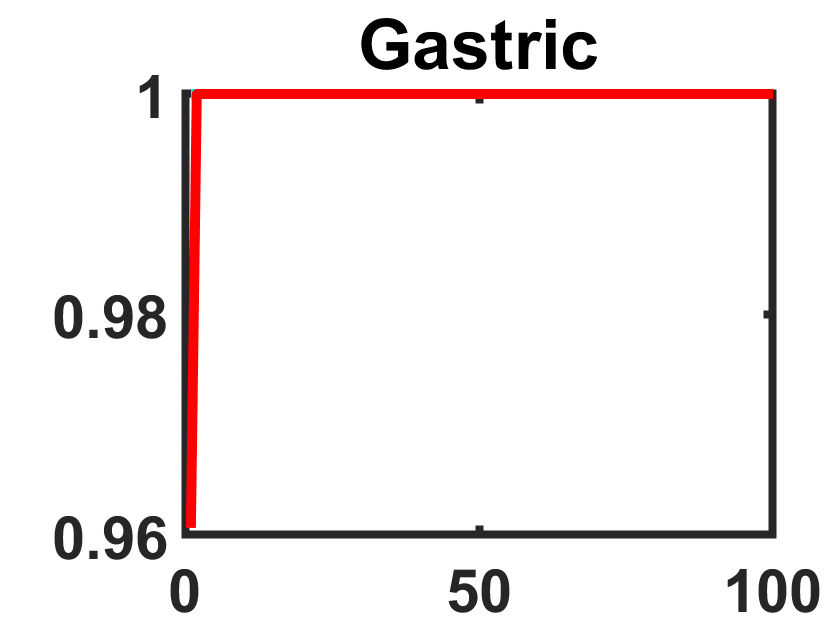}
  \end{minipage}
  \begin{minipage}[b]{0.1612\textwidth}
    \includegraphics[width=\textwidth]{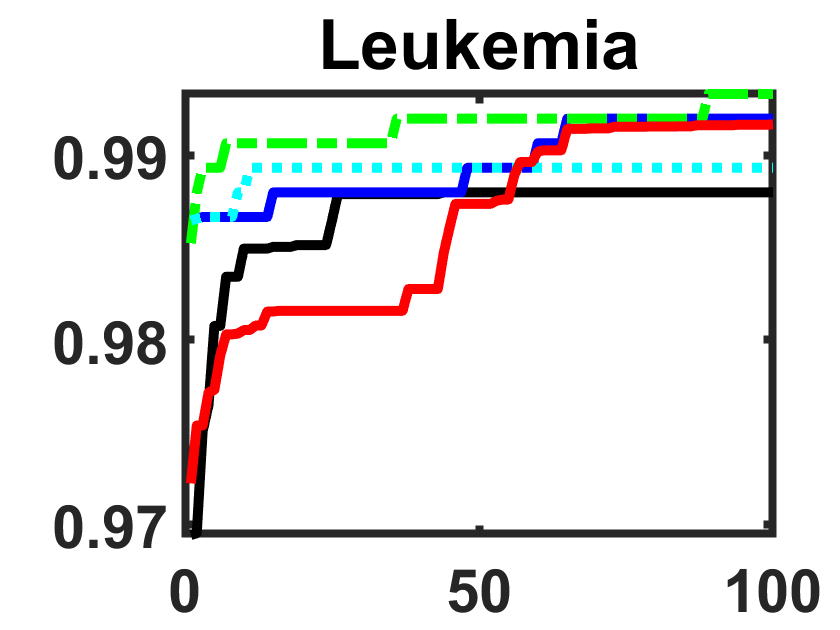}
  \end{minipage}
  \begin{minipage}[b]{0.1612\textwidth}
    \includegraphics[width=\textwidth]{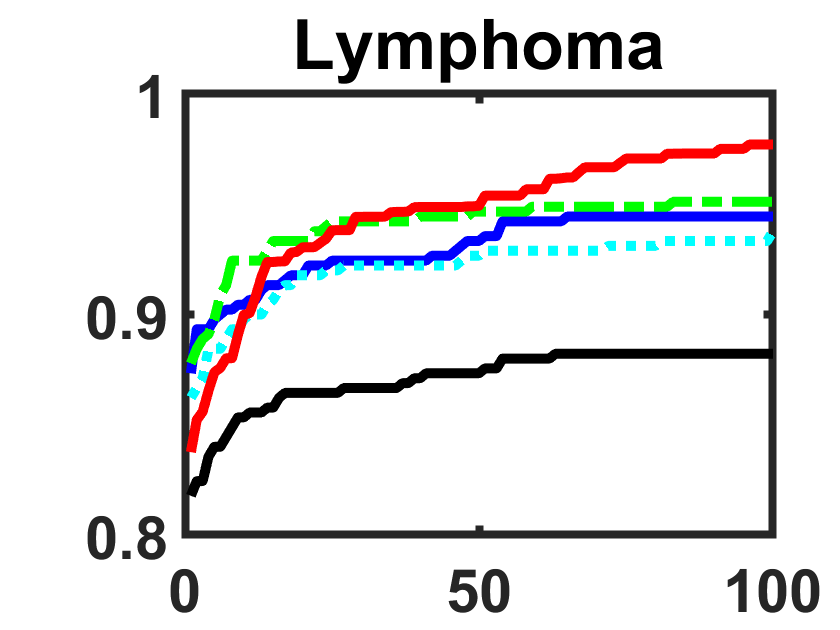}
  \end{minipage}
  \begin{minipage}[b]{0.1612\textwidth}
    \includegraphics[width=\textwidth]{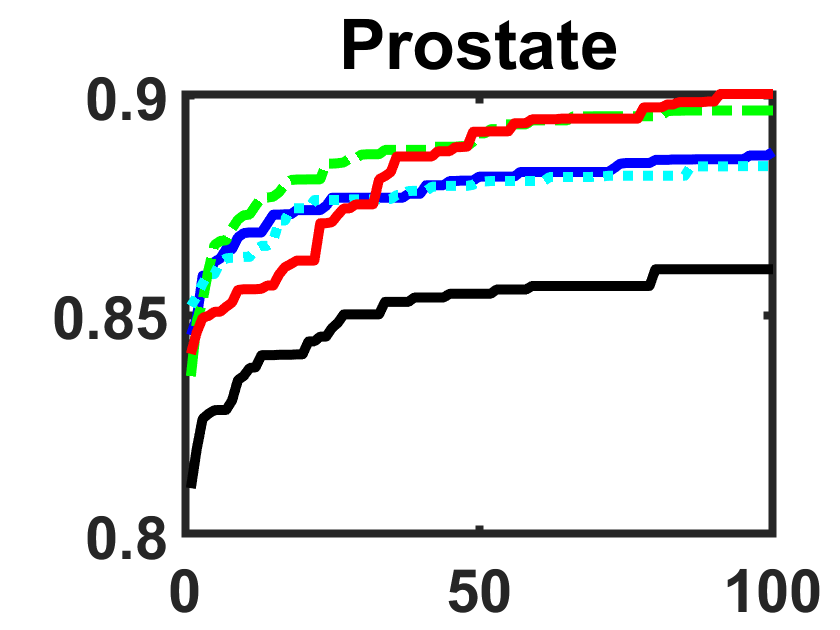}
  \end{minipage}
  \begin{minipage}[b]{0.1612\textwidth}
    \includegraphics[width=\textwidth]{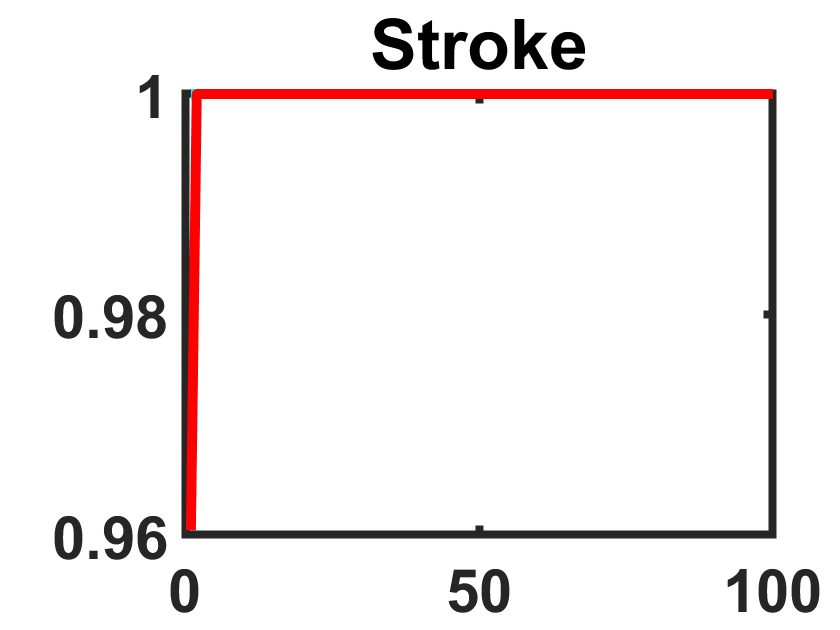}
  \end{minipage}
  \begin{minipage}[b]{0.19\textwidth}
    \includegraphics[width=\textwidth]{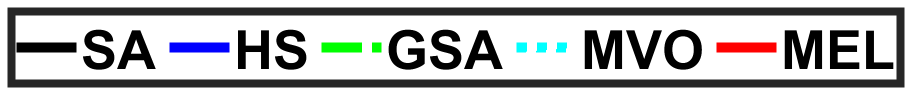}
  \end{minipage}
  \caption{Convergence Curves of Physics-based EC Algorithms in Terms of Accuracy}
\end{figure*}

\begin{figure*}[htp!]
  \centering
  \begin{minipage}[b]{0.1612\textwidth}
    \includegraphics[width=\textwidth]{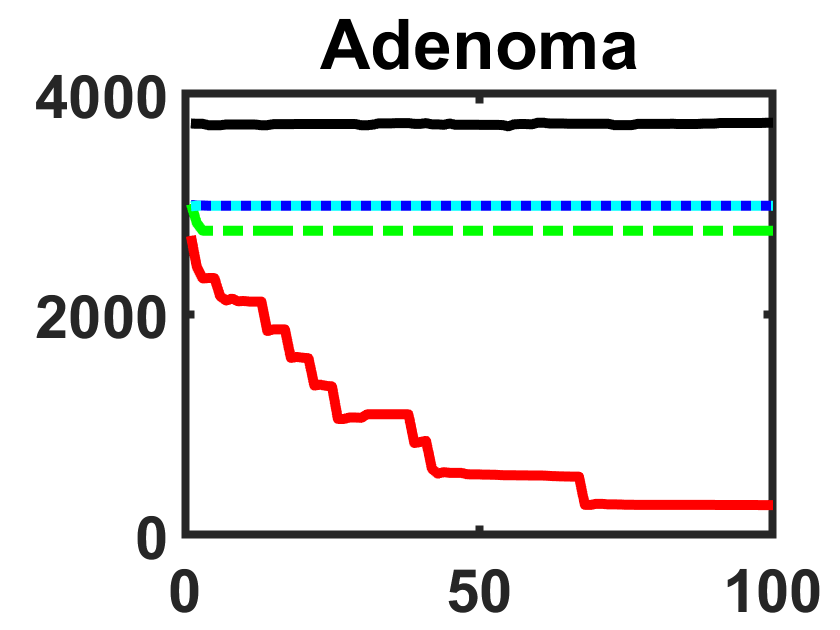}
  \end{minipage}
  \begin{minipage}[b]{0.1612\textwidth}
    \includegraphics[width=\textwidth]{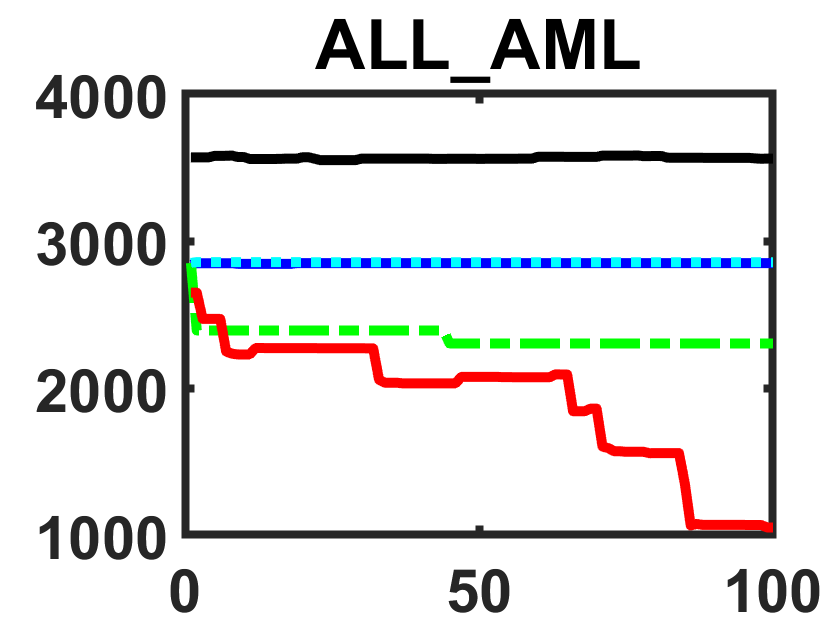}
  \end{minipage}
  \begin{minipage}[b]{0.1612\textwidth}
    \includegraphics[width=\textwidth]{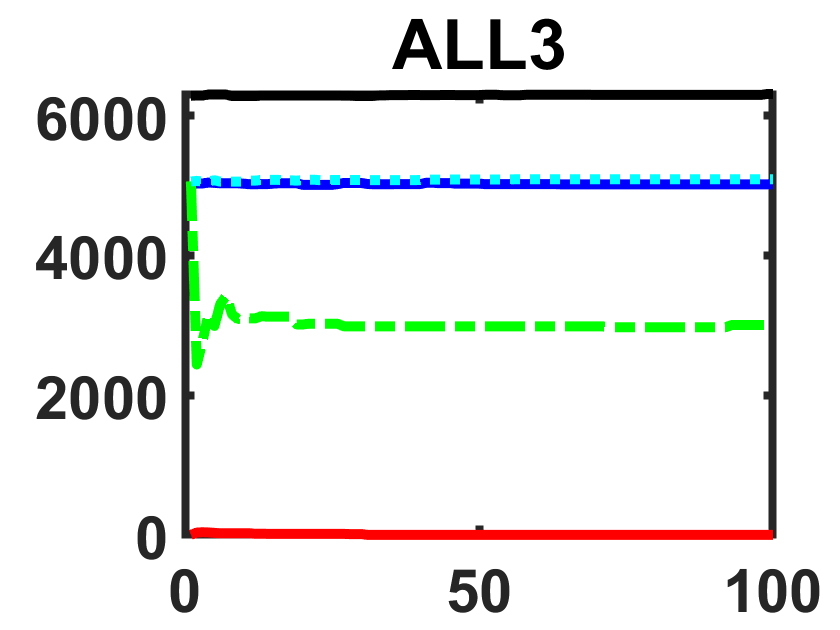}
  \end{minipage}
  \begin{minipage}[b]{0.1612\textwidth}
    \includegraphics[width=\textwidth]{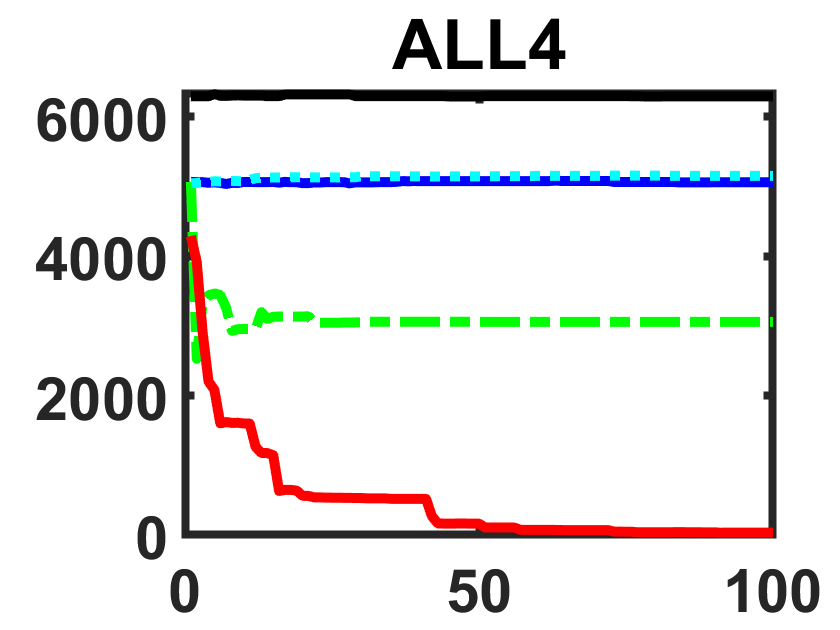}
  \end{minipage}
  \begin{minipage}[b]{0.1612\textwidth}
    \includegraphics[width=\textwidth]{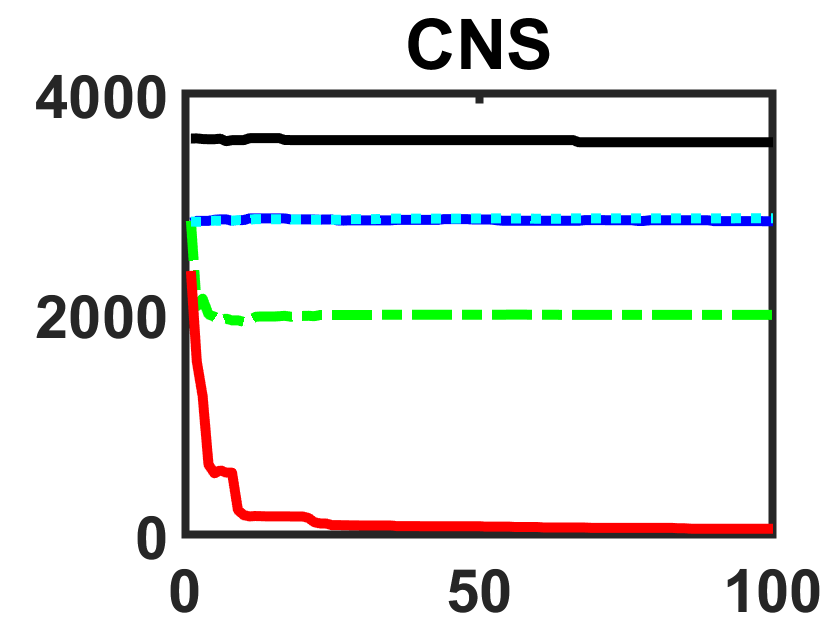}
  \end{minipage}
  \begin{minipage}[b]{0.1612\textwidth}
    \includegraphics[width=\textwidth]{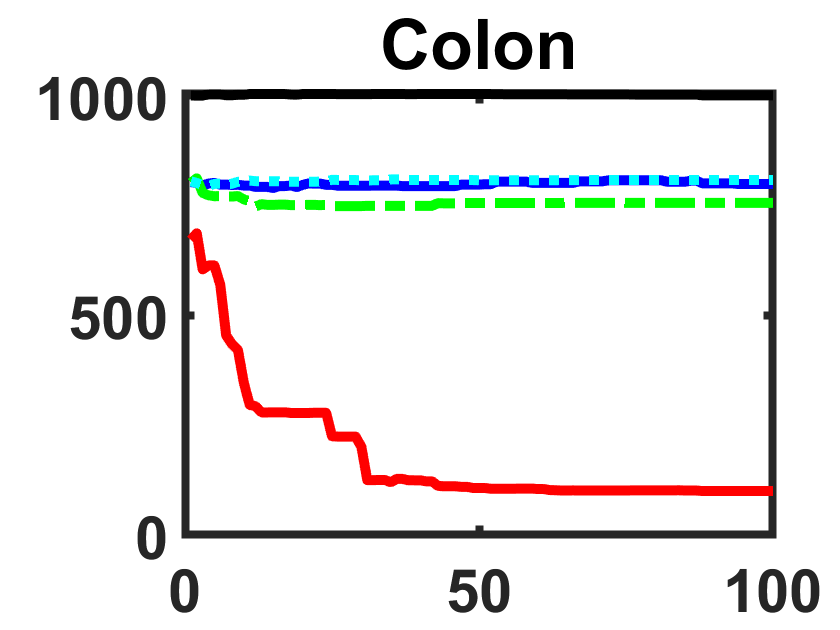}
  \end{minipage}
    \centering
  \begin{minipage}[b]{0.1612\textwidth}
    \includegraphics[width=\textwidth]{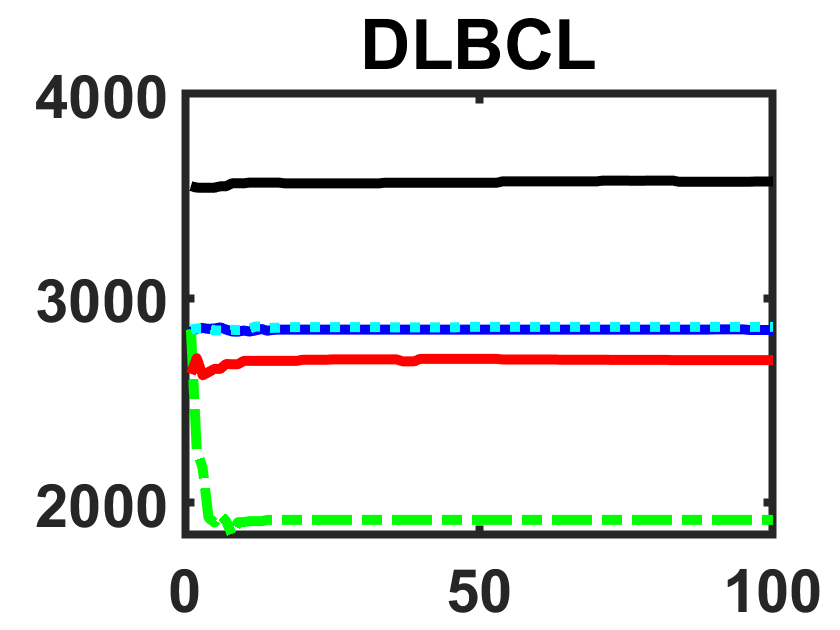}
  \end{minipage}
  \begin{minipage}[b]{0.1612\textwidth}
    \includegraphics[width=\textwidth]{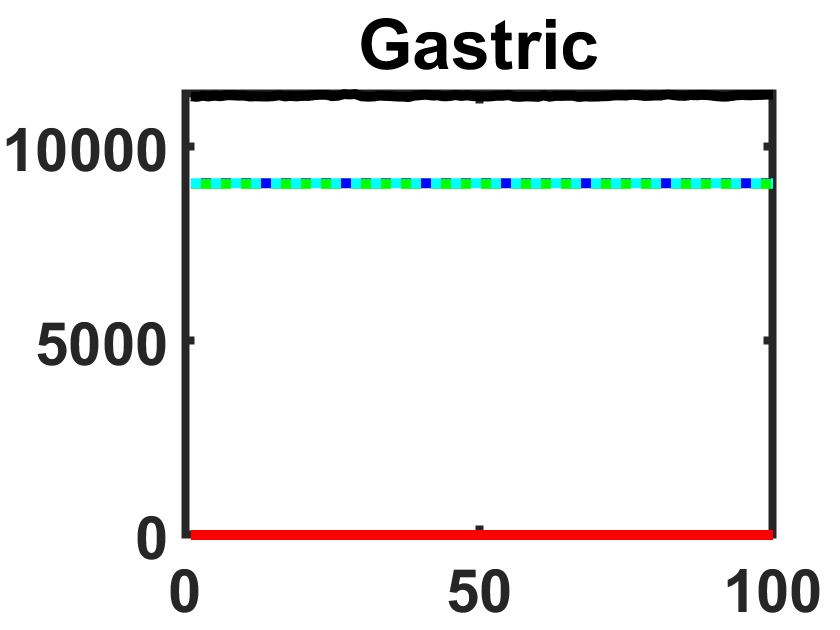}
  \end{minipage}
  \begin{minipage}[b]{0.1612\textwidth}
    \includegraphics[width=\textwidth]{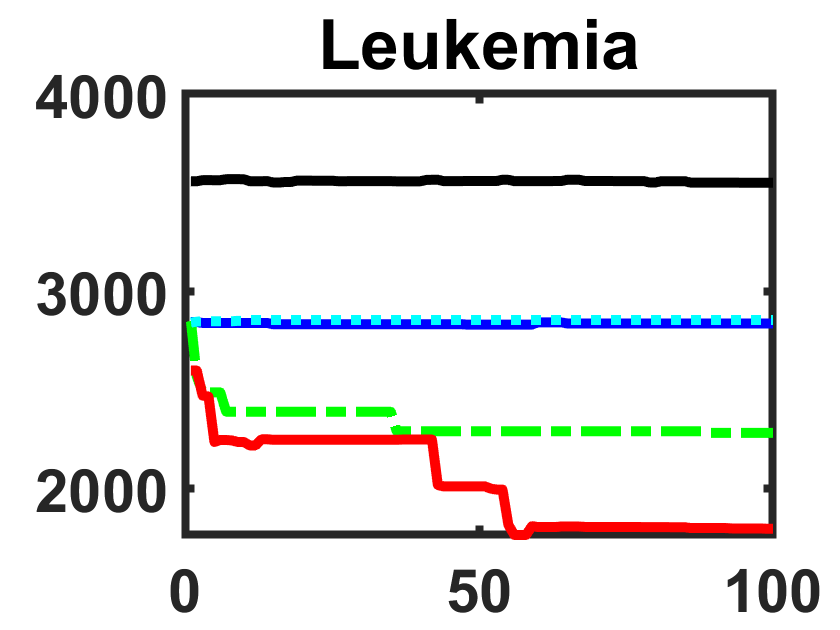}
  \end{minipage}
  \begin{minipage}[b]{0.1612\textwidth}
    \includegraphics[width=\textwidth]{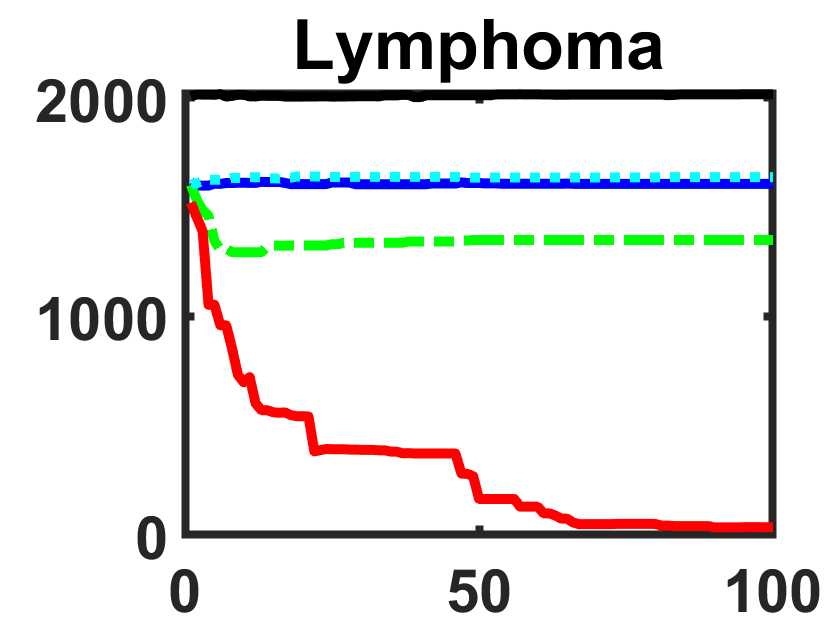}
  \end{minipage}
  \begin{minipage}[b]{0.1612\textwidth}
    \includegraphics[width=\textwidth]{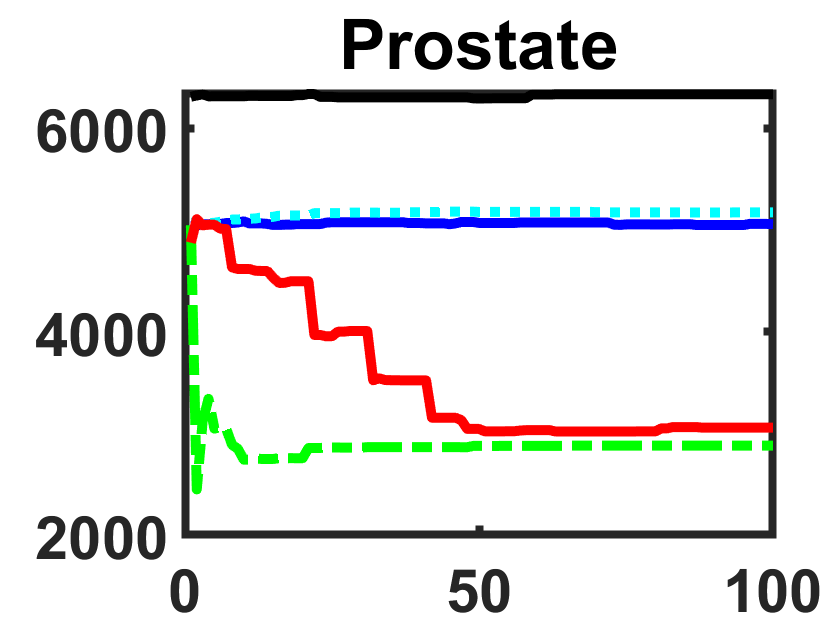}
  \end{minipage}
  \begin{minipage}[b]{0.1612\textwidth}
    \includegraphics[width=\textwidth]{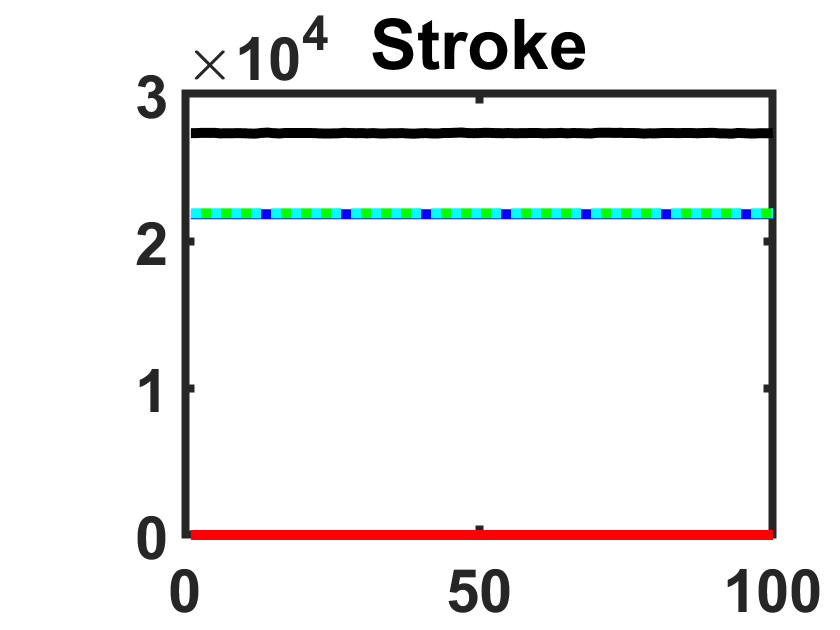}
  \end{minipage}
   \begin{minipage}[b]{0.19\textwidth}
    \includegraphics[width=\textwidth]{physics/label.png}
  \end{minipage}
  \caption{Convergence Curves of Physics-based EC Algorithms in Terms of the Size of Feature Subset}
\end{figure*}

For ABC, max\_limit is the maximum limits allowed. The tau, rat, alpha, beta and rho in ACO are the pheromone value, heuristic desirability, control pheromone, control heuristic and pheromone trail decay coefficient respectively. The c1, c2, w and Vmax in PSO are the cognitive factor, social factor, inertia weight and maximum velocity respectively. The peri, p, Smax, BAR, num\_land1 and beta in MBO are the migration period, ratio, maximum step, butterfly adjusting rate, number of butterflies in land 1 and levy component respectively. For BAT, fmax is the maximum frequency, fmin is the minimum frequency, alpha and gamma are two constants, A\_max is the maximum loudness and r0\_max is the maximum pulse rate. The Pa, alpha and beta in CS are the discovery rate, constant and levy component respectively. The alpha0, beta0, gamma and calpha in FA are the constant, light amplitude, absorbtion coefficient and control alpha respectively. In FPA, beta is the levy component and P is the switch probability. In DE, CR is the crossover rate and F is the constant factor. For genetic algorithm (GA), we use Roulette Wheel and Tournament for selection, which have the same crossover rate (CR) and mutation rate (MR). Tournament also sets the tournament size (Tour\_size). The beta in HHO is the levy component. The b in WOA is the constant. For SA, c is the cooling rate and t0 is the initial temperature. In HS, PAR, HMCR and bw are the pitch adjusting rate, harmony memory considering rate and bandwidth respectively. The G0 in GSA is the initial gravitational constant and the alpha is a constant. For MVO, P is the control TDR, Wmax is the maximum WEP and Wmin is the minimum WEP. The lb and ub in the parameter table are the lower boundary and the upper boundary.

In addition, for all algorithms, we set the number of populations (NP) to 20 and the maximum number of iterations (T) to 100. At the same time, we set a threshold value ($\theta$) of 0.6 to convert the value of the real number field into a discrete value, so as to decide whether to choose the feature of the corresponding field. In this study, KNN classifier is employed with K equals to 3. Our classification accuracy is calculated using a five-fold cross-validation average. We performed each experiment ten times and averaged the data to make sure the experimental results were stable.

\subsection{Converge Curves for Physics-based Heuristic Methods}
We also conducted a comparison between MEL and four physical-based heuristic methods, namely Simulated Annealing (SA), Harmony Search (HS), Gravitational Search Algorithm (GSA), and Multi-Verse Optimizer (MVO). Figures 11 and 12 illustrate the convergence curves of these physical-based methods.

\subsection{Supplementary Tables for Section V, Subsection G}
This part provides the two tables from the seventh sub-section ``Experiments with Larger Data Samples" of the ``Results and Analysis'' section, comparing the subset size and running time. The first table shows the comparison of the average subset sizes generated by different algorithms on 10 larger datasets. It reports the number of selected features by each method. The second table compares the average running times of the algorithms in seconds. It lists the execution time of each algorithm on each dataset. These two tables present a quantitative analysis of how MEL performs relative to other methods in terms of the parsimony of the selected feature subsets and computational efficiency, using larger real-world classification problems. They complement the classification accuracy results discussed in the previous sub-section.

\begin{table}[!htb]
\centering
\resizebox{87mm}{!}{
\begin{tabular}{ l l l l l l l }
\hline \textbf{Dataset}& \textbf{SaWDE}	&	\textbf{FWPSO}& \textbf{DENCA} & \textbf{PSO-EMT } & \textbf{MTPSO} & \textbf{MEL (Ours)} \\  
\hline
BASEHOCK	&	2247.9 	&	\textbf{1.3} 	&	1514.9	&	108.5	&	172.9 	&	1953.9 	\\ \hline
COIL20	&	426.9 	&	\textbf{3.3} 	&	363.6	&	48.1	&	219.5 	&	345.2 	\\ \hline
HAPTDataSet	&	255.8 	&	\textbf{1.3} 	&	177.3	&	20.7	&	79.1 	&	217.2 	\\ \hline
Isolet	&	257.0 	&	\textbf{1.0} 	&	227.0 	&	35.3	&	94.1 	&	243.6 	\\ \hline
madelon	&	205.0 	&	\textbf{1.8} 	&	214.1	&	10.9	&	18.7 	&	204.1 	\\ \hline
MultipleFeaturesDigit	&	258.5 	&	\textbf{1.7} 	&	199.1	&	18.2	&	107.6 	&	240.1 	\\ \hline
Pancancer	&	770.4 	&	\textbf{1.0} 	&	799.7	&	-	&	261.6 	&	708.4 	\\ \hline
PCMAC	&	1453.7 	&	\textbf{1.0} 	&	1083.7	&	40.9	&	162.0 	&	1361.4 	\\ \hline
RELATHE	&	1929.9 	&	\textbf{1.1} 	&	1434.4	&	49.7	&	250.6 	&	1746.9 	\\ \hline
USPS	&	139.8 	&	\textbf{1.0 }	&	97.9	&	-	&	38.7 	&	78.9 	\\ \hline
Average	&	794.5 	&	\textbf{1.5} 	&	611.2 	&	41.5 	&	140.5 	&	710.0 	\\ \hline
\end{tabular}}
\caption{Subset Size Comparison on Datasets with Large Sample Size}
\end{table}

\begin{table}[!htb]
\centering
\resizebox{87mm}{!}{
\begin{tabular}{ l l l l l l l }
\hline \textbf{Dataset}& \textbf{SaWDE}	&	\textbf{FWPSO}& \textbf{DENCA} & \textbf{PSO-EMT } & \textbf{MTPSO} & \textbf{MEL (Ours)} \\ 
\hline
BASEHOCK	&	\textbf{256.3} 	&	1884.8 	&	2884.9	&	372330.9	&	7489.4 	&	862.4 	\\ \hline
COIL20	&	\textbf{35.2} 	&	268.9 	&	234.7	&	85504.1	&	2415.4 	&	121.9 	\\ \hline
HAPTDataSet	&	\textbf{14.9} 	&	127.9 	&	90.2	&	22998.8	&	394.2 	&	58.2 	\\ \hline
Isolet	&	\textbf{26.9} 	&	221.5 	&	132.3	&	82507.8	&	1165.3 	&	105.6 	\\ \hline
madelon	&	\textbf{43.6} 	&	3687.3 	&	604.1	&	198703.8	&	1325.1 	&	169.3 	\\ \hline
MultipleFeaturesDigit	&	\textbf{14.0} 	&	121.1 	&	78.8	&	19404.5	&	442.9 	&	55.0 	\\ \hline
Pancancer	&	\textbf{543.0} 	&	3693.6 	&	8396.1	&	-	&	20684.9 	&	1803.7 	\\ \hline
PCMAC	&	\textbf{157.3} 	&	1241.6 	&	2370.3	&	226997.1	&	5316.9 	&	589.0 	\\ \hline
RELATHE	&	\textbf{121.3} 	&	897.8 	&	1733.5	&	164727.4	&	5028.1 	&	434.3 	\\ \hline
USPS	&	\textbf{208.1} 	&	1499.1 	&	1788.9	&	-	&	14718.1 	&	661.3 	\\ \hline
Average	&	\textbf{142.1} 	&	1364.4 	&	1831.4 	&	119223.2 	&	5898.0 	&	486.1 	\\ \hline
\end{tabular}}
\caption{Running Time Comparison on Datasets with Large Sample Size}
\end{table}

\end{document}